\documentclass[final]{cvpr}

\usepackage{times}
\usepackage{epsfig}
\usepackage{graphicx}
\usepackage{amsmath}
\usepackage{amssymb}
\usepackage{diagbox}
\usepackage{adjustbox}
\usepackage{multirow}
\usepackage{caption}

\makeatletter
\renewcommand{\paragraph}{%
  \@startsection{paragraph}{4}%
  {\z@}{1.25ex \@plus .5ex \@minus .2ex}{-1em}%
  {\normalfont\normalsize\bfseries}%
}
\makeatother

\newif\ifdrafting
\draftingtrue 
\draftingfalse 
\ifdrafting
    \newcommand{\ds}[1]{{\color{red}[DS: #1]}}
    \newcommand{\dv}[1]{{\color{green}[DV: #1]}}
    \newcommand{\mk}[1]{{\color{blue}[MK: #1]}}
    \newcommand{\newedit}[1]{{#1}}
\else
	\newcommand{\ds} [1] {}
	\newcommand{\dv} [1] {}
	\newcommand{\mk} [1] {}
	\newcommand{\newedit}[1]{{#1}}
\fi

\usepackage{pifont}
\newcommand{\cmark}{\ding{51}}%
\newcommand{\xmark}{\ding{55}}%
\newcommand{\mv}[1]{\mathbf{#1}}
\newcommand{\uni}{\mathcal{U}}

\DeclareMathOperator*{\argmin}{arg\,min}
\newcommand{\fig}{Fig.}

\newcommand{\raftsintelclean}{1.95 }
\newcommand{\raftsintelfinal}{2.57}
\newcommand{\raftkitti}{4.23}

\newcommand{\raftsintelcleanbase}{2.08}
\newcommand{\raftsintelfinalbase}{2.75}
\newcommand{\raftkittibase}{4.66}

\newcommand{\pwcsintelclean}{2.17}
\newcommand{\pwcsintelfinal}{2.91}
\newcommand{\pwckitti}{5.76}

\newcommand{\chairnumber}{22,872}
\newcommand{\minautoflownumber}{4}

\usepackage[pagebackref=true,breaklinks=true,colorlinks,bookmarks=false]{hyperref}

\def\cvprPaperID{1698} 
\def\confYear{CVPR 2021}

\newcommand{\ourmethod}{AutoFlow}
\newcommand{\ignore}[1]{}
\begin{document}

\title{AutoFlow: Learning a Better Training Set for Optical Flow}

\author{Deqing Sun, Daniel Vlasic, Charles Herrmann, Varun Jampani, Michael Krainin,  Huiwen Chang, \\
Ramin Zabih, William T. Freeman, and Ce Liu\\
\vspace{1ex}
{Google Research}
}

\maketitle

\begin{figure}[t]
\vspace{-3.2in}
\begin{minipage}{\textwidth}
	\begin{center}
		\newcommand{\figwidth}{0.43\linewidth}
		\newcommand{\Figwidth}{0.53\linewidth}
		\newcommand{\Figheight}{0.4\linewidth}
		\newcommand{\shiftfigure}{\hspace{-1mm}}
		\begin{tabular}{cc}
		\includegraphics[width = \Figwidth]{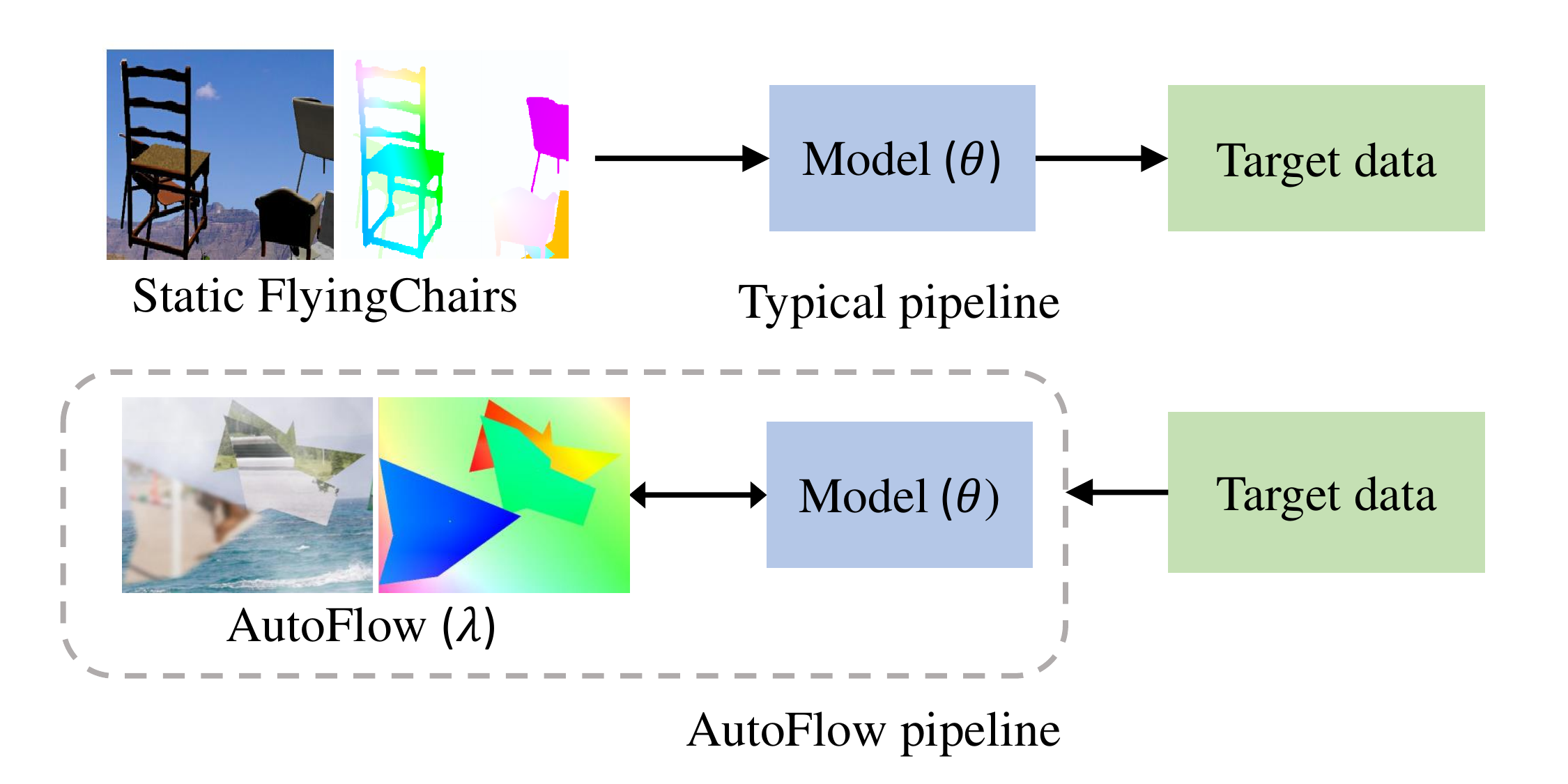} &
		\includegraphics[width = \figwidth]{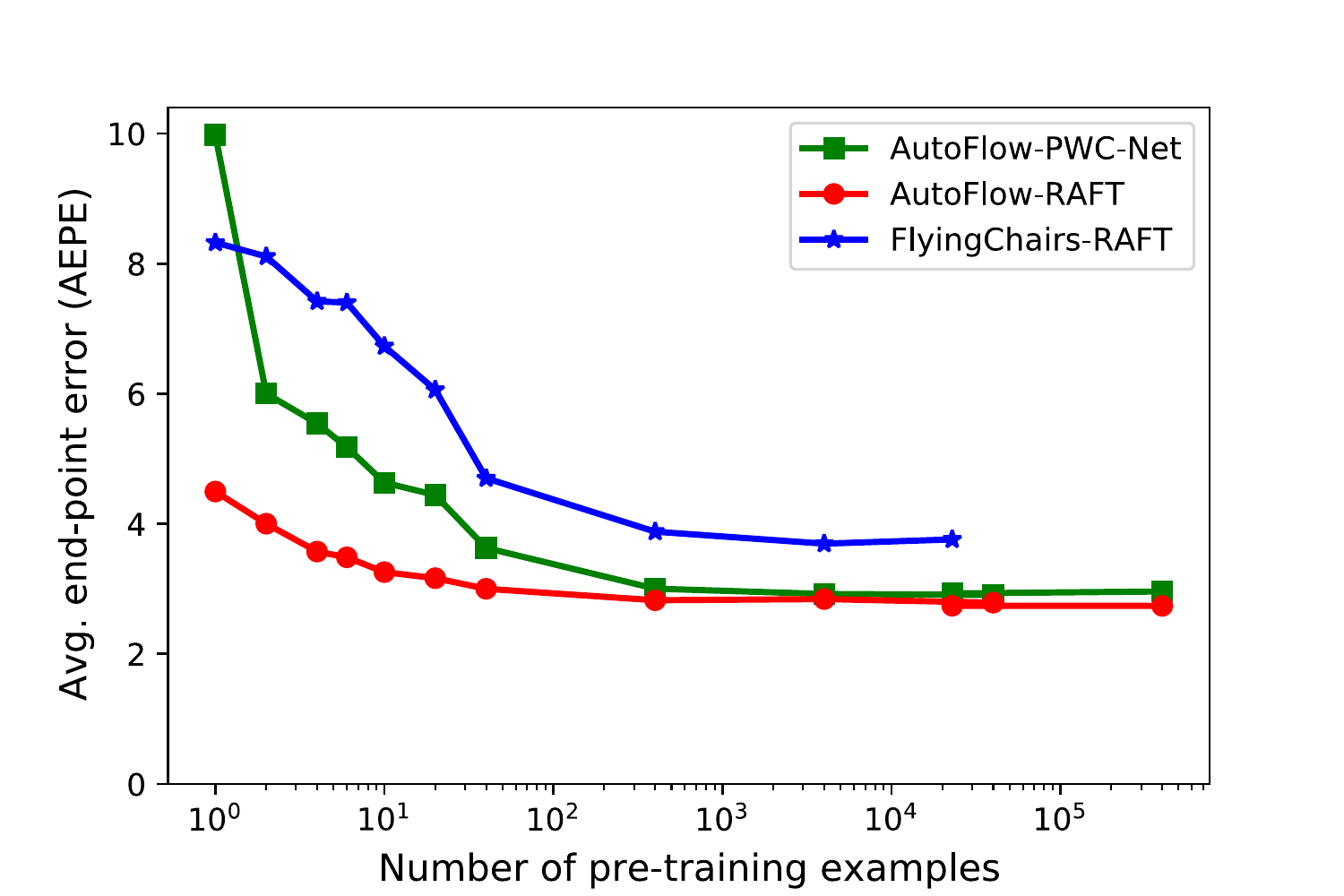} 
		\end{tabular}
		\caption{Left: \textbf{Pipelines for optical flow}. A typical pipeline pre-trains models on static datasets, \eg, FlyingChairs, and then evaluates the performance on a target dataset, \eg, Sintel.  \ourmethod{} learns pre-training data which is  optimized on a target dataset.
		Right: \textbf{Accuracy \wrt number of pre-training examples on Sintel.final.} 
		Four \ourmethod{}  pre-training examples with augmentation achieve lower errors than \chairnumber{} FlyingChairs pre-training examples with augmentation. 
		The  gap between PWC-Net and RAFT becomes small when pre-trained on enough \ourmethod{} examples.
		}\vspace{-15pt}		
		\label{fig:teaser}
	\end{center}
\end{minipage}	
\end{figure}

\begin{abstract}
Synthetic datasets play a critical role in pre-training CNN models for optical flow, but they are painstaking to generate and hard to adapt to new applications. 
To automate the process,  we present {\ourmethod{}}, a simple and effective  method to  render training data for optical flow that optimizes the performance of a model on a target dataset.
\ourmethod{} takes a layered approach to render synthetic data, where the motion, shape, and appearance of each layer are controlled by learnable hyperparameters.
Experimental results show that  \ourmethod{} achieves state-of-the-art accuracy in pre-training both  PWC-Net and RAFT.
Our code and data are available at  \url{autoflow-google.github.io}.
\end{abstract}

\section{Introduction}
\label{sec:intro}

Datasets have been a driving force for the development of AI algorithms.
Convolutional neural networks (CNNs)~\cite{Lecun1998Gradient} were proposed in the 1990's but were not widely adopted for vision tasks until the early 2010's, with the advent of  AlexNet~\cite{krizhevsky2012imagenet}. One key ingredient for deep CNN models was the large amount of manually labeled images,  \eg, from ImageNet~\cite{russakovsky2015imagenet}. The performance gain by AlexNet over shallow models stimulated a paradigm shift in high-level vision tasks. Since then, new models have been invented in rapid succession, even achieving ``superhuman'' performance on image classification tasks~\cite{He2016Deep,huang2016densely}. 

Manual labeling,  however,  cannot provide reliable ground truth for a variety of low-level vision tasks like optical flow and stereo. Since these labels are either difficult or impossible to obtain, synthetic data play a key role in enabling deep models to perform well on such tasks. For example, all top-performing CNN models for optical flow are pre-trained on  two large synthetic datasets, FlyingChairs~\cite{Dosovitskiy:2015Flownet} and  FlyingThings3D~\cite{Mayer:2016:Large}, before being fine-tuned on limited target datasets, \eg, Sintel~\cite{Butler:ECCV:2012} and KITTI~\cite{Geiger:2012:We}.

However, the success of FlyingChairs raises some interesting questions. For example, \emph{how realistic should the rendering be?} 
Several new datasets have been developed to be more realistic than FlyingChairs, such as virtual KITTI~\cite{gaidon2016virtual}, VIPER~\cite{Richter_2017},  and REFRESH~\cite{Lv18eccv}, but none of them have proven more effective than FlyingChairs and FlyingThings3D at pre-training models. In fact, a comprehensive study has revealed that ``realism is overrated''~\cite{Mayer:2016:Large}. 
There are some hypotheses for why FlyingChairs works, \eg, that it has been designed to match the motion statistics  of Sintel, or that it has many thin structures and fine motion details. However, it still remains unclear what set of principles makes an effective optical flow dataset.

To address these questions, we argue that we should make explicit the objective function for rendering training data.
We formulate the generation of training data as a joint optimization problem, which couples rendering the data with training the model. This generation process depends on a set of hyperparameters being optimized. The hyperparameters are evaluated by the performance of the trained  model on a target dataset, as shown in \fig~\ref{fig:teaser}.

To understand what matters, we ask: \emph{how simple can the rendering be?} Thus, we start from an even simpler rendering pipeline than FlyingChairs, a 2D layered approach that requires neither manual labeling nor 3D models.
The motion and shape of each layer are randomly generated according to hyperparameters, as shown in \fig~\ref{fig:rendering}. We can then learn the rendering hyperparameters to optimize the performance of a  model on a target dataset. 

This simple rendering pipeline is surprisingly effective at generating training datasets for optical flow. Trained on its rendered data from scratch, both the recent RAFT model and the widely-used PWC-Net model obtain consistent improvements in accuracy on Sintel and KITTI over the same models trained on FlyingChairs (\fig~\ref{fig:teaser} and Table~\ref{tab:scratch}). 
Further, using \minautoflownumber{} \ourmethod{} examples with  augmentation results in lower errors on Sintel.final for RAFT than using \chairnumber{}  FlyingChairs examples with  augmentation. 
More interestingly, the gap between PWC-Net and RAFT becomes small when trained on enough \ourmethod{} examples.

An analysis of the rendered data also suggests some interesting properties. For example, the motion statistics of the \ourmethod{} dataset  and its augmented version do not resemble those of Sintel (\fig~\ref{fig:motion:stat}) and underrepresent small motions. Though at first glance this distribution may seem abnormal, there may be a simple, intuitive explanation: tiny motion matters little in the overall error.

To summarize, our contributions are the following.
\begin{itemize}
    \item We have introduced, to our knowledge, the first learning approach to render training data for optical flow.
    \item \ourmethod{} compares favorably against FlyingChairs and FlyingThings3D in pre-training RAFT.
    \item \ourmethod{} also leads to  a significant performance gain for PWC-Net, even competitive against RAFT.
    \item We present a detailed analysis of what features are important to dataset generation for optical flow.
\end{itemize}

\begin{table}[t!]
\begin{center}
\begin{adjustbox}{width=0.48\textwidth}
\small
\begin{tabular}{llccc} 
Model &  Dataset  & Sintel.clean & Sintel.final & KITTI\\ 
 \hline
\multirow{3}{*}{PWC-Net} & FlyingChairs &  3.27	& 4.42 & 11.43\\ 
& Chairs~$\!\rightarrow\!$~Things &	2.39	& 3.90 & 9.81 \\
&\ourmethod{} & \textbf{\pwcsintelclean} &	\textbf{\pwcsintelfinal} & \textbf{\pwckitti} \\ 
\hline 
\multirow{3}{*}{ RAFT } & FlyingChairs &2.27	&3.76	&7.63 	\\ 
& Chairs~$\!\rightarrow\!$~Things & \textbf{1.68}	&2.80	&5.92\\
& \ourmethod{}  & {\raftsintelclean} &	\textbf{\raftsintelfinal} & \textbf{\raftkitti} \\ 
\end{tabular}
\end{adjustbox}
\vspace{3pt}
\caption{ \textbf{AEPE results for pre-training}. \ourmethod{} can better train RAFT and PWC-Net from scratch than the widely-used FlyingChairs dataset and perform competitively against the FlyingChairs~$\!\rightarrow\!$~FlyingThings3D schedule. 
}
\label{tab:scratch}
\vspace{-10pt}
\end{center}
\end{table}

\section{Related Work}
\label{sec:related}
\paragraph{Datasets for high-level computer vision}

Manually labeled datasets, such as ImageNet~\cite{russakovsky2015imagenet}, PASCAL~\cite{everingham2010pascal}, MS-COCO~\cite{lin2014microsoft}, and CityScapes~\cite{cordts2016cityscapes}, have been widely adopted for high-level vision tasks.
However, manual labeling is  hard to scale, and quite a few synthetic datasets have been developed~\cite{ros2016synthia,dosovitskiy2017carla,gaidon2016virtual,kar2019meta}.
Meta-Sim~\cite{kar2019meta} learns to minimize the distribution gap between the rendered  and  target datasets and can also optimize  task performance.  However,  Meta-Sim can model only limited scenes because it relies on obtaining valid scene structures from a grammar. 

RenderGAN~\cite{sixt2018rendergan} learns to augment the dataset for handwriting classification.
{Differentiable rendering}~\cite{loper2014opendr,ravi2020accelerating} enables  gradients to be passed to rendering parameters, which, however, do not directly relate to the scene distribution hyperparameters. 
Yang and Deng~\cite{yang2020learning} proposed a ``hybrid gradient'' approach to make use of analytical gradients whenever available. 
These methods focus on the generation of a single image and cannot directly apply to the generation of optical flow.

\paragraph{Datasets for optical flow}

Similar to other vision tasks, datasets have been the driving force behind the development of optical flow. 
However, unlike high-level vision tasks, it is only possible to obtain ground truth under controlled lab environments~\cite{Baker:2011:DEO} or rigid scenes/objects~\cite{Geiger:2012:We,kondermann2016hci}. Early work relied on synthetic datasets for evaluation, such as the well-known ``Yosemite'' sequence~\cite{Barron:1994:PO}. MPI-Sintel~\cite{Butler:ECCV:2012}, one of the leading benchmark datasets for optical flow, was rendered using the Blender engine.
Roth and Black~\cite{roth2007spatial} used real depth data to render synthetic data, which is limited to static scenes. KITTI~\cite{Geiger:2012:We} was created using LIDAR for static scenes and later extended to rigidly moving cars~\cite{Menze2015ISA} for autonomous driving applications.

Dosovitskiy \etal~\cite{Dosovitskiy:2015Flownet} created a synthetic dataset, FlyingChairs. Mayer \etal~\cite{Mayer:2016:Large} further introduced a large dataset for optical flow and related tasks, FlyingThings3D. Ilg \etal~\cite{Ilg:2016:Flownet2} found that sequentially training on FlyingChairs and then on FlyingThings3D obtains the best results; this has since become standard practice in the field.
Efforts to improve these two datasets include the autonomous driving scenario~\cite{gaidon2016virtual}, more realistic rendering~\cite{Richter_2017},  realistic backgrounds from SLAM~\cite{Lv18eccv}, and human datasets~\cite{ranjan2020learning}. However, none have proven more effective than FlyingChairs and FlyingThings3D for pre-training.

Mayer \etal~\cite{Mayer2018makes} performed a comprehensive study of synthetic datasets for optical flow and disparity estimation. They developed each synthetic dataset heuristically, with no regard for target dataset performance. Our rendering pipeline is largely inspired by their 2D rendering techniques. But instead of designing each dataset by hand, we learn these parameters via jointly solving rendering and training to optimize the performance on a target dataset.

\paragraph{CNN models for optical flow}
The seminal FlowNet paper~\cite{Dosovitskiy:2015Flownet} pioneered the CNN-based approach for optical flow. Its follow-up, FlowNet2~\cite{Ilg:2016:Flownet2}, significantly improved FlowNet's performance by stacking several sub-networks into one large model. 
Spy-Net~\cite{Ranjan:2016:SpyNet}, PWC-Net~\cite{sun2018pwc}, and LiteFlowNet~\cite{Hui_2018_CVPR} were designed using several well-established principles for optical flow. For the first time, PWC-Net obtained more accurate results on the Sintel and KITTI benchmarks than traditional approaches. Quite a few new network architectures were proposed based on the PWC-Net framework~\cite{hur2019iterative,yang2019volumetric,jiang2019sense,yin2019hierarchical}.
Recently, Teed and Deng~\cite{teed2020raft} introduced the RAFT architecture, which used a recurrent architecture to obtain a significant performance gain over its predecessors on Sintel and KITTI. 

The advances in these network architectures have significantly improved their performance on benchmark datasets. However, all these models follow nearly the same training procedures, \ie, pre-training on FlyingChairs and FlyingThings3D and then fine-tuning on limited training data on the target domain. 
In this paper, we focus on dataset generation and show that it is possible to achieve accuracy similar to  or better than that of FlyingChairs and FlyingThings3D in pre-training by learning to render training data. We learn the rendering hyperparameters for the recent  RAFT model  and find that they also apply to PWC-Net.

\paragraph{Evaluating CNN models for optical flow}
The improvement in accuracy comes from innovations on both the model architecture and the training procedures. Previous work~\cite{sun2019models} shows that changes in training procedure result in significant performance boosts for FlowNetC and PWC-Net. Here we find that changing the datasets and incorporating recent practices in training significantly improves PWC-Net and narrows down its performance gap from  RAFT.

\paragraph{Self-supervised and semi-supervised learning of optical flow}
Significant progress has been made on self-supervised optical flow~\cite{jonschkowski2020matters,liu2019selflow}. However, state-of-the-art, self-supervised methods still lag behind supervised ones,  e.g., models pre-trained on FlyingChairs and FlyingThings~\cite{teed2020raft,yang2019volumetric} are more accurate on Sintel than models~\cite{jonschkowski2020matters,liu2019selflow} trained on Sintel image pairs using self-supervised loss.

\paragraph{Learning to learn}
A recent trend in neural network research is learning to learn, which aims at automating the manual process of network design or hyperparameter selection. Existing methods mainly focus on learning hyperparameters for the architecture~\cite{zoph2016neural}, loss function, optimization, and augmentation~\cite{cheng2020improving,cubuk2018autoaugment}. In contrast, we focus on learning to render synthetic training data for optical flow.

\section{Generating training data}
\label{sec:data}
\begin{figure}[t]
	\begin{center}
		\includegraphics[width=\columnwidth]{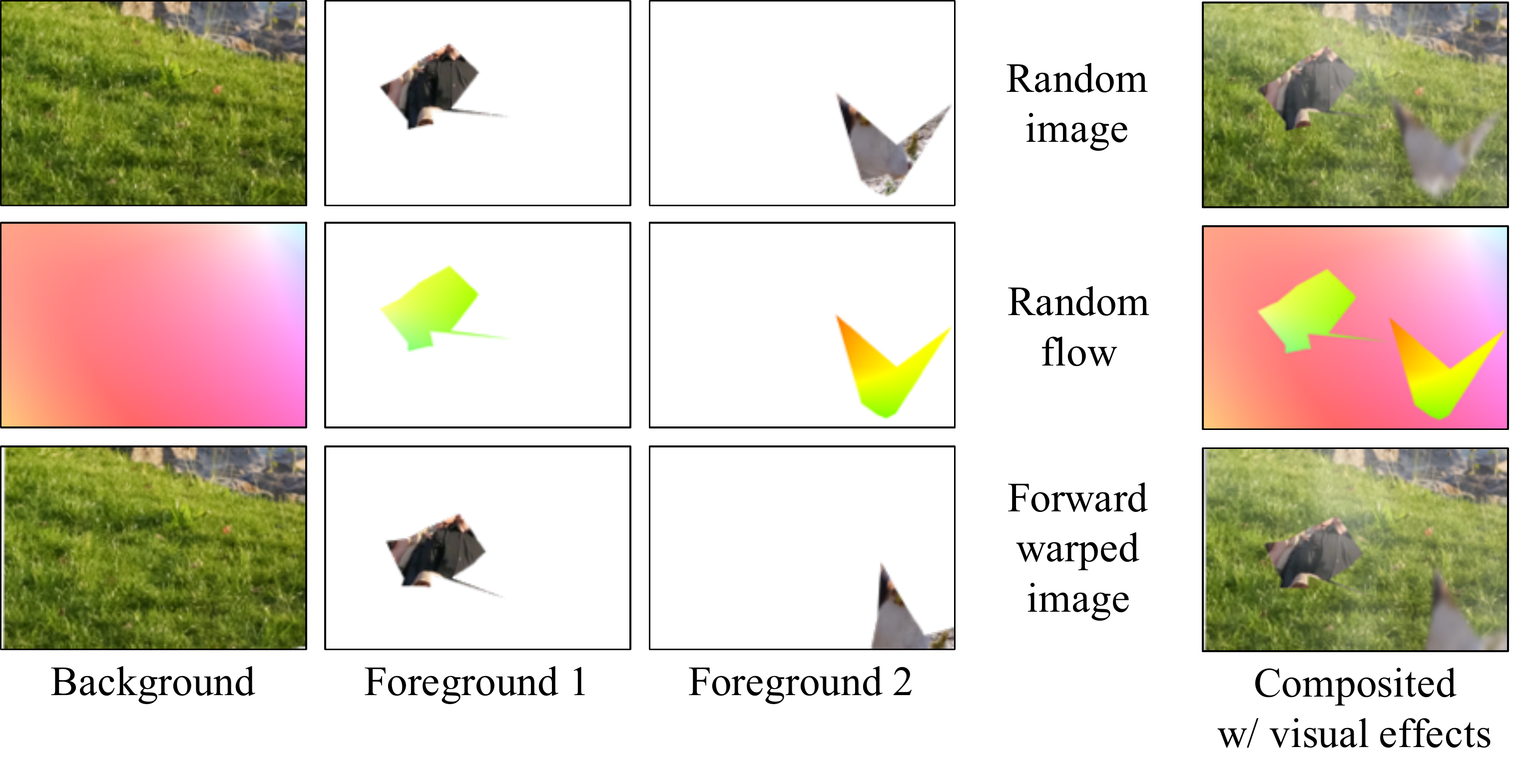} 
		\caption{\textbf{Rendering pipeline} for \ourmethod{} uses a layered approach.  Each layer is created from a random image and mask (top row), its flow field is randomly generated (middle row), and the layer is accordingly warped (bottom row).  All layers are alpha-composited back-to-front and undergo visual effects such as motion blur and fog (right column).}
		\label{fig:rendering}
	\end{center}
\end{figure}

We take a layered approach~\cite{Wang:1994:RMIL,Sun:NIPS:10} to rendering image pairs and their optical flow, as shown in \fig~\ref{fig:rendering}. For the first frame, we randomly sample K images $\mv{I}_1^k$ from an image dataset and order them by depth, with the first layer being the background. Next, we sample an alpha mask $\mv{M}_1^k$ (section~\ref{sec:data_mask}) and an optical flow field $\mv{W}^k$ (section~\ref{sec:data_warp}) for each layer according to the rendering hyperparameters (section~\ref{sec:hyperparameters}). The optical flow field is used to warp the image and the mask into the second frame:
\begin{align}
    \mv{I}_2^k &= f(\mv{I}_1^k, \mv{W}^k) & 1 \leq k \leq K, \\
    \mv{M}_2^k &= f(\mv{M}_1^k, \mv{W}^k) & 1 \leq k \leq K, \nonumber
\end{align}
where $f$ represents the forward warping function according to the flow field.

We composite the images and the flow with back-to-front alpha blending, starting with the background layer:
\begin{align}
    \mv{I}^k &= \mv{M}^k \odot \mv{I}^k + \left( 1 - \mv{M}^k \right) \odot \mv{I}^{k-1}, \\
    \mv{W}^k &= \mv{\bar{M}}^k \odot \mv{W}^k + \left( 1 - \mv{\bar{M}}^k \right) \odot \mv{W}^{k-1}, \nonumber
\end{align}
where $\mv{\bar{M}}$ is the alpha mask binarized around its middle value, and  $\odot$ denotes the element-wise product and broadcasts to the channel dimension. We slightly abuse the notation, using the same symbols for  images and masks before and after composition, as well as dropping  subscripts.

Finally, we apply certain visual effects (section~\ref{sec:data_effects}) to the images to cover some natural variations in videos. Figure~\ref{fig:samples} shows examples of complete images and their flows.

\subsection{Object Masks}
\label{sec:data_mask}

The background layer has a fully opaque mask. For each foreground layer, we test two ways of generating the mask: random polygons and manual segmentation.

\begin{figure}[t]
	\begin{center}
		\newcommand{\ImgWidth}{0.25\columnwidth}
		\begin{tabular}{@{} c @{} c @{} c @{} c @{}}
		\includegraphics[width = \ImgWidth{}]{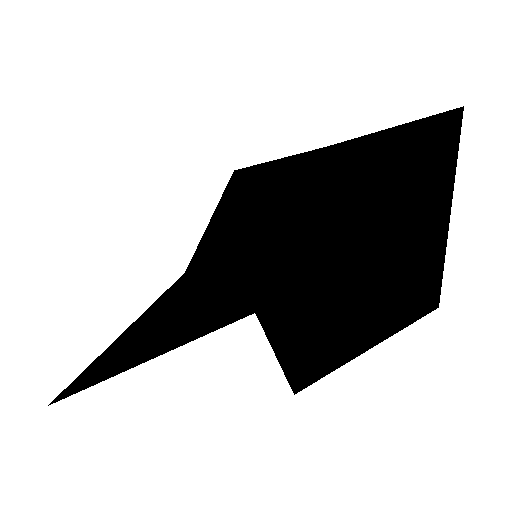} &
		\includegraphics[width = \ImgWidth{}]{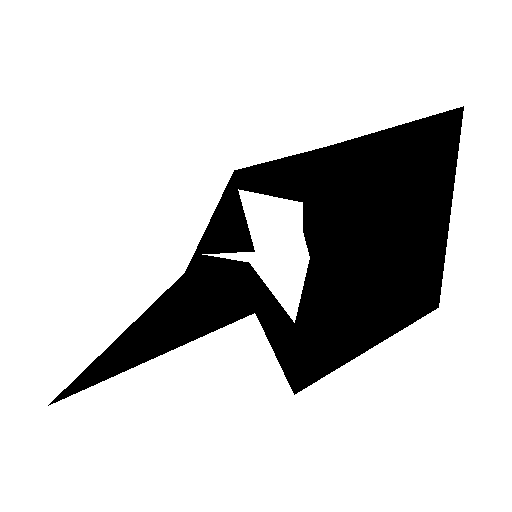} &
		\includegraphics[width = \ImgWidth{}]{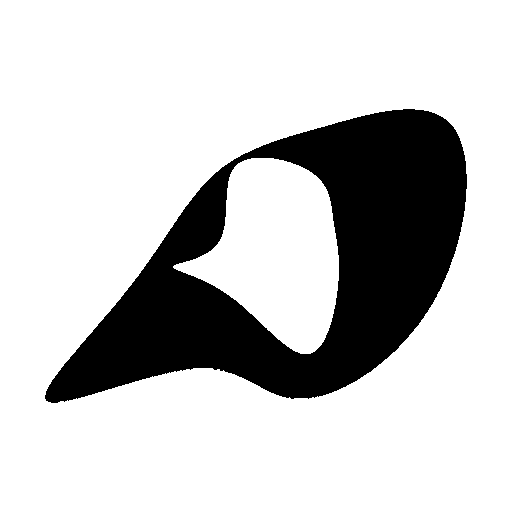} &
		\includegraphics[width = \ImgWidth{}]{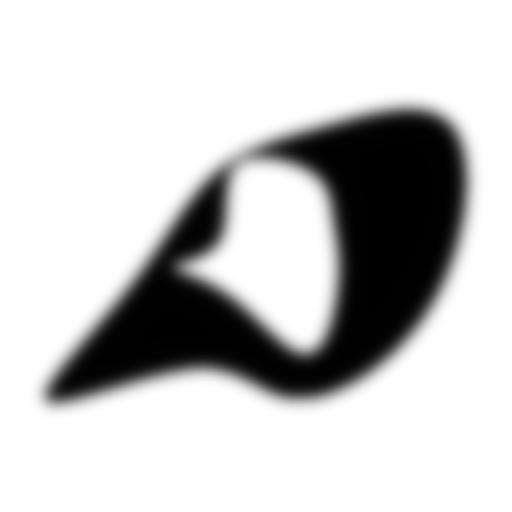}
		\end{tabular}
		\vspace{-10pt}
		\caption{\textbf{Foreground object masks} are random polygons that can have holes, smoothed edges, and blurred boundaries. Mask values have been inverted for visualization.}
		\label{fig:polygon}
		\vspace{-10pt}
	\end{center}
\end{figure}

\paragraph{Random polygons} For each foreground layer, we generate a random polygon~\cite{Mayer2018makes} to serve as its alpha mask. Each polygon has a random number of sides, with vertices randomly sampled in angle and radius around a center. Each polygon can also have a hole, which itself is a smaller random polygon. Further, we can control polygon smoothness through subdivision. Finally, the mask can be blurred with a Gaussian filter in order to feather its boundary (this is applied to both polygon and manual object masks). Examples of random polygon masks are shown in \fig~\ref{fig:polygon}.

\paragraph{Manual segmentation} To make foreground objects more semantically congruent, we use the images and manual labels from OpenImages~\cite{OpenImages}. The location and size of each foreground object within the image are randomly sampled.  

\subsection{Motion Model}
\label{sec:data_warp}

The motion of each layer is a combination of rigid transformation (scale, rotation, translation), perspective distortion, and a \emph{bilinear grid warp}. A bilinear grid warp of size $n$ is a set of flow vectors defined on the vertices of a $n \times n$ grid, then bilinearly interpolated in the interior of the grid (the grid being uniformly distributed over the image)~\cite{szeliski2010computer}. This allows for more complex forward flow with a fast analytic solution for forward image warping (we invert the bilinear interpolating function within each grid cell, which boils down to solving a quadratic equation). In fact, all of our base motions can be modeled with a bilinear grid warp: rigid transform can be expressed by rigidly moving the corners of a grid, and perspective distortion by independently moving the corners of a $2 \times 2$ grid.

\begin{figure}[t]
	\begin{center}
        \includegraphics[width=.9\columnwidth]{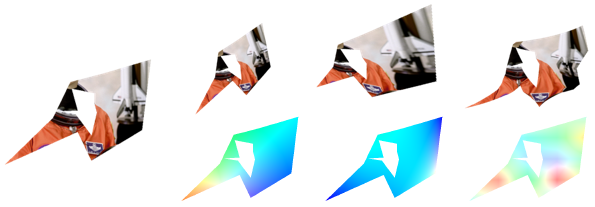}
		\caption{\textbf{Base motions} a foreground object (left) can undergo (left-to-right): rigid transformation, perspective distortion, and bilinear grid warp.}
		\label{fig:warp}
		\vspace{-10pt}
	\end{center}
\end{figure}

For the foreground layers, we employ all modalities of motion (rigid + grid), while for the background we only apply moderate perspective distortion. Figure~\ref{fig:warp} demonstrates our motion modalities on a sample foreground object.

\subsection{Visual Effects}
\label{sec:data_effects}

\begin{figure}[t]
    \begin{center}
		\begin{tabular}{ccc}
		\includegraphics[width=0.27\columnwidth]{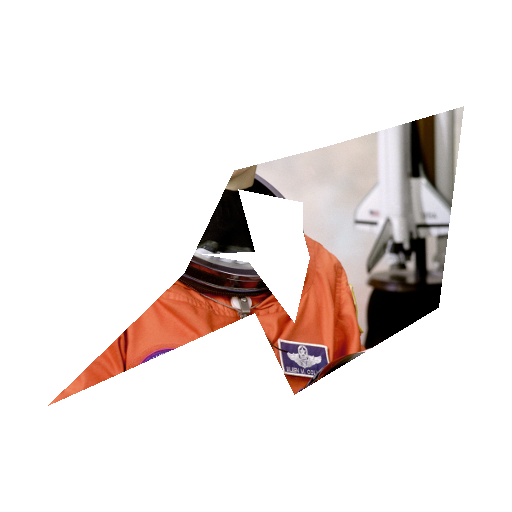} &
		\includegraphics[width=0.27\columnwidth]{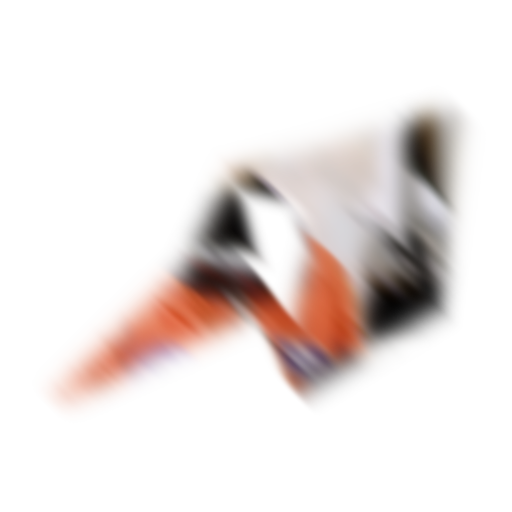} &
		\includegraphics[width=0.27\columnwidth]{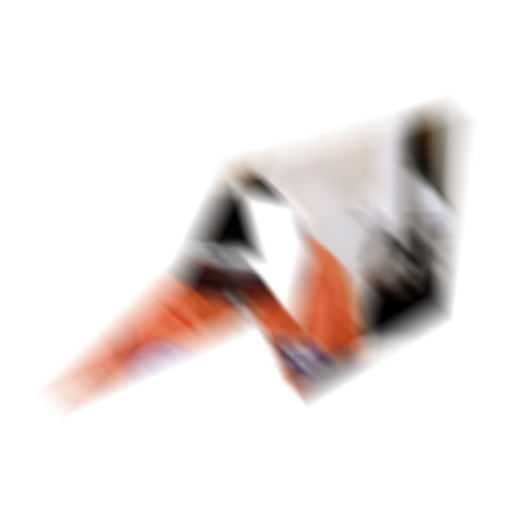} \\
		\includegraphics[width=0.27\columnwidth]{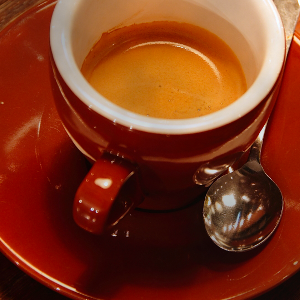} &
		\includegraphics[width=0.27\columnwidth]{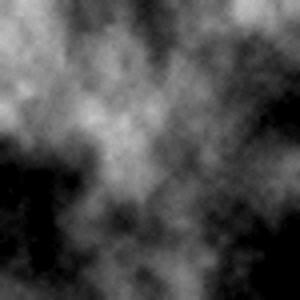} &
		\includegraphics[width=0.27\columnwidth]{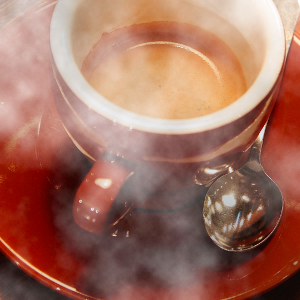}
		\end{tabular}
		\caption{\textbf{Visual effects} improve performance on realistic datasets. Top row shows an object motion-blurred due to diagonal movement using Gaussian (middle) or box (right) filters. Bottom row shows a random semi-transparent fog overlaid on top of an image.}
        \label{fig:visual_effects}
        \vspace{-10pt}
    \end{center}
\end{figure}

To generalize better to more realistic video data, we simulate common visual effects including motion blur and fog (\fig~\ref{fig:visual_effects}). These effects only modify the image data and have no influence over the ground truth flow.

\paragraph{Motion blur}
We approximate the motion blur of each layer by applying a  filter to both the image and the mask. Standard deviations of the filter are computed by taking a proportion of the average absolute flow in each dimension over all the pixels within the mask. 
We apply the same motion blur filter to both the first and second images.

\paragraph{Fog}
To simulate fog, we generate a white image with a random semi-transparent alpha mask and overlay it on top of the composited initial and final images. The fog does not move between the images, nor does it affect the ground truth flow. To compute the alpha mask, we generate several random normal images of various resolutions, with their standard deviations being inversely proportional to their resolutions. We then bicubically resample each to the desired fog resolution and sum them up. Finally, we adjust the resulting image so that its mean and standard deviation match controllable hyperparameters.

\subsection{Data Augmentation}
\label{sec:randaug}
To increase the diversity of the training data, we apply data augmentation~\cite{Dosovitskiy:2015Flownet,sun2018pwc} to the rendered data. 
Inspired by RandAugment~\cite{cubuk2020randaugment}, we randomly select several transformations among rotation, scale, squeeze, translation, and additive noise at each iteration. The number of transformations and their strength levels are hyperparameters to learn.

\subsection{Hyperparameters}
\label{sec:hyperparameters}

During training, we tune a number of hyperparameters that dictate data generation and augmentation, including the shape, size, and position of masks, the complexity and magnitude of motion,  and the visual effects. Respective values are uniformly sampled from the specified ranges, and the ranges are hyperparameters to learn. Please refer to the appendix for the detailed list of hyperparameters.

\ignore{\begin{figure*}[t]
	\begin{center}
		\newcommand{\Figwidth}{\linewidth}
		\includegraphics[width = \Figwidth]{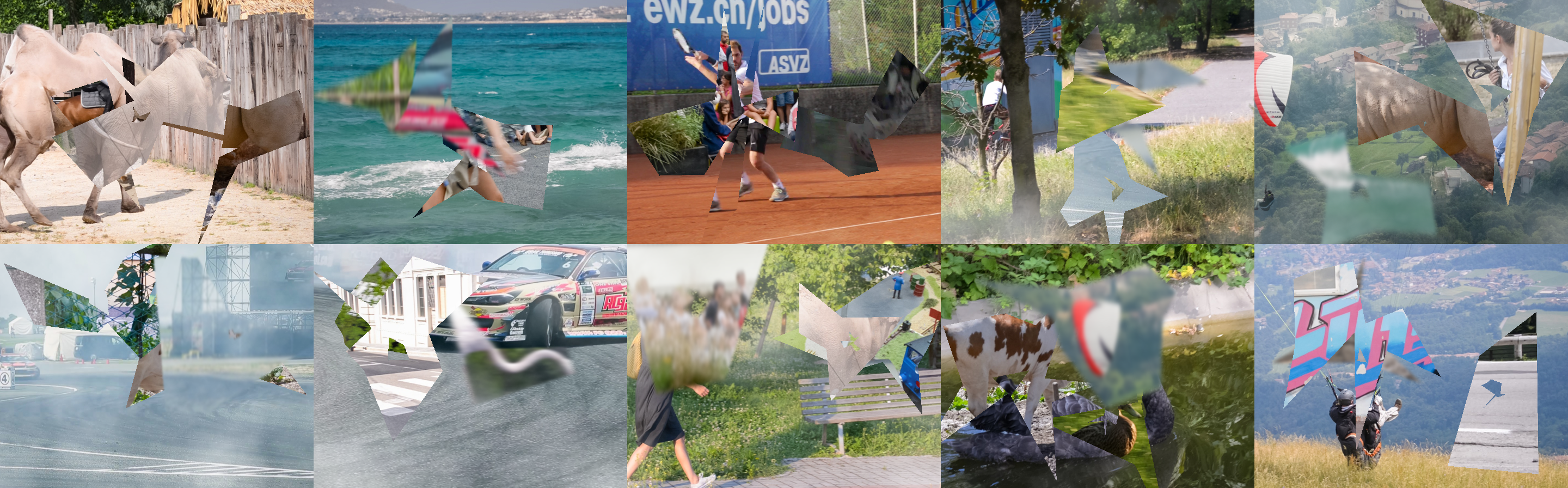} \\
		\includegraphics[width = \Figwidth]{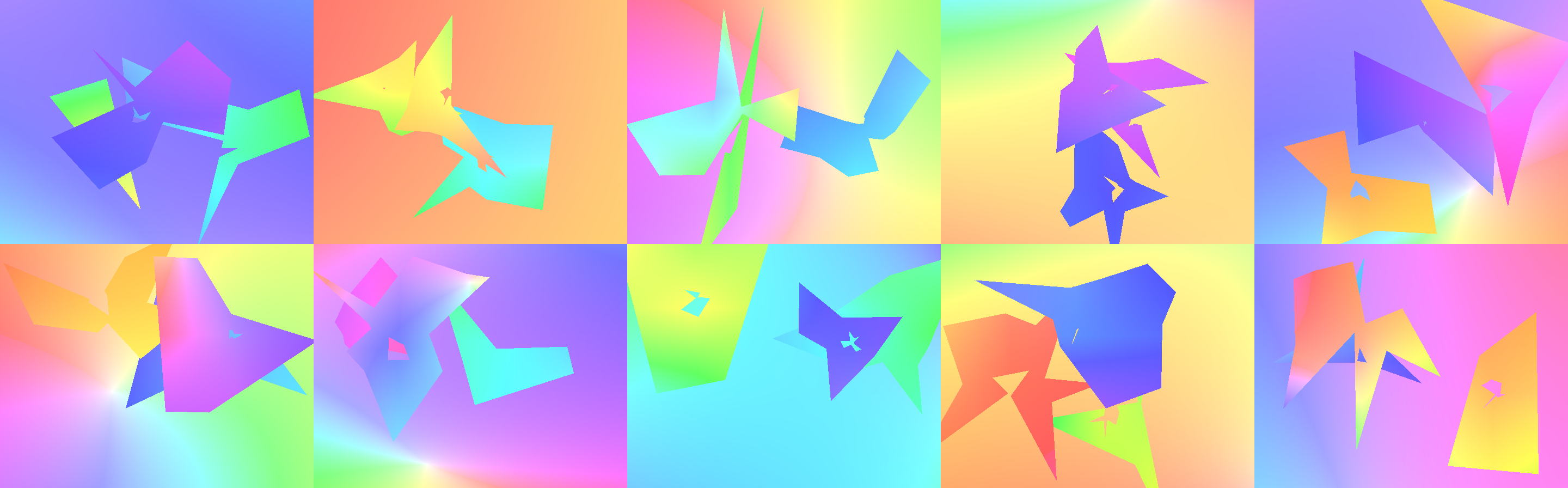} 
		\caption{Samples of synthetic data. Top: first images; bottom: visualized flow fields. See the appendix for more examples. \ds{Daniel will update} }
		\label{fig:samples}
	\end{center}
\end{figure*}
}

\begin{figure*}[t]
  \centering
  \setkeys{Gin}{width=0.2\linewidth}
  
  \includegraphics{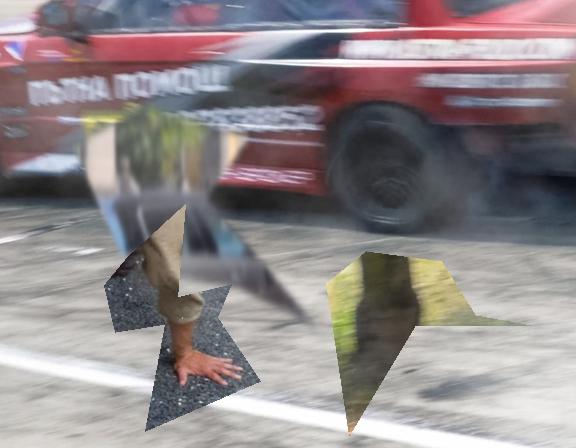}\,%
  \includegraphics{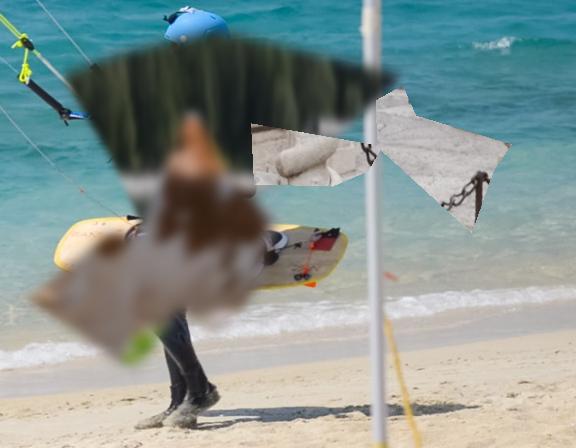}\,%
  \includegraphics{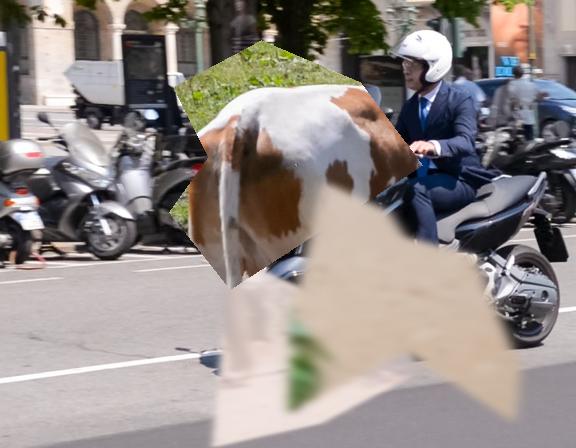}\,%
  \includegraphics{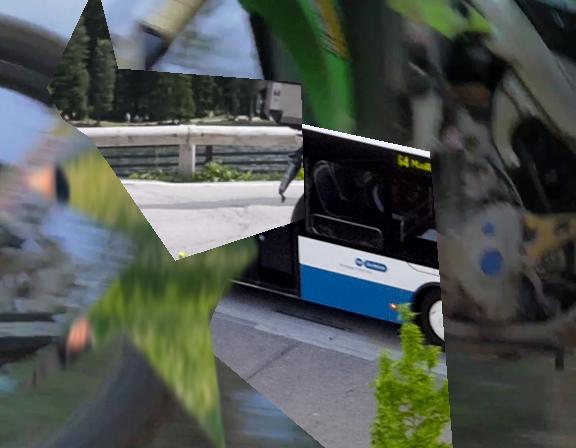}\,%
  \includegraphics{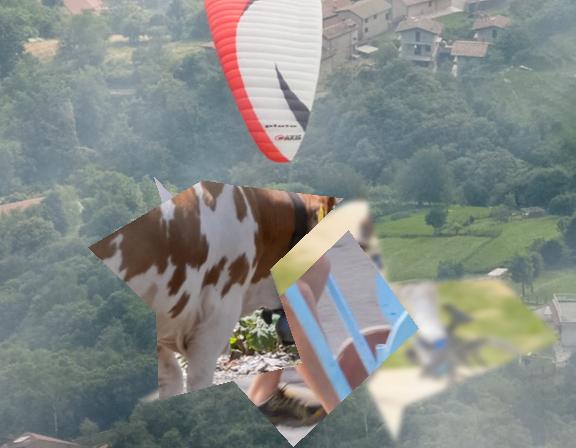}

  \includegraphics{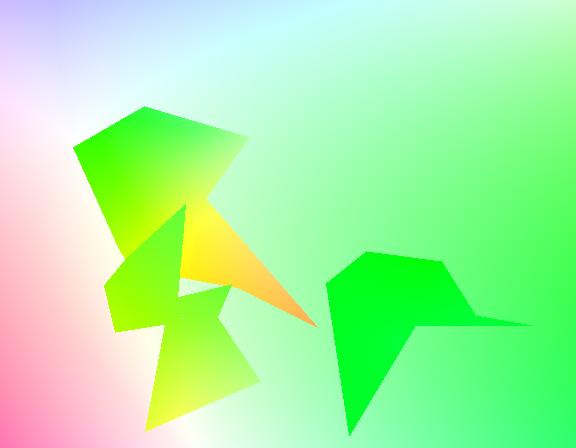}\,%
  \includegraphics{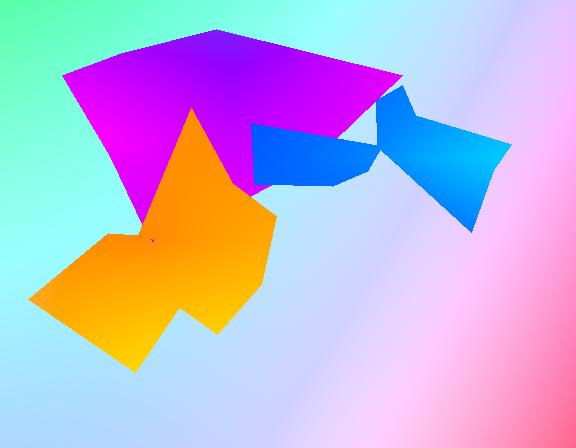}\,%
  \includegraphics{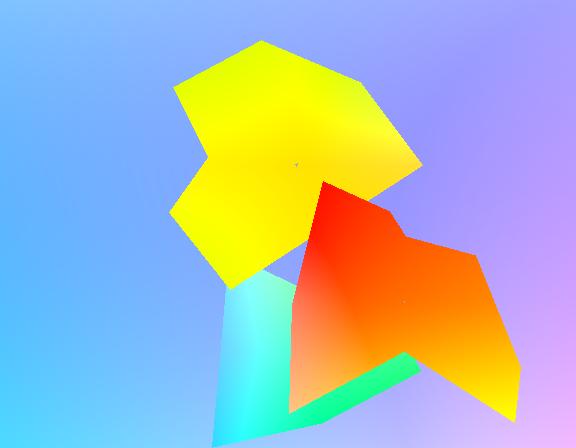}\,%
  \includegraphics{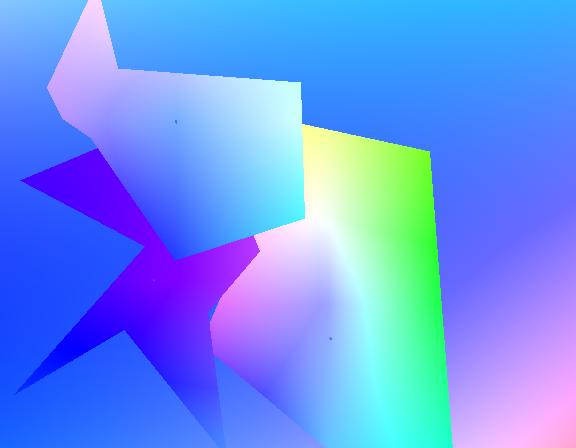}\,%
  \includegraphics{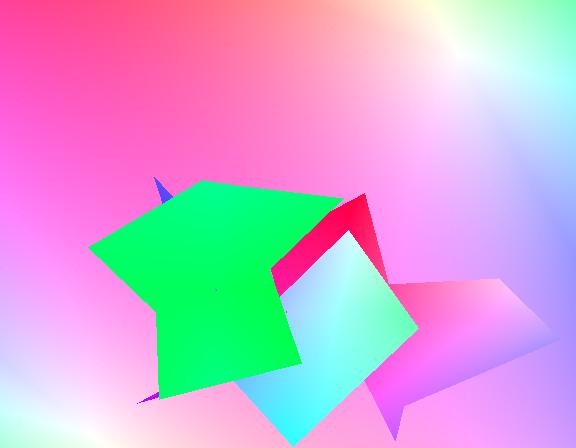}
  \caption{\textbf{Samples of \ourmethod{}}. Top: first images; bottom: visualized flow field. 
  }		
\label{fig:samples}
\end{figure*}

\section{Learning to Render Training Data}
\label{sec:formulation}

Given a target dataset for optical flow, \eg, Sintel or KITTI, we want to learn the hyperparameters to render training data so that a CNN model trained on the rendered data has optimal performance on the target dataset. Every set of hyperparameters corresponds to a rendered training dataset. In this section, we will first present the learning objective and then the search algorithm.

\subsection{Problem Formulation}

Given the  rendering pipeline for generating training data and the  range for the rendering  hyperparameters $\Lambda$, our goal is to search for the set of optimal hyperparameters $\lambda$ that optimizes a  metric $\Omega$ on a model $\theta$, 
\begin{align}
    \lambda^* = \argmin_{\lambda \in \Lambda} {\Omega} \left (\theta(\lambda) \right ).
\end{align}
The model $\theta$ minimizes a loss function $\mathcal{L}$ on the rendered datasets according to the set of hyperparameters $\lambda$
\begin{align}
    \theta(\lambda) = \argmin_{\theta} \mathcal{L} \left (\mv{W}(\lambda), \phi_{\theta}(\mv{I}_{1} \big (\lambda), \mv{I}_{2}(\lambda) \big ) \right),
\end{align}
where the model $\theta$ includes the parameters of a network $\phi_{\theta}$ that maps two input images to their optical flow. Be default, we use the sequence loss function proposed by RAFT as the loss function and the average end-point error (AEPE) as the metric on the target datasets unless  stated otherwise.

\subsection{Hyperparameter Search Algorithm}
\label{subsec:search}

To learn the hyperparameters for rendering the  dataset, we develop a hybrid algorithm based on the population-based training  (PBT) algorithm~\cite{jaderberg2017population,cheng2020improving} and the Covariance Matrix Adaptation Evolution Strategy (CMA-ES) algorithm~\cite{hansen2016cma}.
Specifically, we classify the hyperparameters into subgroups and use the  CMA-ES algorithm to search the selected subgroups of hyperparameters to optimize the learning metric. 
CMA-ES  maintains a sampling distribution over the search space. It samples a few points, evaluates them, and updates the distribution based on the ranking of the points \wrt to the learning metric. The sampling distribution is a multivariate Gaussian whose covariance matrix is adapted over time.
Our algorithm takes $\mathcal{N}$ iterations, with each  iteration training $\mathcal{M}$ in parallel. 
The time complexity grows linearly w.r.t. the number of search iterations and the number of training steps per search.

\section{Experimental Results}
\label{sec:results}

\paragraph{Implementation details}
We randomly sample images from different sequences of the Davis dataset~\cite{Perazzi2016} as appearances for each layer. Our baseline is a TensorFlow implementation of RAFT~\cite{tf-raft}, the  performance of which is similar to that of the official PyTorch implementation.  Throughout this section, we refer to our method or the data it generates as \ourmethod{}. By default, we use the average end-point error (AEPE) on the final pass of Sintel training dataset (Sintel.final) as the learning metric because it is currently the most challenging dataset.

Empirically, it takes about 7 days to finish 8 searching iterations using 48 NVIDIA P100 GPUs, with each iteration training 8 models in parallel. Hyperparameters about 5\% less accurate are often found within 2 days.  Alternatively, the time can also be reduced to less than 2 days by using fewer training steps (40k) and then reusing the searched hyperparameters for the full 200k steps. That has roughly a 3\% drop in accuracy on Sintel.

\subsection{\ourmethod{} Versus the State of the Art}
\begin{figure*}
    \begin{center}
    \small
        \newcommand{\figwidth}{0.035\linewidth}
        \newcommand{\Figwidth}{0.24\linewidth}
        \newcommand{\Figheight}{0.4\linewidth}
        \newcommand{\shiftfigure}{\hspace{-1mm}}
        \begin{tabular}{cccc}
        \shiftfigure \includegraphics[width = \Figwidth]{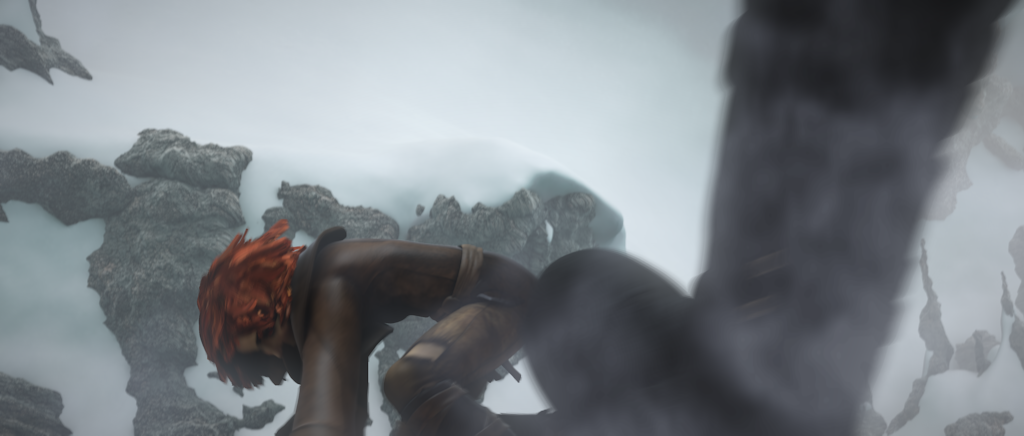} &
        \shiftfigure \includegraphics[width = \Figwidth]{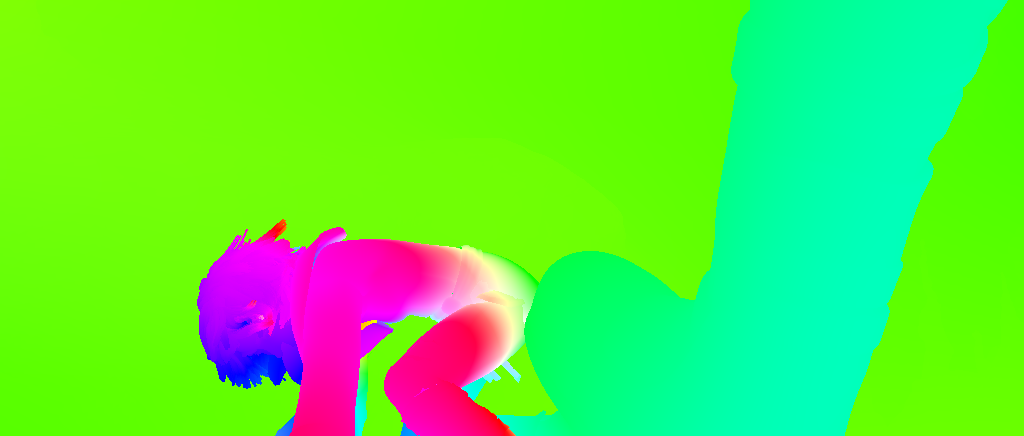}  &
        \shiftfigure \includegraphics[width = \Figwidth]{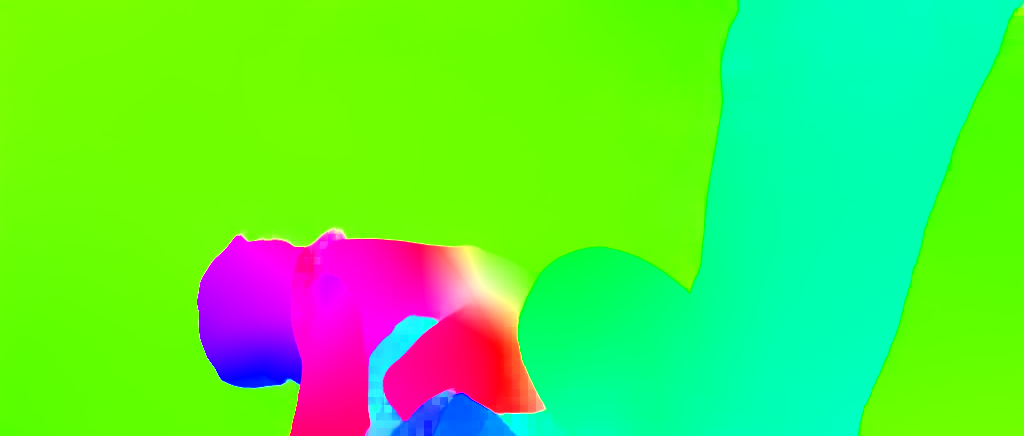} &
        \shiftfigure \includegraphics[width = \Figwidth]{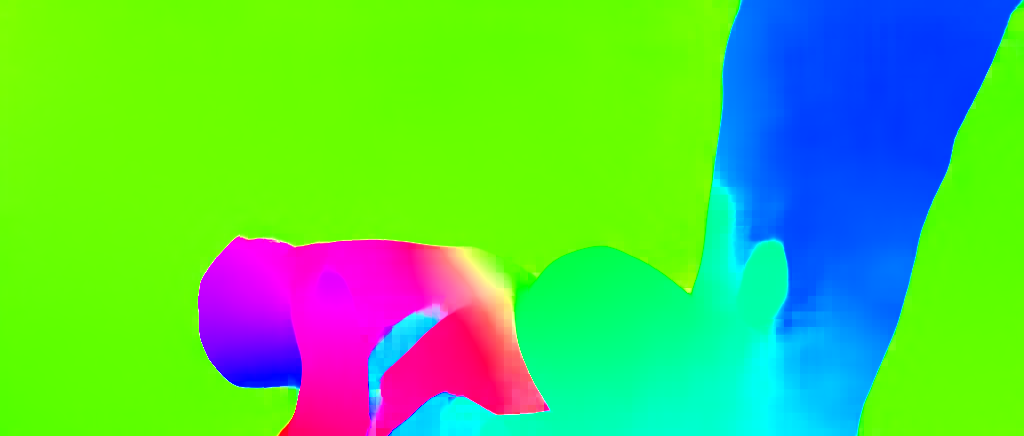}   \\ 
        \shiftfigure  Frame 29 of final Ambush\_4 & \shiftfigure Ground truth & \shiftfigure  \ourmethod{} & \shiftfigure  FlyingChairs \\

        \shiftfigure \includegraphics[width = \Figwidth]{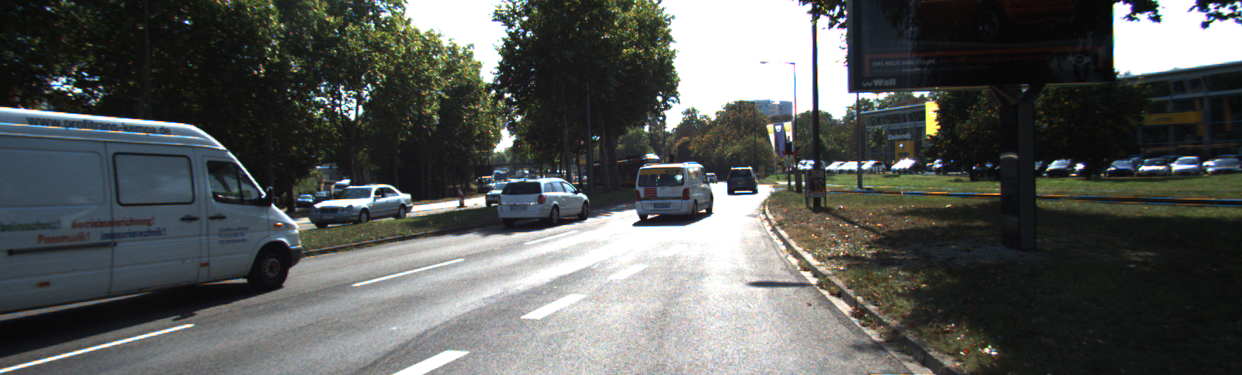} &
        \shiftfigure \includegraphics[width = \Figwidth]{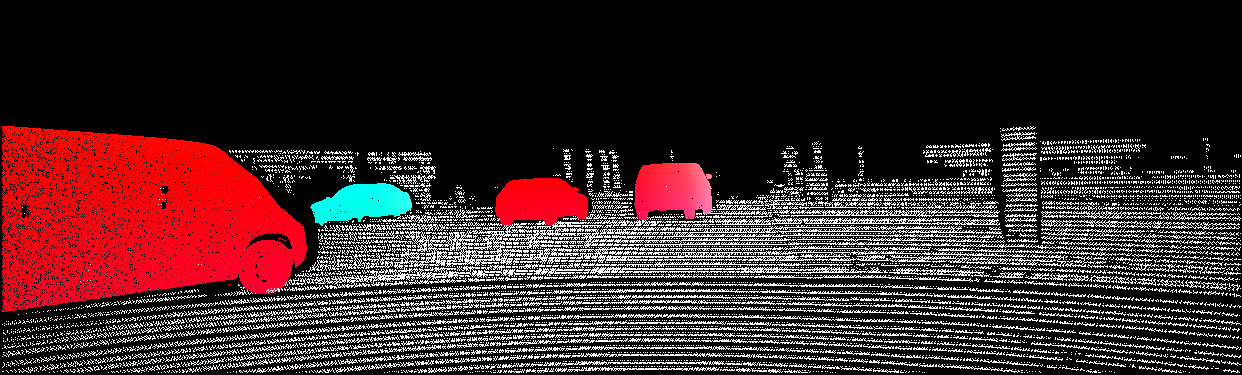}  &
        \shiftfigure \includegraphics[width = \Figwidth]{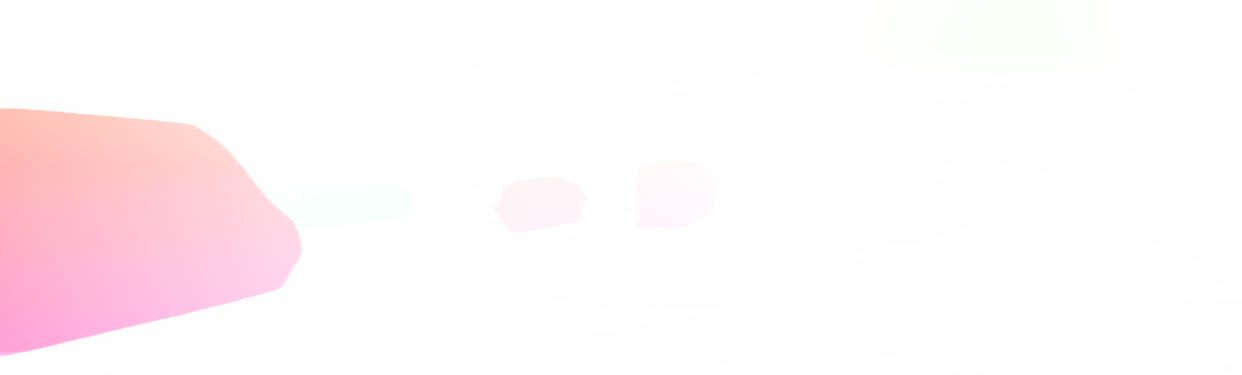} &
        \shiftfigure \includegraphics[width = \Figwidth]{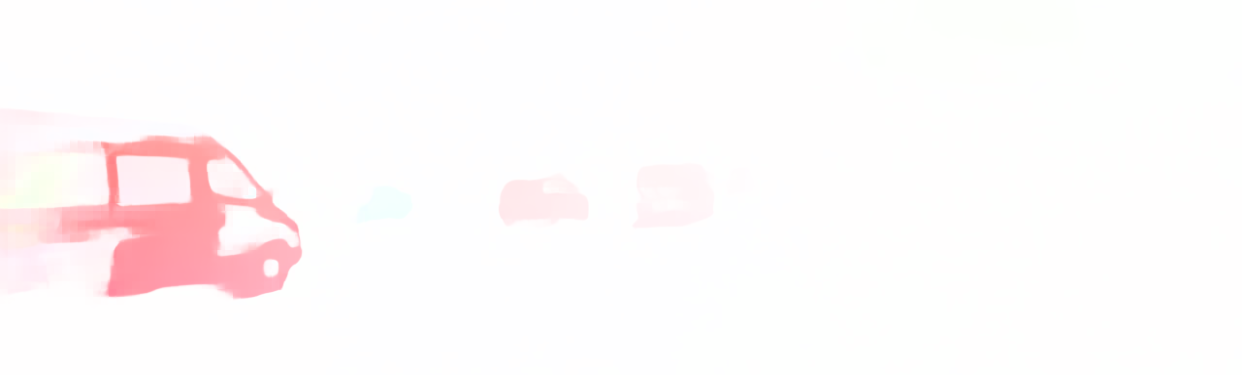}   \\ 
        \shiftfigure  Frame 000094 of KITTI training & \shiftfigure Ground truth & \shiftfigure  \ourmethod{} & \shiftfigure  FlyingChairs \\  
   
        \shiftfigure \includegraphics[width = \Figwidth]{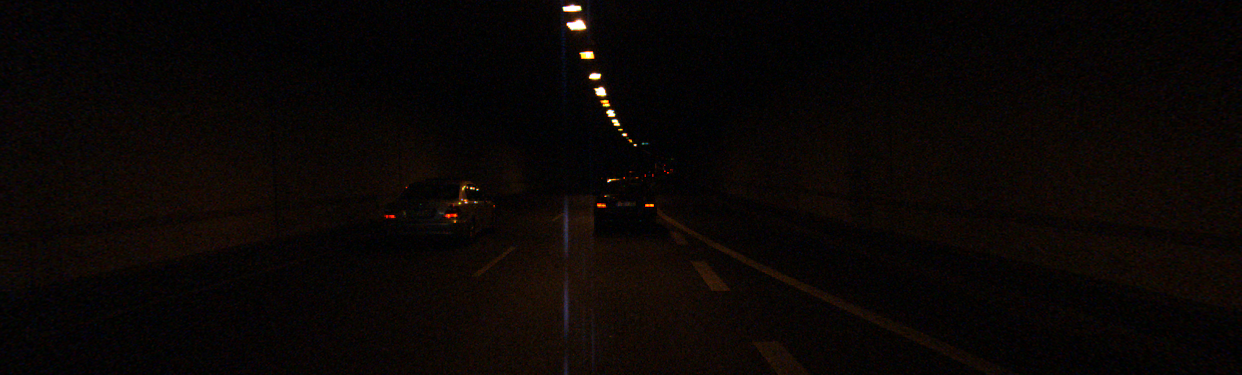} &
        \shiftfigure \includegraphics[width = \Figwidth]{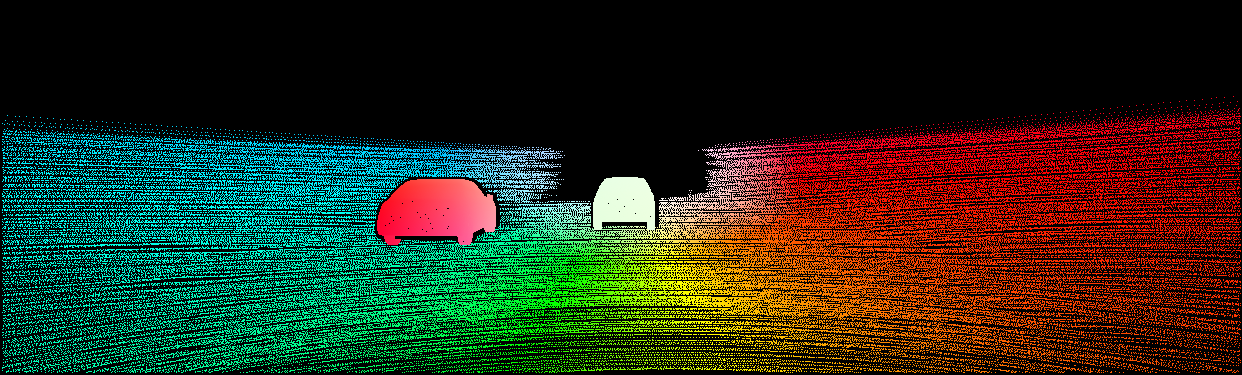}  &
        \shiftfigure \includegraphics[width = \Figwidth]{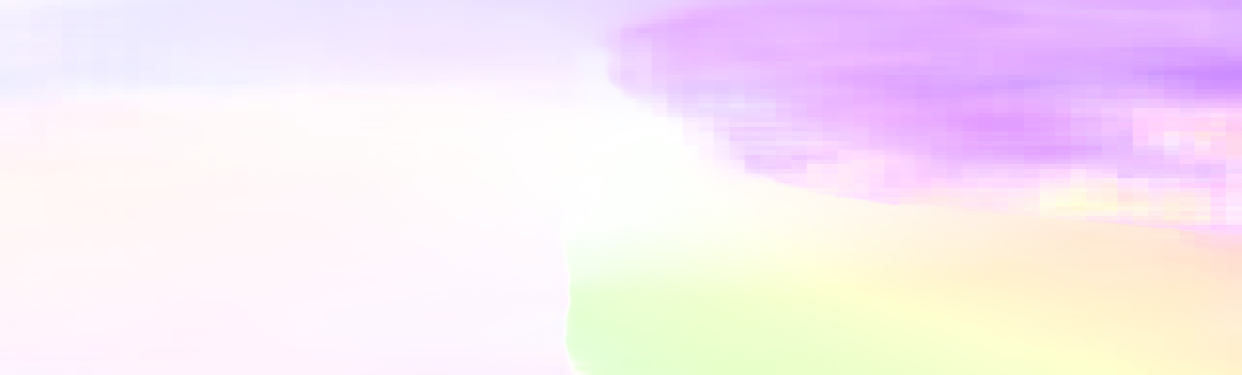} &
        \shiftfigure \includegraphics[width = \Figwidth]{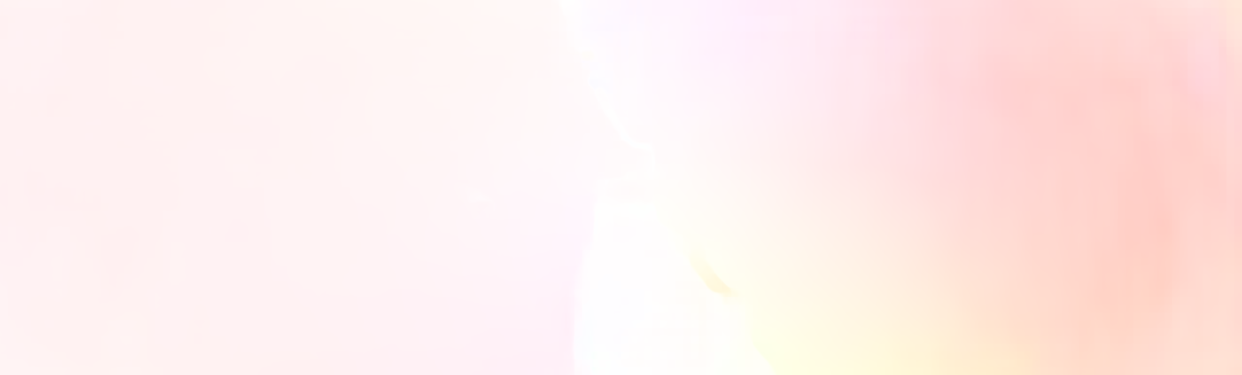}   \\ 
        \shiftfigure  Frame 000104 of KITTI training & \shiftfigure Ground truth & \shiftfigure  \ourmethod{} & \shiftfigure  FlyingChairs \\    
        \end{tabular}
        \vspace{3pt}
		\caption{\textbf{Visual comparison.} 
		Row 1: RAFT trained on \ourmethod{} works better for frames with strong motion blur. 
		Row 2: RAFT trained on \ourmethod{} can capture the car structure. 
		Row 3: low levels of light cause both methods to struggle. 
		}	
        \label{fig:result}
        \vspace{-20pt}
    \end{center}
\end{figure*}

\paragraph{Pre-training results}
We pre-trained RAFT and PWC-Net from scratch using different datasets. The hyperparameters for \ourmethod{} have been learned for RAFT. 
As summarized in Table~\ref{tab:scratch}, models trained from scratch using \ourmethod{} are comparable to or more accurate than models trained on FlyingChairs or FlyingChairs~$\!\rightarrow\!$~FlyingThings3D. 
As shown in \fig~\ref{fig:result}, RAFT trained on \ourmethod{} can successfully recover blurry objects under large motion in the final pass of Sintel.
Table~\ref{tab:detail:comp} summarizes the errors for regions with different motion magnitude.  RAFT trained on \ourmethod{} performs better  than RAFT  trained on FlyingChairs, especially in  regions with large motion. \newedit{Using end-point error (epe) as the learning metric results in better accuracy in regions with large motion than using angular error (ae).}

\begin{table}[t!]
\begin{center}
\footnotesize
\begin{tabular}{ll lcccc } 
\multicolumn{2}{c}{GT range} & $<$ 1 & [1,10] & (10,20] & (20,30] & $>$ 30\\ 
 \hline
\multirow{3}{*}{AEPE} & Chairs & {0.43} &0.89 &3.13 &5.63 &19.61 \\
 & \ourmethod{}-epe & {0.35} & {0.65} & {1.87} & {3.36} & \textbf{15.08} \\
& \ourmethod{}-ae & \textbf{0.31} & \textbf{0.63} & \textbf{1.86} & \textbf{3.24} &16.00 \\ \hline
\multirow{3}{*}{AAE} & Chairs & {10.86} & 6.94 & 6.63 & 9.61 & 13.37	\\ 
& \ourmethod{}-epe &  {10.88} & {5.41} & {4.95} & {6.19} & \textbf{10.35} \\ 
& \ourmethod{}-ae & \textbf{9.77} & \textbf{5.11} & \textbf{4.85} & \textbf{5.96} & {10.68} \\
\end{tabular}
\vspace{5pt}
\caption{\textbf{Results in different motion ranges}. RAFT trained by \ourmethod{} tends to perform better for medium to large motion than RAFT trained by FlyingChairs.  Using end-point error (epe) as the learning metric results in more accurate large motion than using angular error (ae).}
\label{tab:detail:comp}
\vspace{-5pt}
\end{center}
\end{table}

\paragraph{Generalization across datasets}
We further compared with the recent {DSMNet}~\cite{zhang2020domain} method that aims at narrowing down  domain gaps. 
{DSMNet} reported an F-all score of 11.2\% in \emph{non-occlusion} regions on the KITTI 2015 \emph{training} set for a modified PWC-Net trained on FlyingThings3D and Sintel. 
The  F-all scores by RAFT and PWC-Net trained on AutoFlow that has been optimized for Sintel.final are  8.7\% and 11.0\%, respectively, suggesting that \ourmethod{} generalizes well across datasets.

\paragraph{Improving PWC-Net}
We modified the pre-training procedure of PWC-Net~\cite{sun2018pwc} using the one-cycle learning rate schedule and gradient clipping from RAFT~\cite{teed2020raft},  as summarized in Table~\ref{tab:pwcnet}.  Applying gradient clipping not only improves accuracy but also makes training more stable: two out of eight runs diverged without gradient clipping.
 
\begin{table}[h!]
\begin{center}
\small
\begin{tabular}{ cc|ccc } 
\multirow{2}{*}{Learning rate}& \multirow{2}{*}{Gradient clipping} & \multicolumn{2}{c}{Sintel} & \multirow{2}{*}{KITTI} \\
  &    & clean & final & \\ \hline
Piecewise & \xmark & 2.64 &3.44 & 7.26\\ 
Piecewise & \cmark & 2.40	&3.11 & 6.26\\
One-cycle  & \cmark & \textbf{\pwcsintelclean} &	\textbf{\pwcsintelfinal} & \textbf{\pwckitti}  \\
\end{tabular}
\caption{ \textbf{Improvements on pre-training PWC-Net}. Both one-cycle learning rate schedule and gradient clipping help.}
\label{tab:pwcnet}
\end{center}
\end{table}

\paragraph{Fine-tuning results}

We followed the TF-RAFT procedure to fine-tune the model pre-trained by AutoFlow and denoted the method as RAFT-A.
We applied the same fine-tuned model to Sintel and KITTI, as summarized in Table~\ref{tab:benchmark}. RAFT-A is more accurate than TF-RAFT on the more challenging Sintel.final and KITTI benchmarks, demonstrating the benefits of pre-training on AutoFlow.

\begin{table}[t]
\begin{center}
\begin{adjustbox}{width=0.48\textwidth}
\small
\begin{tabular}{llccc} 
Method & Dataset schedule & S.clean & S.final & KITTI \\ \hline
FlowNet2 & C$\rightarrow$T$\rightarrow$S & 3.96 & 6.02 &11.48\%$^*$ \\
PWC-Net & C$\rightarrow$T$\rightarrow$S  &  3.86 & 5.13  & 9.60\%$^*$\\
VCN & C$\rightarrow$T$\rightarrow$SKHTC  & 2.81 & 4.40 & 6.30\%$^*$ \\ 
RAFT~\cite{teed2020raft} &  C$\rightarrow$T$\rightarrow$SKHT/K & {1.94} & {3.18}& {5.10}\%$^*$ \\ 
\hline
TF-RAFT~\cite{tf-raft} & C$\rightarrow$T$\rightarrow$SKHTV & \textbf{1.84} & 3.32 &  5.56\%\\
RAFT-A & A$\rightarrow$SKHTV & 2.01 & \textbf{3.14} &  \textbf{4.78}\%\\
\end{tabular}
\end{adjustbox}
\vspace{5pt}
\caption{\textbf{Results on public benchmarks} (AEPE for Sintel and Fl-all for KITTI).
A, C, H, K, S, and T stand for \ourmethod{}, FlyingChairs, HD1K, KITTI, Sintel,  and FlyingThings3D, respectively.
$^*$indicates where weights for KITTI differ from those for Sintel.  
}
\label{tab:benchmark}
\vspace{-5pt}
\end{center}
\end{table}

\subsection{Ablation Study}
To further analyze \ourmethod{}, we performed a series of ablation studies designed to determine how different design choices affect  performance.  Since it is computationally expensive to learn all the hyperparameters for each setup, we fixed the learned hyperparameters unless explicitly specified. 
For each experiment, we ran 8 independent trials, and Table~\ref{tab:ablation} summarizes the most accurate one for each setup.

\paragraph{Motion blur and fog}
Removing the motion blur effect leads to a significant drop in performance on the final pass of Sintel and KITTI, despite the rough approximation used for simulating the motion blur.  Gaussian and box filters have similar results. Removing the fog effect also results in a moderate performance drop in the final pass of Sintel and KITTI. Neither motion blur nor fog effects have a significant effect on the clean pass of Sintel. 

\paragraph{Appearance}

We tested three different image sources for the appearance image of each layer: Davis, OpenImages~\cite{OpenImages}, and Sintel (\cf Table~\ref{tab:ablation}). 
Neither OpenImages nor Sintel achieves better results than Davis. 
By default, we downsample Davis images to $1280\!\times\!720$ (720p)  resolution as appearance images for each layer. Downsampling to $960\!\times\!540$ (540p) has similar results while $1920 \!\times\! 1080$ (1080p) has degraded performance, likely because the hyperparameters have been learned for the 720p resolution. 
\paragraph{Foreground object masks}

We tested three versions of masks for the foreground objects: random polygons with sharp edges, random polygons with smooth edges (default), and instance segmentation from the OpenImage~\cite{OpenImages} dataset. 
Polygons with smooth edges perform consistently better than those with sharp edges.
We also experimented with instance segmentation from the OpenImage dataset due to its diverse set of segmentation masks, but there we only observed a small improvement on Sintel.clean.

\begin{table}[h!]
\begin{center}
\small
\begin{tabular}{ lcccc } 
\multirow{2}{*}{Experiment} &  & \multicolumn{2}{c}{Sintel} & \multirow{2}{*}{KITTI} \\
  &    & clean & final & \\ \hline
\multirow{2}{*}{ Fog } & \underline{On} & \raftsintelcleanbase &	\raftsintelfinalbase &	\raftkittibase \\ 
 & Off  & 2.07	& 3.11 & 4.92 \\ \hline  
\multirow{3}{*}{ Motion blur } & \underline{Box} & \raftsintelcleanbase &	\raftsintelfinalbase &	\raftkittibase \\ 
& Gaussian & 2.17 &	2.75 & 4.71\\ 
& Off & 2.10 &	3.77 & 5.68 \\ \hline 
 \multirow{3}{*}{Appearance} & \underline{Davis} & \raftsintelcleanbase &	\raftsintelfinalbase &	\raftkittibase \\ 
 & OpenImages & 2.20 &	2.85 & 4.83\\ 
 & Sintel-540p &2.17	&2.88	&4.75 \\ \hline 
\multirow{3}{*}{Resolution} & 540p  &2.06	&2.88	&4.87\\ 
& \underline{720}p & \raftsintelcleanbase &	\raftsintelfinalbase &	\raftkittibase \\ 
& 1080p &2.33	&2.85	&5.09\\ \hline
\multirow{3}{*}{Object mask}  & Sharp & 2.09	& 2.80 & 4.94 \\ 
  & \underline{Smooth} & \raftsintelcleanbase &	\raftsintelfinalbase &	\raftkittibase \\ 
 & Instance & 1.99 &	2.78	& 4.85 \\\hline
& 0   &5.32	&5.74	&9.02\\
& 1 &2.38	&3.07	&5.22 \\ 
Number of & 2 &2.17	&2.93	&4.99  \\
{foreground} & 3 &2.11	&2.76	&4.64  \\
objects& \underline{4}  & \raftsintelcleanbase &	\raftsintelfinalbase &	\raftkittibase \\ 
& 5 &2.02	&2.87	&4.66  \\
& 6 &2.05	&2.84	&4.57   \\\hline
\multirow{2}{*}{Grid warp } & \underline{On} & \raftsintelcleanbase &	\raftsintelfinalbase &	\raftkittibase \\ 
 & Off & 2.30	& 2.92 &5.26 \\ \hline
\multirow{3}{*}{Training steps} &  50k & 2.42 &	3.27 &	5.81\\
& \underline{200k} &  {\raftsintelcleanbase}  & {\raftsintelfinalbase} & \raftkittibase \\ 
& 800k &  \raftsintelclean &	\raftsintelfinal & {\raftkitti} \\  \hline
\multirow{2}{*}{Target dataset} &  \underline{Sintel.final} &  {\raftsintelcleanbase}  & {\raftsintelfinalbase} & \raftkittibase \\ 
& KITTI & {2.09}		&2.82	& {4.33} \\  \hline
 \multirow{4}{*}{Augmentation} 
& All &2.22	& 2.87 & 4.87  \\ 
& \underline{RandAugment} & \raftsintelcleanbase &	\raftsintelfinalbase &	\raftkittibase \\ 
& No spatial &2.78&	3.37&	5.22\\ 
& No color  &2.24&	2.92	&14.06 \\ 
\hline
\end{tabular}
\vspace{5pt}
\caption{\textbf{Ablation study}. Baseline options are underlined.}
\label{tab:ablation}
\vspace{-5pt}
\end{center}
\end{table}

\paragraph{Number of foreground objects}

The number of foreground objects determines the complexity of a scene. 
Using only a background layer, \ie, 0 foreground object, results in large errors. 
Adding one foreground object significantly improves the performance. 
Using three or four foreground objects tends to work best, while more than four foreground objects bring no further gain. 

\paragraph{Motion model}
Removing the bilinear grid warping results in a performance degradation on both Sintel and KITTI, suggesting that more complex and flexible motion than parametric motion is critical.

\paragraph{Number of training steps}
We learned the hyperparameters using 200k training steps for RAFT. With the same hyperparameters, running more iterations to train RAFT, such as 800k, results in moderate gains on both Sintel and KITTI.

\paragraph{Target datasets}
\ourmethod{} directly optimizes the performance on a target dataset.  To test how well \ourmethod{} generalizes, we learned hyperparameters for Sintel.final and KITTI separately and found  that the generalization gap is small. 
It is likely that the rendering pipeline and the small number of hyperparameters act as a form of regularization, which helps generalization.

\paragraph{Data augmentation}
RandAugment leads to moderate improvement over applying the same augmentation at every training step, likely because RandAugment increases the diversity of training data. \newedit{Turning off spatial augmentation results in a moderate drop in accuracy on both KITTI and Sintel. Turning off color augmentation results in severe performance degradation on KITTI, likely because KITTI data includes more lighting changes.}

\begin{figure}
    \begin{center}
        \newcommand{\figwidth}{0.035\linewidth}
        \newcommand{\Figwidth}{\linewidth}
        \newcommand{\Figheight}{0.4\linewidth}
        \newcommand{\shiftfigure}{\hspace{-1mm}}
        \begin{tabular}{cc}
        \includegraphics[width = \Figwidth]{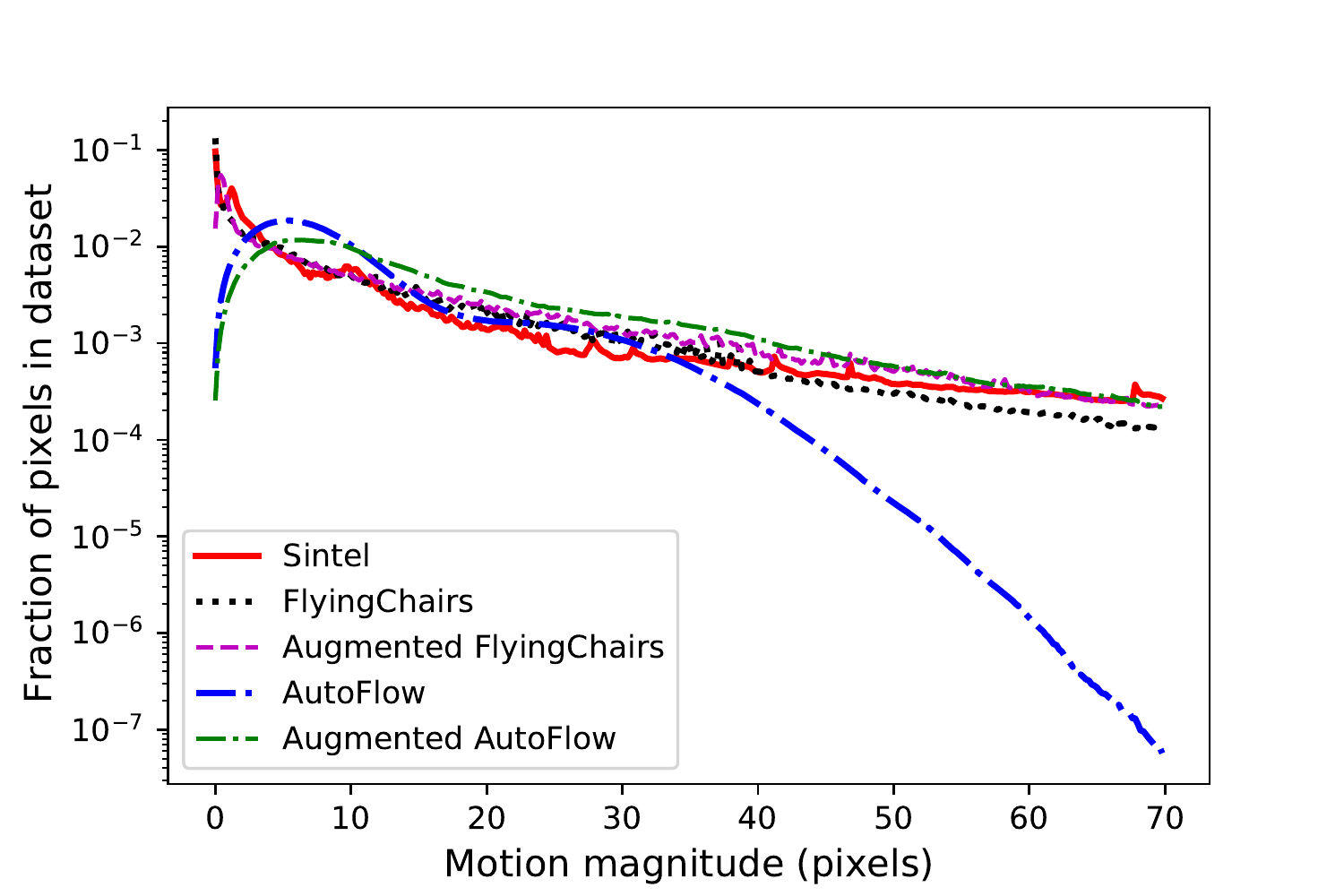} &
        \end{tabular}
		\caption{\textbf{Histogram of motion magnitude for different datasets}. The augmented \ourmethod{} concentrates more on  middle to large-range motion than Sintel, likely because the small motion contributes little to the overall error.
		}		
        \label{fig:motion:stat}
        \vspace{-5pt}
    \end{center}
\end{figure}

\paragraph{Motion statistics}
We compared the statistics of motion magnitude for different datasets  in Figure~\ref{fig:motion:stat}. 
The motion statistics of \ourmethod{} differ from those of Sintel and FlyingChairs. \ourmethod{} has little small motion, concentrates mainly in the middle-range motion, and does not exhibit an exponential falloff. We further analyzed the augmented data, as it is used to train models. 
The augmented \ourmethod{} also has little small motion and concentrates in the middle to high-range motion,  probably because  tiny motion matters little in the overall learning metric.

\begin{figure}[t]
	\begin{center}
		\newcommand{\figwidth}{0.035\linewidth}
		\newcommand{\Figwidth}{\linewidth}
		\newcommand{\Figheight}{0.4\linewidth}
		\newcommand{\shiftfigure}{\hspace{-1mm}}
		\includegraphics[width = \Figwidth]{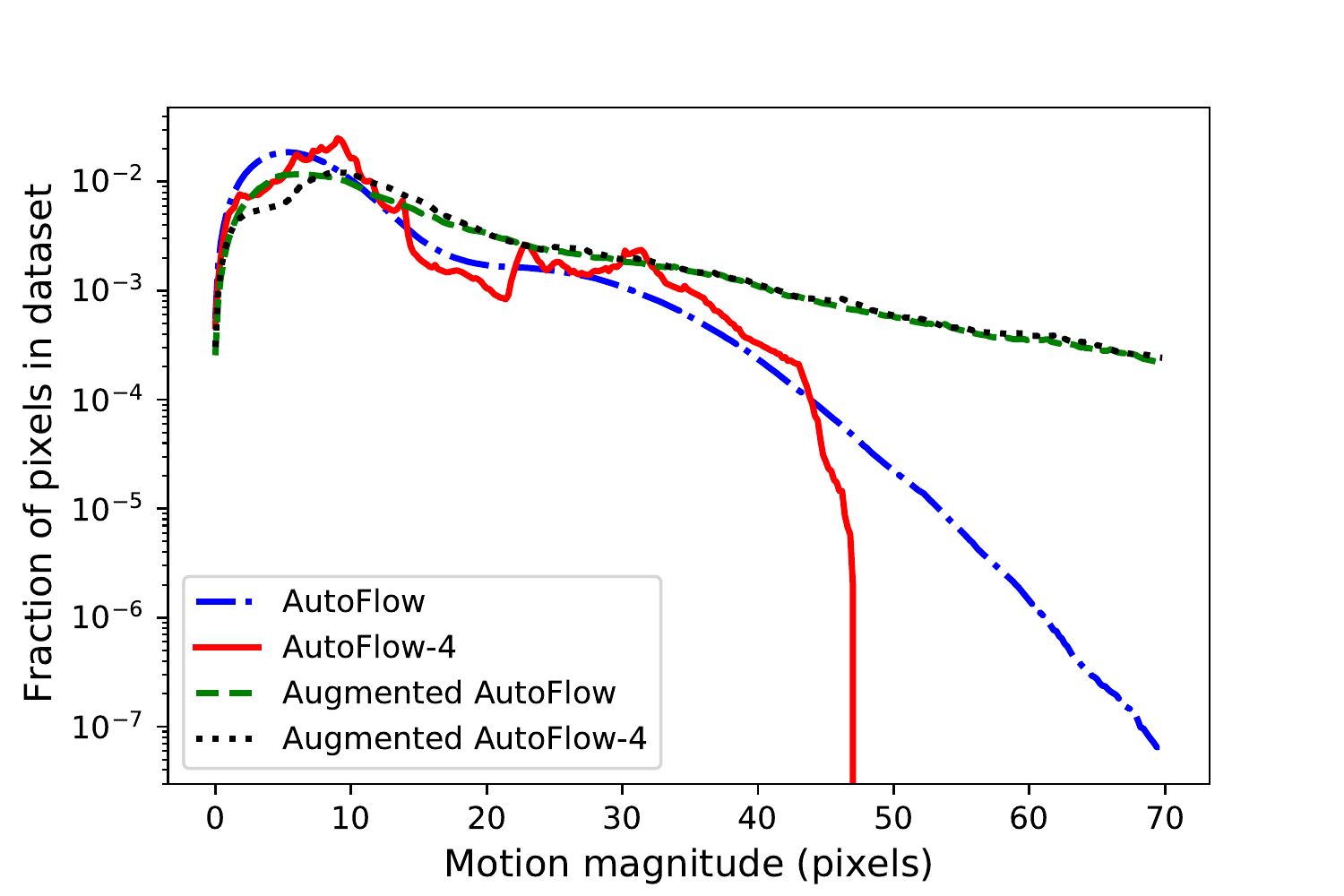}
		\caption{ \textbf{Histogram of motion magnitude for \ourmethod{}}. While statistics differ between the \minautoflownumber{}-example \ourmethod{} and  full \ourmethod{}, they are close for the augmented data.
		}\vspace{-15pt}		
		\label{fig:hist:augment}
	\end{center}
\end{figure}

\paragraph{Number of pre-training examples}
With the  hyperparameters learned, we can render different numbers of pre-training examples, as shown in \fig~\ref{fig:teaser}.
The training of both RAFT and PWC-Net converge using one pre-training example with data augmentation, and more examples lead to better results.  Four \ourmethod{} examples result in lower errors on Sintel.final for RAFT than  22,872 FlyingChairs examples. In this low-data regime, data augmentation plays a key role. Without spatial augmentation, the AEPE by RAFT trained on  \minautoflownumber{} \ourmethod{} examples drops from 3.57 to 7.66,  more severe than the drop from \raftsintelfinalbase{} to 3.37 in Table~\ref{tab:ablation}. Further, as shown in \fig~\ref{fig:hist:augment}, although the statistics of  \minautoflownumber{} \ourmethod{} examples differ significantly from those of the full \ourmethod{}, they are similar for augmented data.

\paragraph{Discussions}
While \ourmethod{} empirically works better than FlyingChairs/FlyingThings3D, we should note that the comparisons are not strictly fair because of differences in implementations and hyperparameters. 
Although comparing motion statistics reveals some interesting properties, learning hyperparameters for a 3D rendering pipeline in the same setup would help identify key design choices. 

\section{Conclusions}

We have introduced \ourmethod, a simple and effective method to learn pre-training data for optical flow. 
AutoFlow uses 2D rendering but achieves results comparable to or better than those obtained by FlyingChairs and FlyingThings3D that have been generated using 3D models. In particular, using as few as \minautoflownumber{} \ourmethod{}  examples with augmentation results in more accurate results on Sintel.final for RAFT than \chairnumber{} FlyingChairs examples with augmentation. \ourmethod{} also significantly improves PWC-Net, even on par with  RAFT. 
We hope that our approach will provide another option for pre-training optical flow and enable further progress and innovation in this direction.

{\small \paragraph{Acknowledgements} We would like to thank Shuyang Cheng, Ekin Dogus Cubuk, Alex Dosovitskiy, Rico Jonschkowski, David Kao, Ang Li, Aaron Sarna, Austin Stone,  and Barret Zoph for helpful discussions and support. 
}

{\small
\bibliographystyle{ieee_fullname}
\bibliography{autoflow_arxiv}
}

\newpage
\appendix

\renewcommand\thefigure{\thesection.\arabic{figure}}    
\setcounter{figure}{0}

\section{Rendering Hyperparameters}
During training, we tune a number of hyperparameters that dictate data generation, including the shape, size, and position of masks, the complexity and magnitude of motion, and the visual effects. Respective values are uniformly sampled from the specified ranges, and the ranges are hyperparameters to learn.  $\uni$ denotes a random number uniformly sampled from $[-1$, $1]$.

\paragraph{Hyperparameters for polygon masks:}
\begin{itemize}
    \item minimum and maximum number of sides
    \item maximum size of the hole's bounding box diagonal, relative to the polygon's
    \item number of polygon subdivisions
\end{itemize}

\paragraph{Hyperparameters for all foreground masks:}
\begin{itemize}
    \item minimum and maximum size of the object's bounding box diagonal, relative to the image diagonal
    \item minimum and maximum object center location, relative to image dimensions
    \item probability of applying mask blur
    \item strength of the mask blur
\end{itemize}

\paragraph{Hyperparameters for motion}
\begin{itemize}
    \item scale strength $p_s$ ($\geq1$), with scale sampled as ${p_s}^{\uni}$ 
    \item rotation strength $p_r$, with rotation angle sampled as $\pi \cdot p_r \cdot \uni$
    \item translation strength $p_t$, with translation in dimension $d$ sampled as $\mathrm{ImageSize}_d \cdot p_t \cdot \uni$
    \item grid strength $p_g$, where each grid vertex offset in dimension $d$ is sampled as $0.5 \cdot \mathrm{CellSize}_d \cdot p_g \cdot \uni$
    \item size of the grid
\end{itemize}

\paragraph{Hyperparameters for visual effects:}
\begin{itemize}
    \item probability of applying motion blur
    \item strength of the motion blur
    \item probability of applying fog
    \item average density of the fog
\end{itemize}

\paragraph{Hyperparameters for RangAugment:}
\begin{itemize}
    \item number of augmentations per iteration
    \item strength level for all the augmentations
\end{itemize}

\section{More Samples}
\begin{figure*}[t]
  \centering
  \setkeys{Gin}{width=0.2\linewidth}
  
  \includegraphics{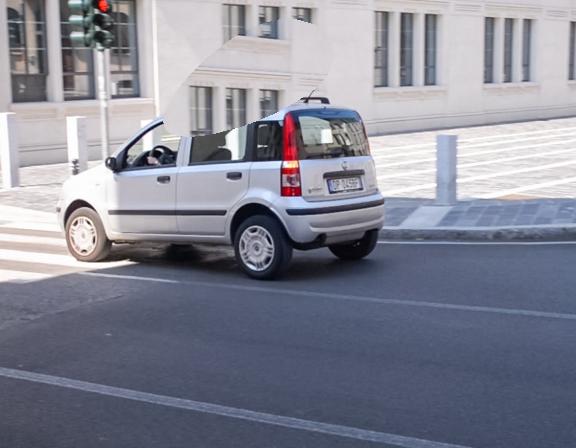}\,%
  \includegraphics{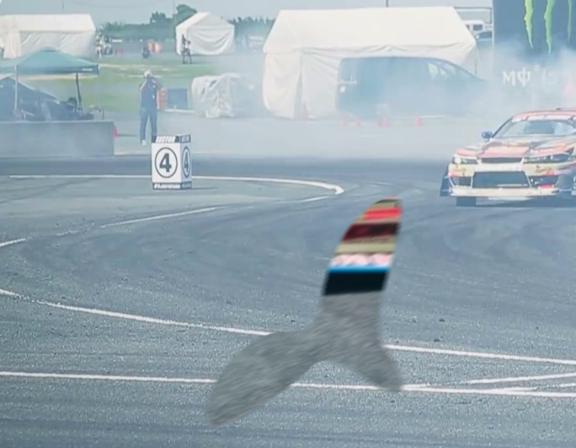}\,%
  \includegraphics{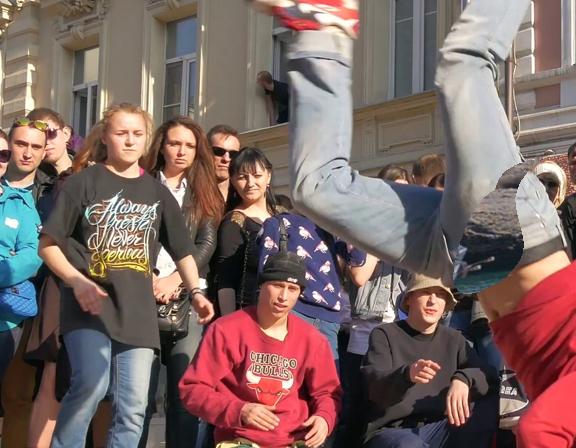}\,%
  \includegraphics{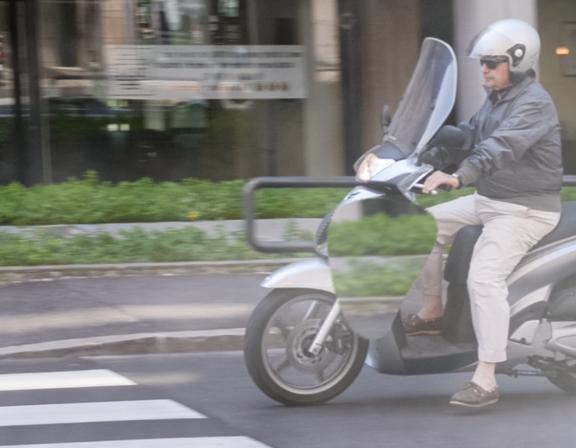}\,%
  \includegraphics{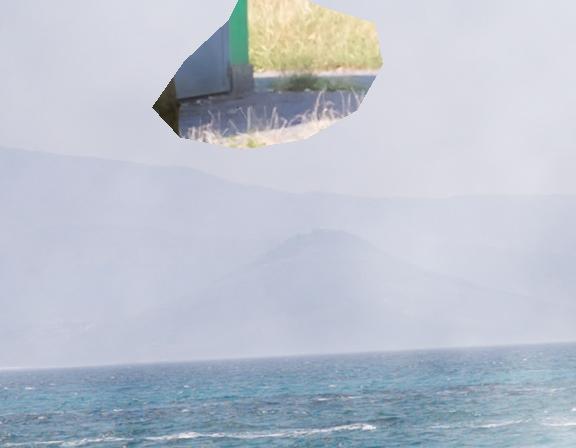}

  \includegraphics{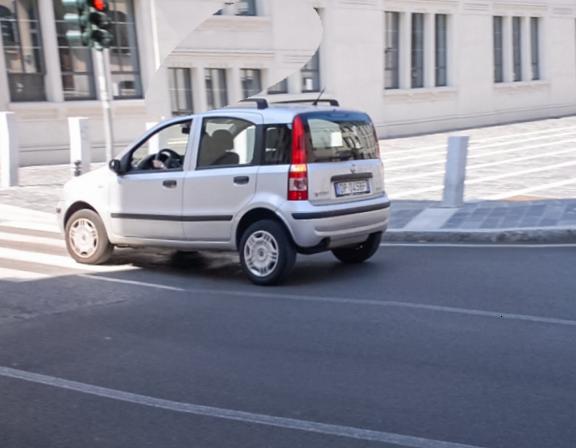}\,%
  \includegraphics{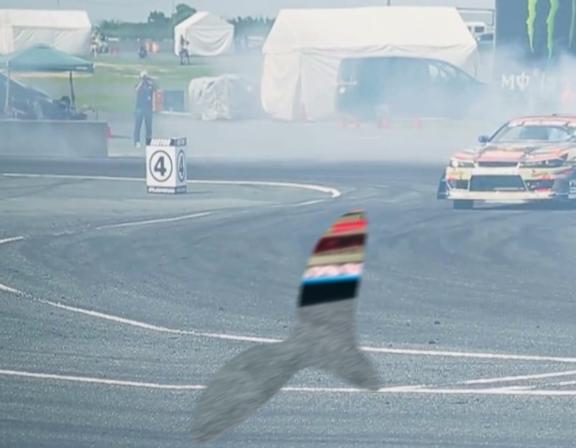}\,%
  \includegraphics{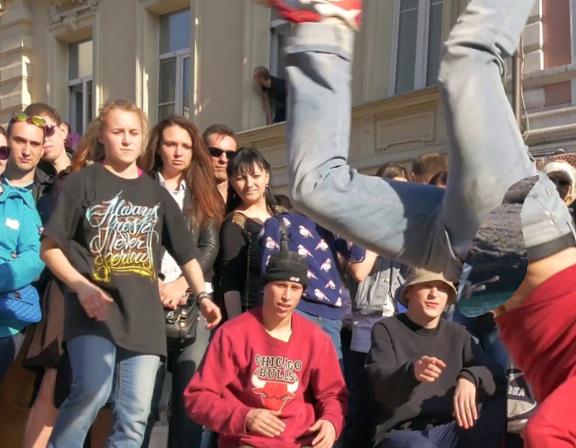}\,%
  \includegraphics{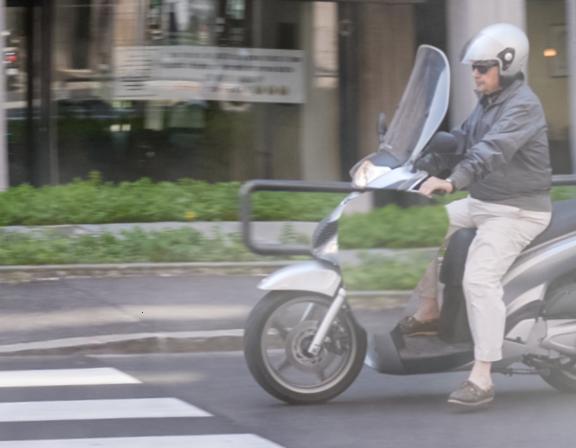}\,%
  \includegraphics{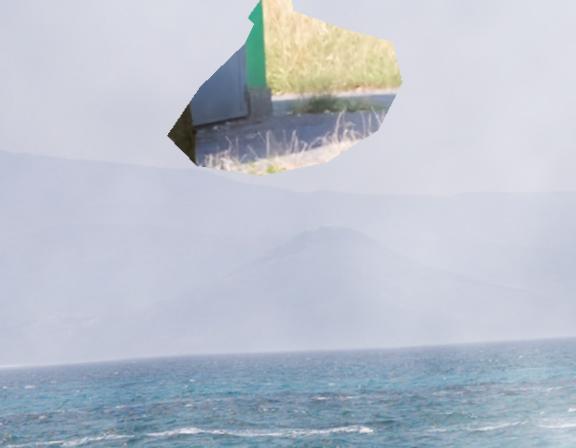}

  \includegraphics{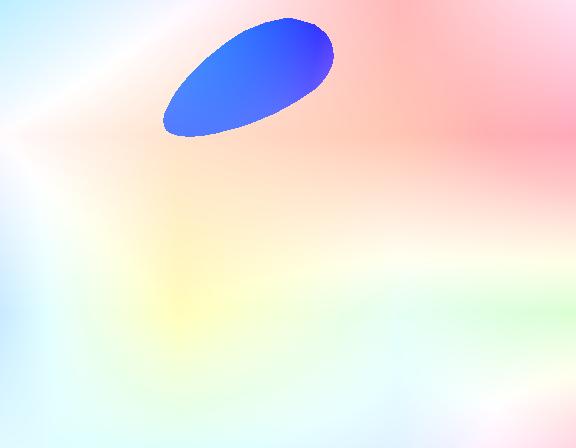}\,%
  \includegraphics{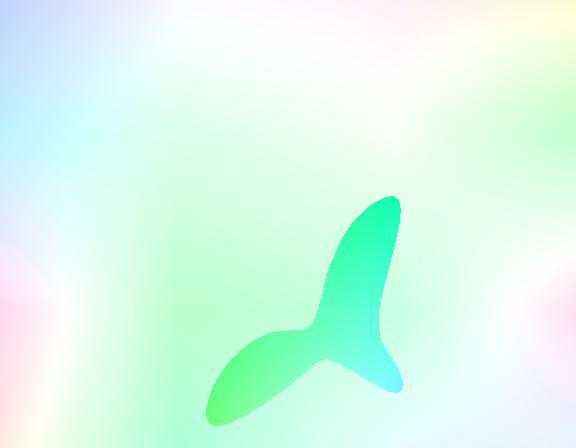}\,%
  \includegraphics{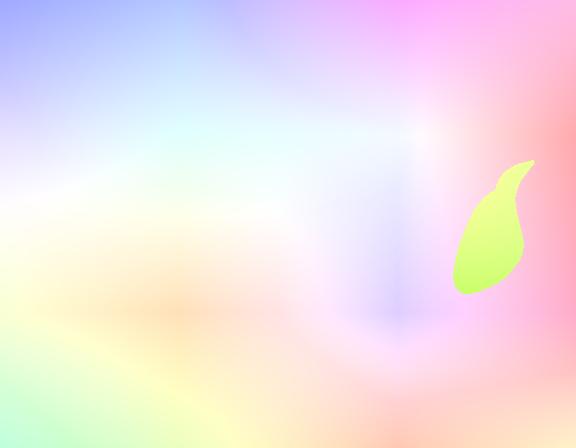}\,%
  \includegraphics{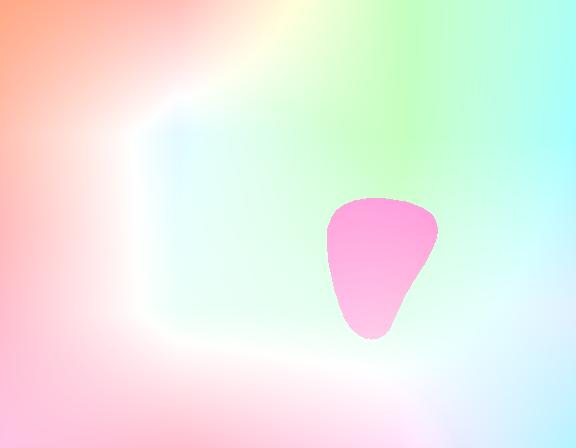}\,%
  \includegraphics{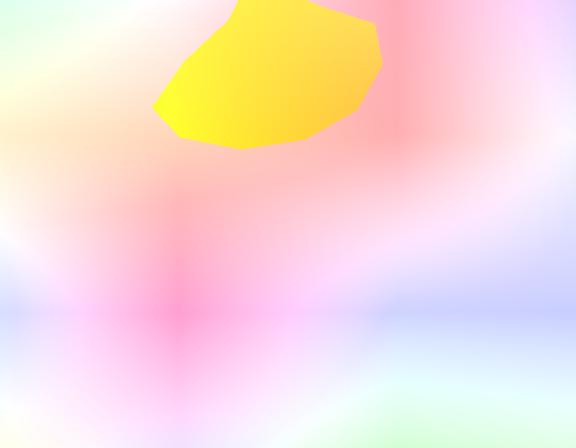}

  \includegraphics{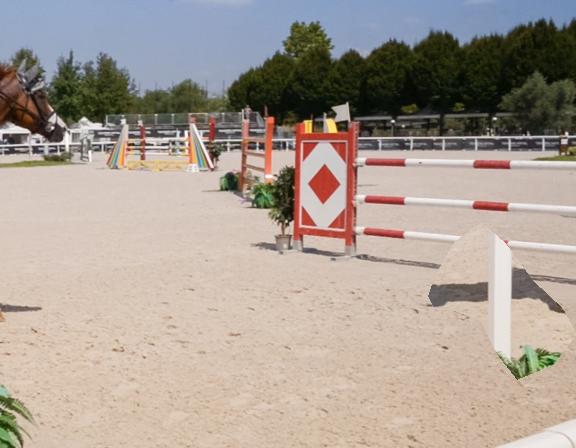}\,%
  \includegraphics{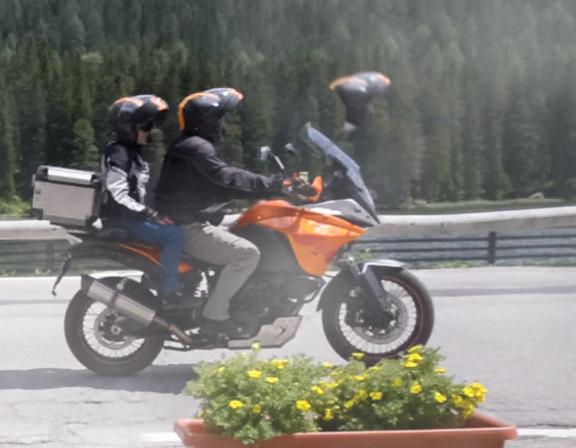}\,%
  \includegraphics{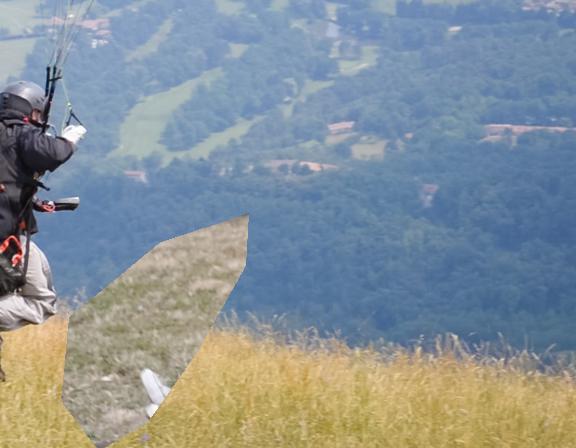}\,%
  \includegraphics{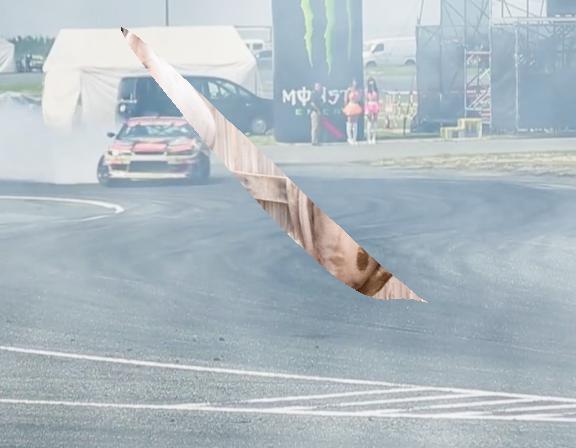}\,%
  \includegraphics{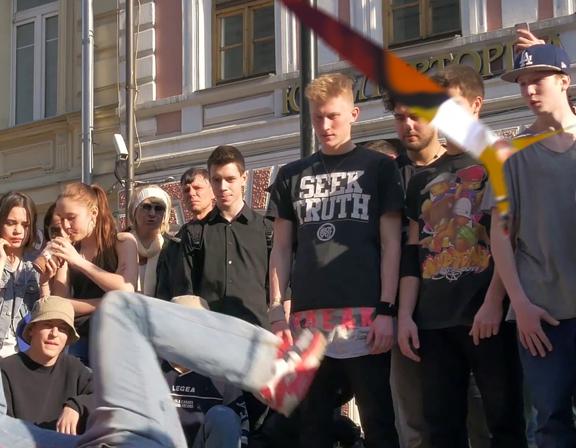}

  \includegraphics{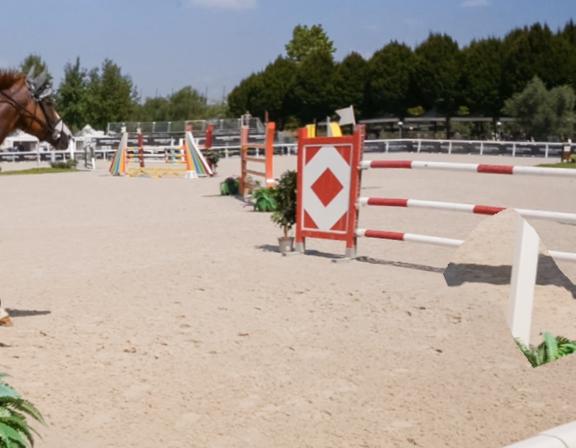}\,%
  \includegraphics{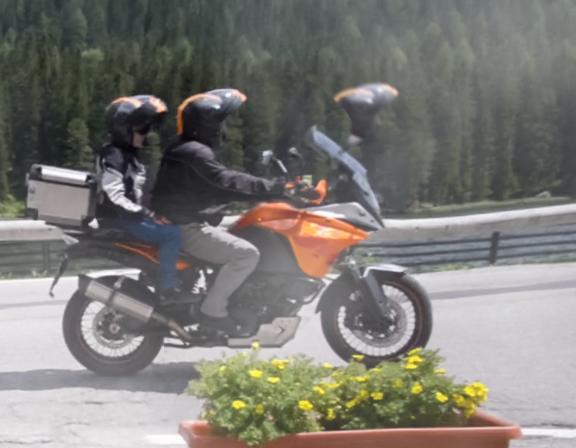}\,%
  \includegraphics{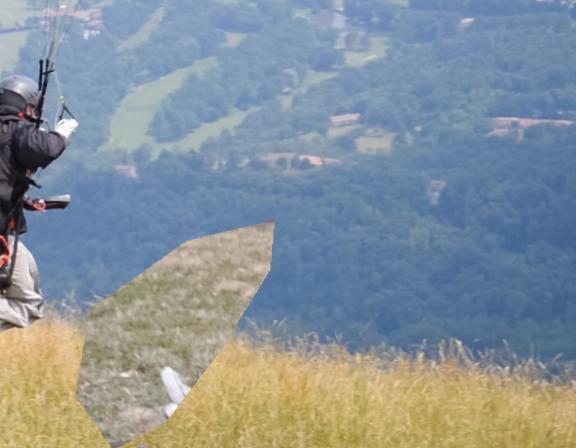}\,%
  \includegraphics{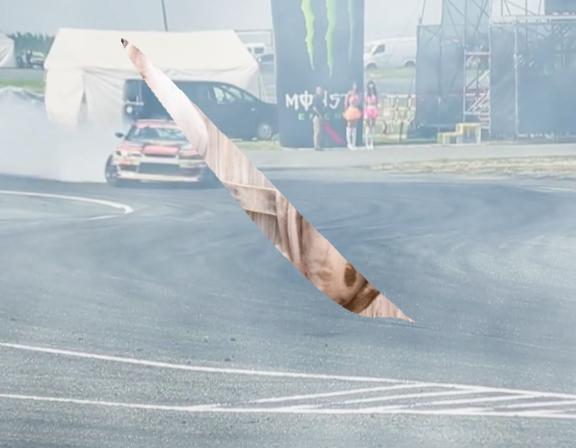}\,%
  \includegraphics{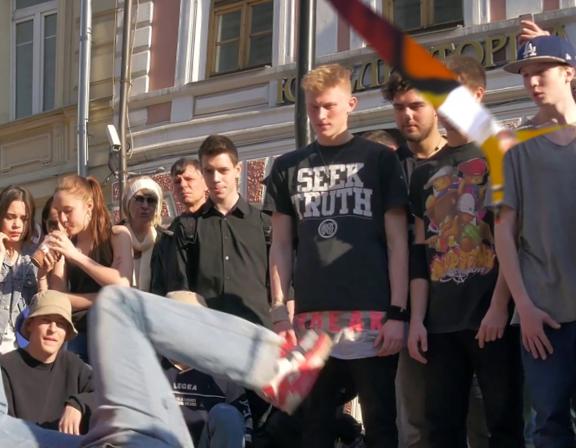}

  \includegraphics{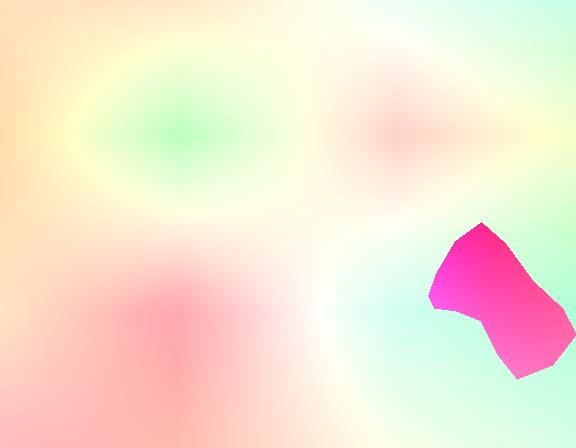}\,%
  \includegraphics{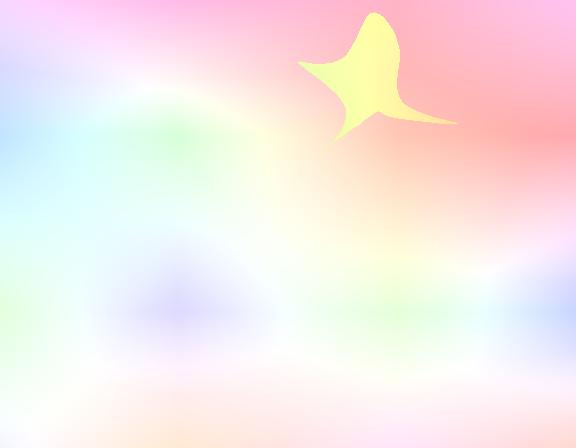}\,%
  \includegraphics{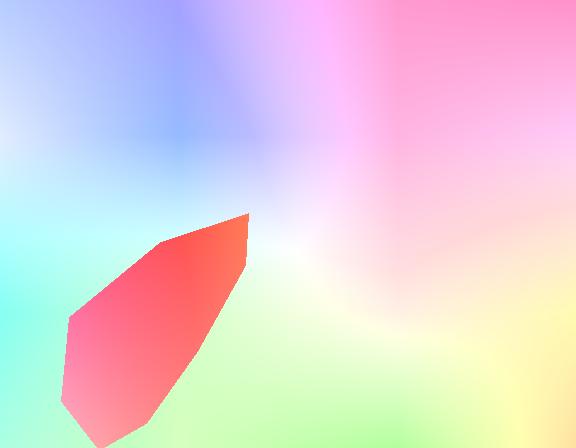}\,%
  \includegraphics{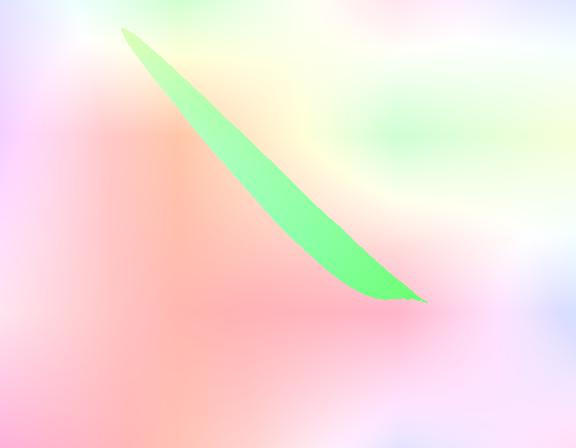}\,%
  \includegraphics{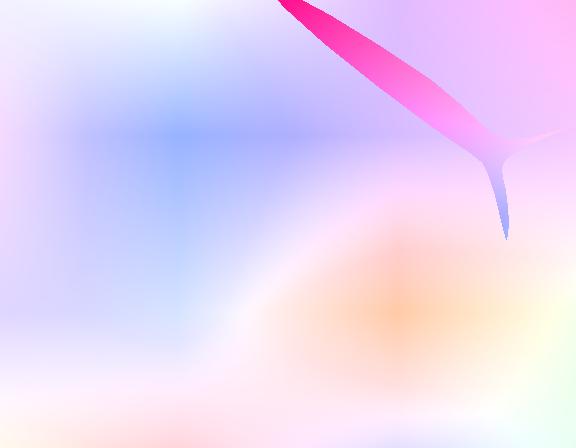}
  
  \caption{\textbf{AutoFlow samples with one foreground object}. First/fourth row: first image; second/fifth row: second image; third/sixth row: visualized optical flow. \textbf{Please go to our webpage \url{autoflow-google.github.io} to see the gif images.}}    
\label{fig:more:samples1}
\end{figure*}

\begin{figure*}[t]
  \centering
  \setkeys{Gin}{width=0.2\linewidth}
  
  \includegraphics{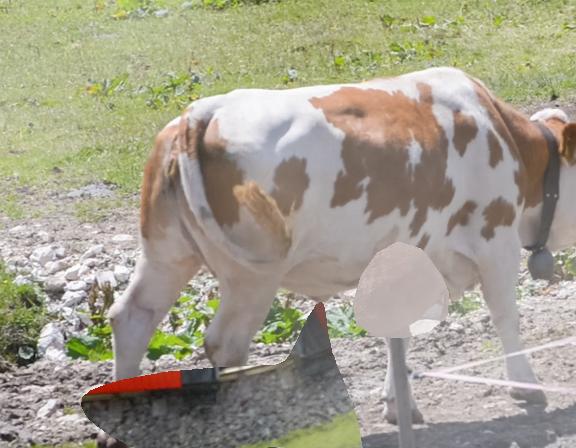}\,%
  \includegraphics{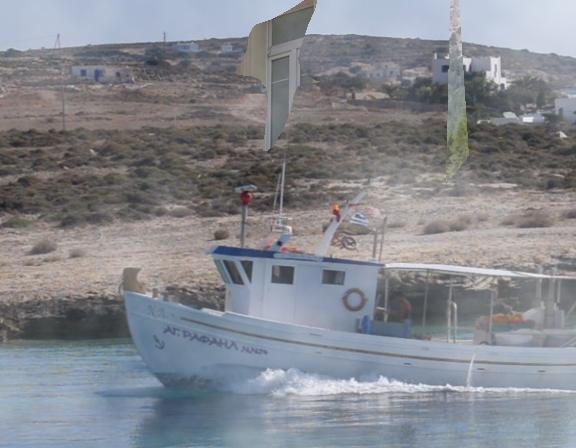}\,%
  \includegraphics{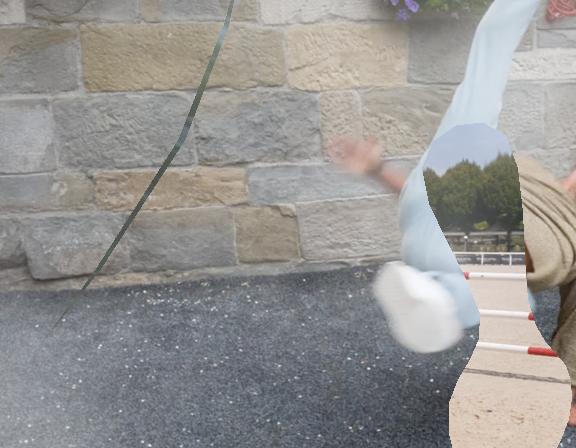}\,%
  \includegraphics{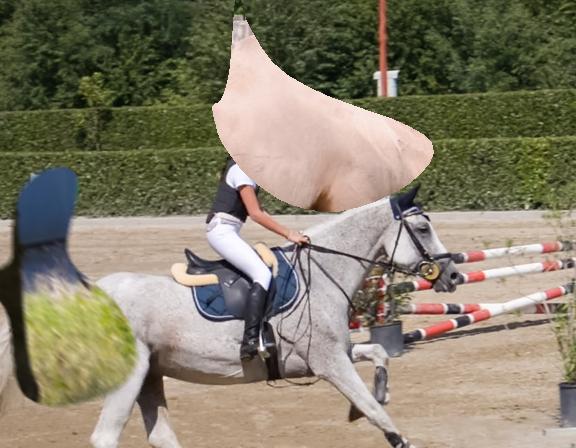}\,%
  \includegraphics{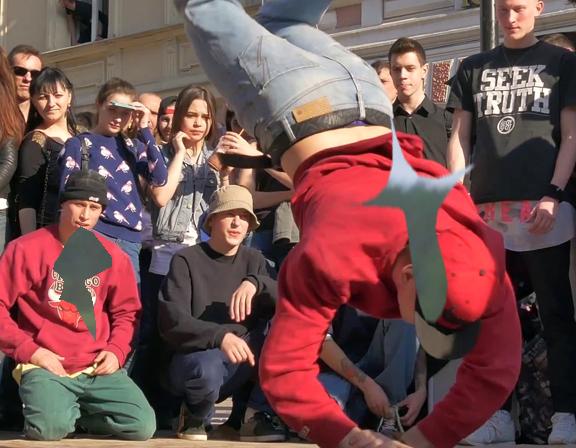}

  \includegraphics{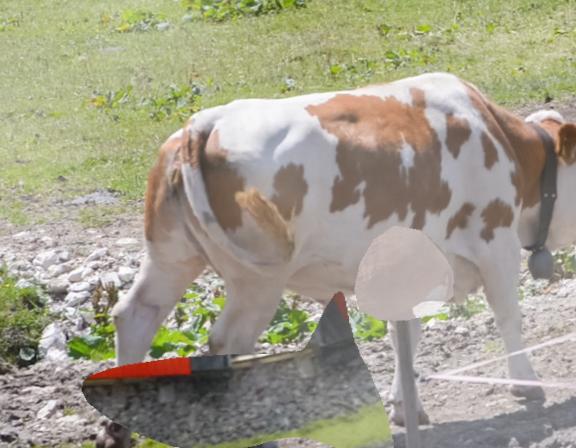}\,%
  \includegraphics{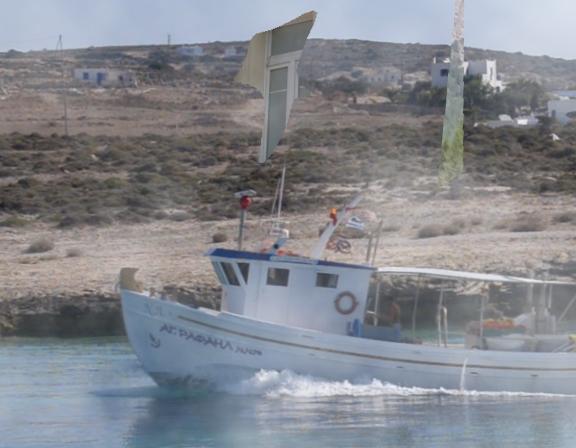}\,%
  \includegraphics{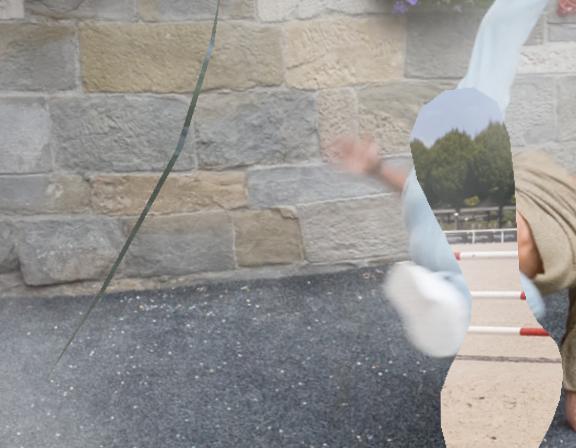}\,%
  \includegraphics{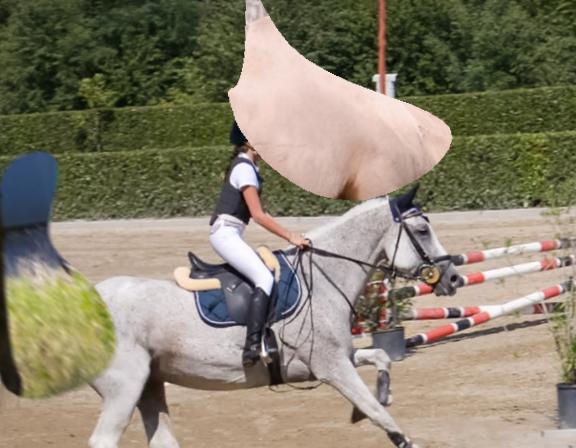}\,%
  \includegraphics{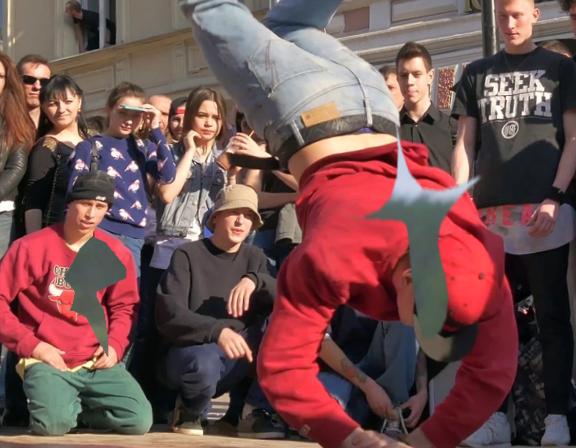}

  \includegraphics{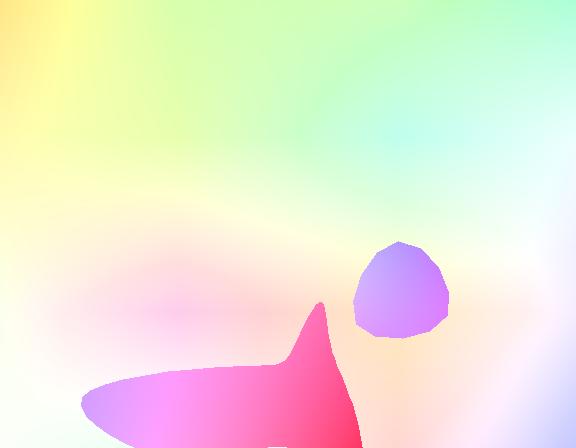}\,%
  \includegraphics{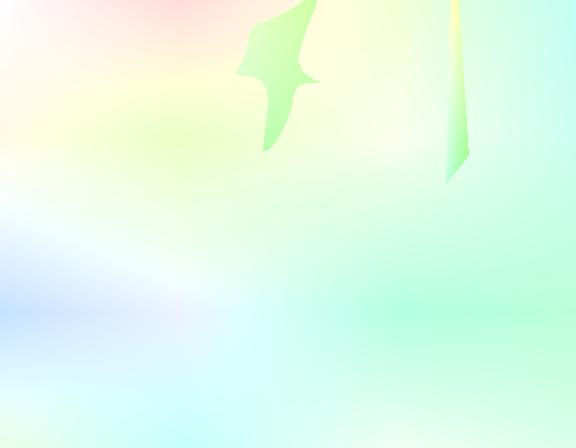}\,%
  \includegraphics{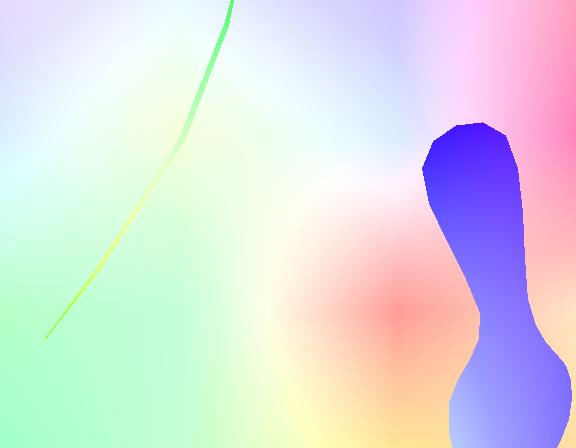}\,%
  \includegraphics{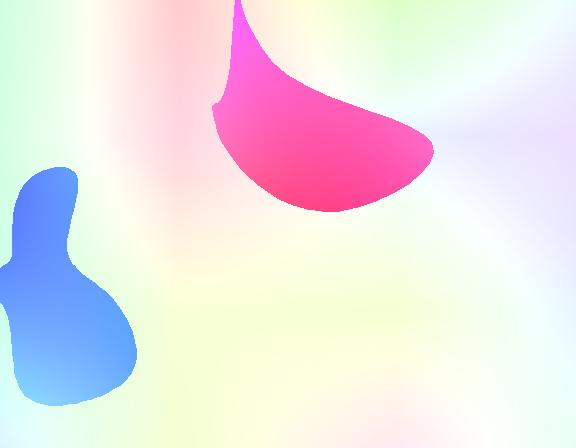}\,%
  \includegraphics{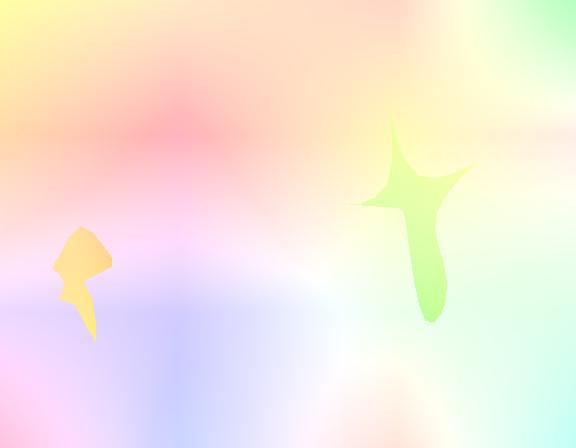}

  \includegraphics{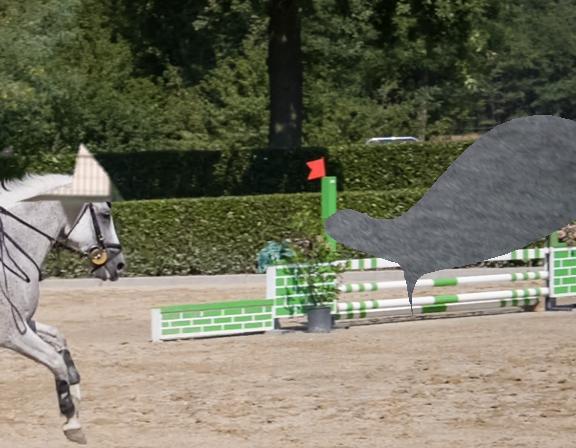}\,%
  \includegraphics{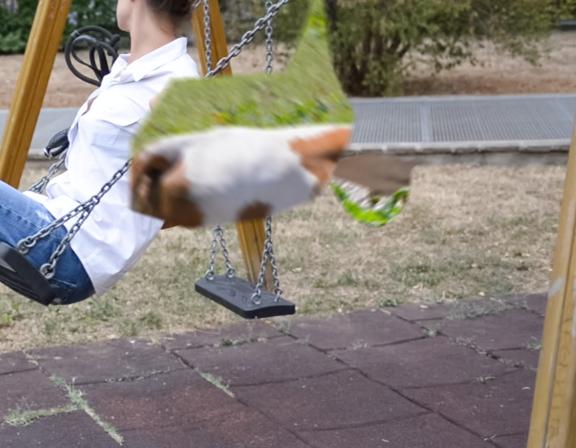}\,%
  \includegraphics{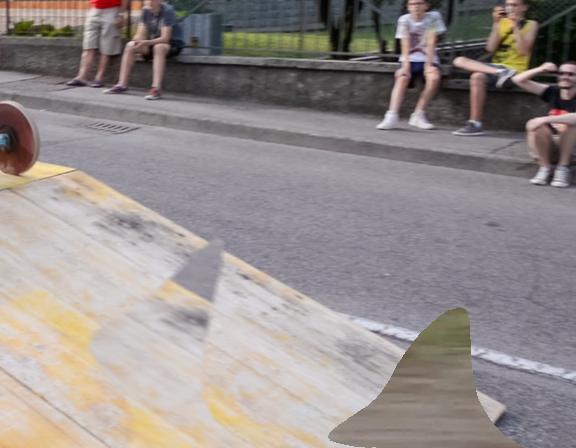}\,%
  \includegraphics{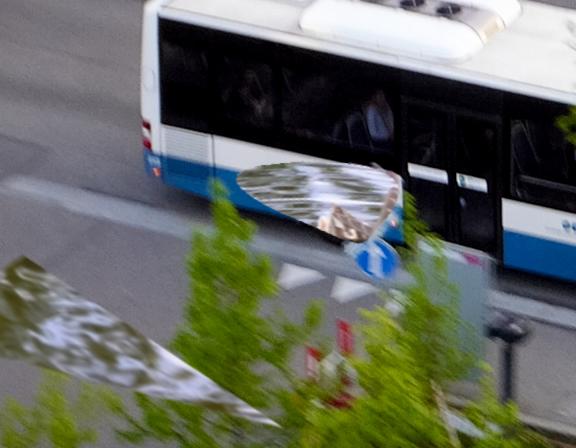}\,%
  \includegraphics{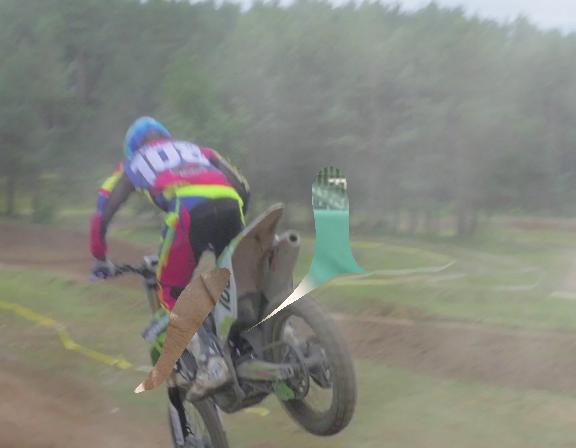}

  \includegraphics{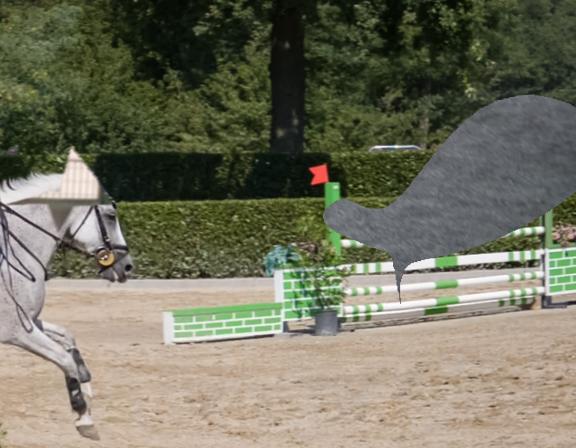}\,%
  \includegraphics{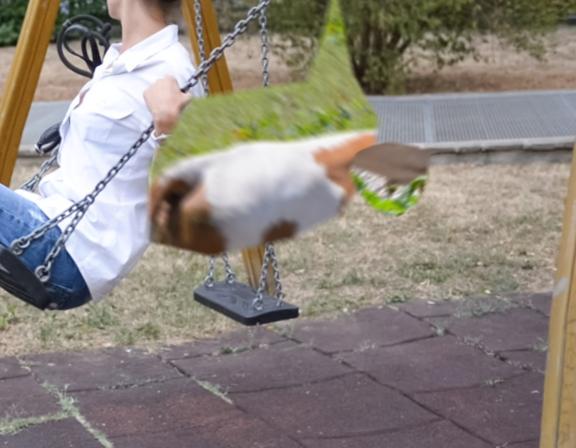}\,%
  \includegraphics{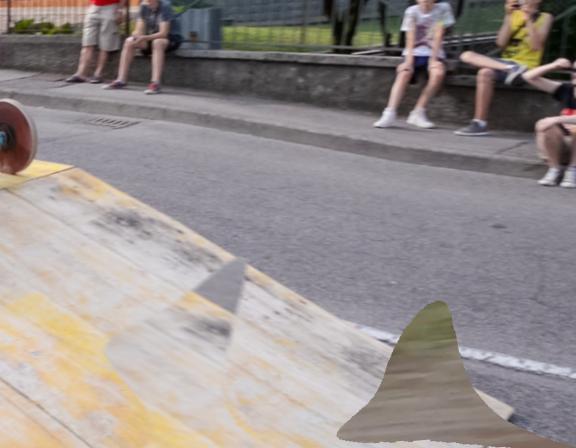}\,%
  \includegraphics{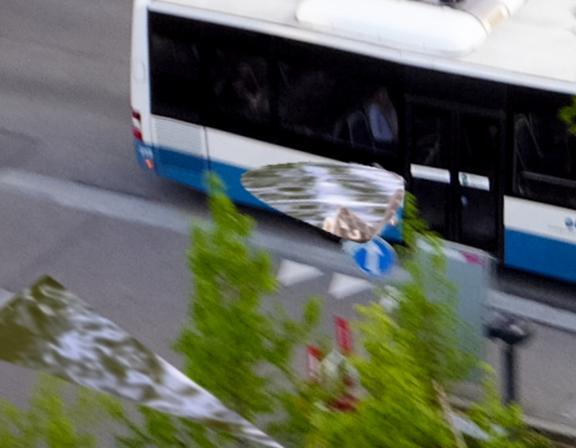}\,%
  \includegraphics{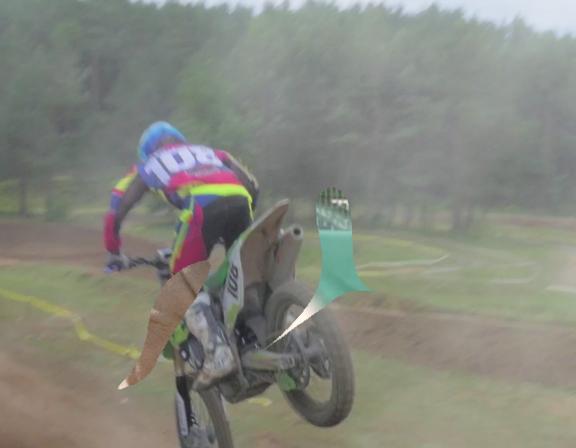}

  \includegraphics{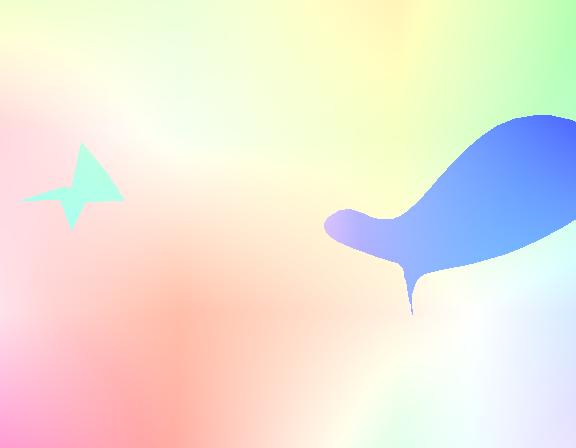}\,%
  \includegraphics{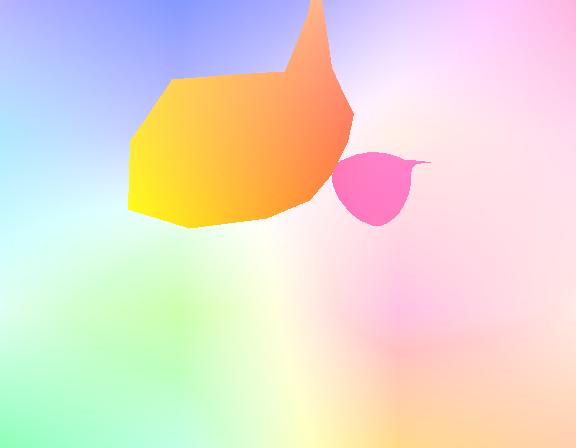}\,%
  \includegraphics{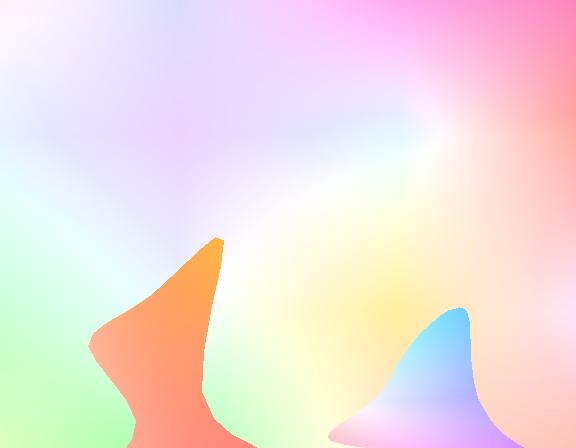}\,%
  \includegraphics{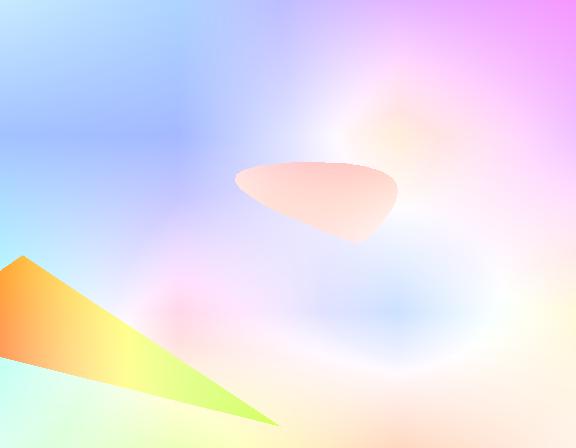}\,%
  \includegraphics{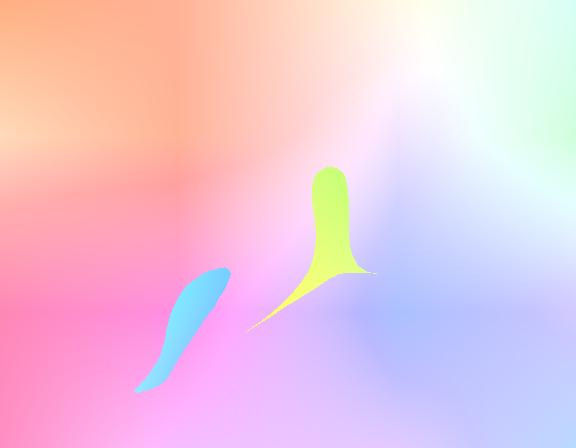}
  
  \caption{\textbf{AutoFlow samples  with two foreground objects}. First/fourth row: first image; second/fifth row: second image; third/sixth row: visualized optical flow. \textbf{Please go to our webpage \url{autoflow-google.github.io} to see the gif images.}}		
\label{fig:more:samples2}
\end{figure*}

\begin{figure*}[t]
  \centering
  \setkeys{Gin}{width=0.2\linewidth}
  
  \includegraphics{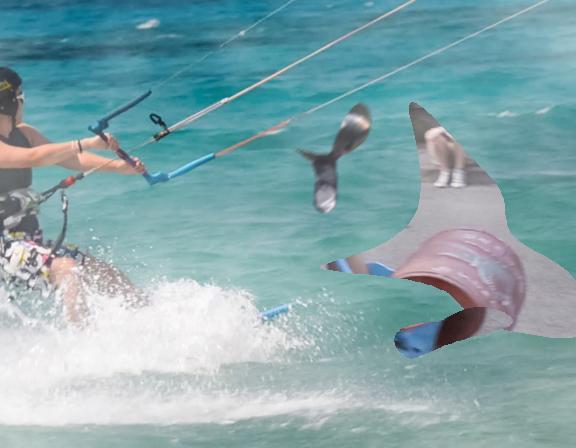}\,%
  \includegraphics{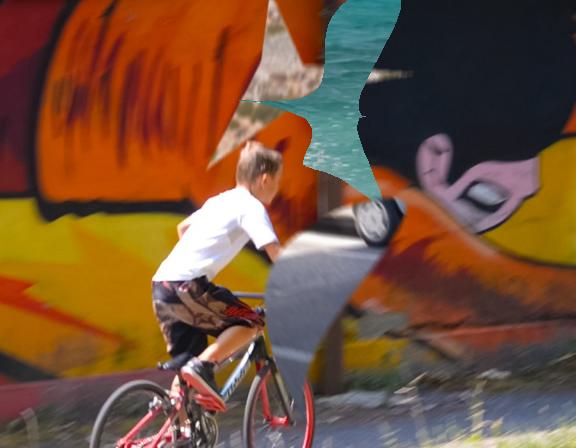}\,%
  \includegraphics{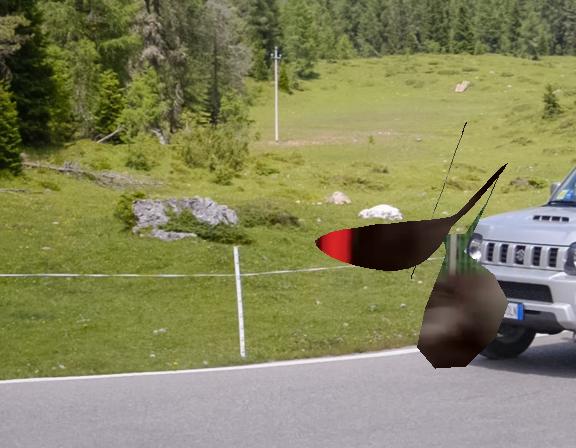}\,%
  \includegraphics{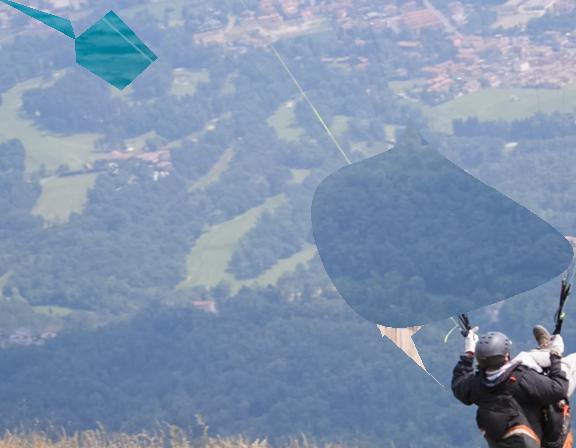}\,%
  \includegraphics{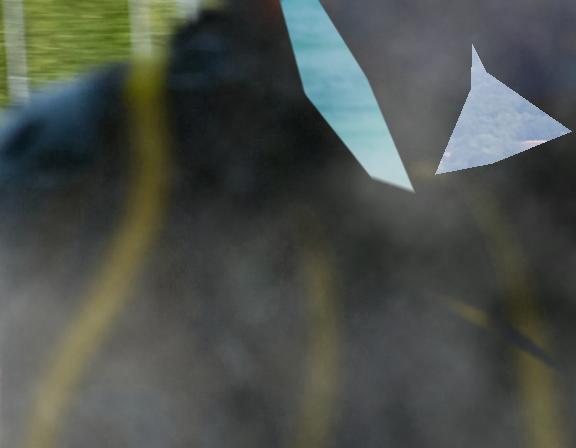}

  \includegraphics{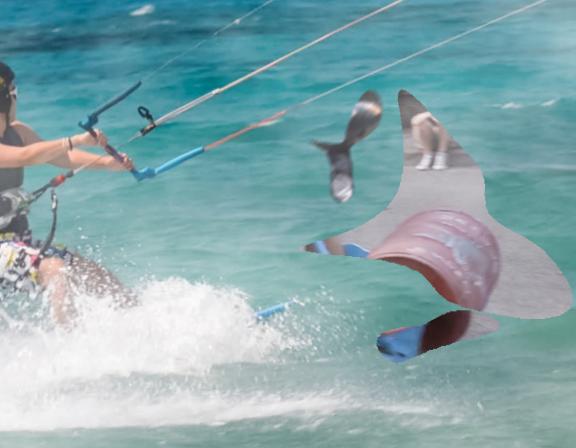}\,%
  \includegraphics{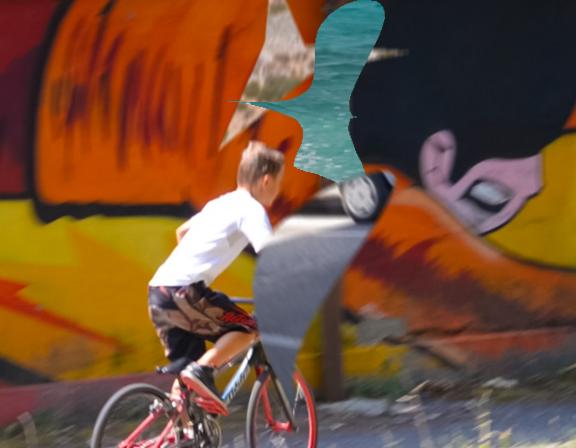}\,%
  \includegraphics{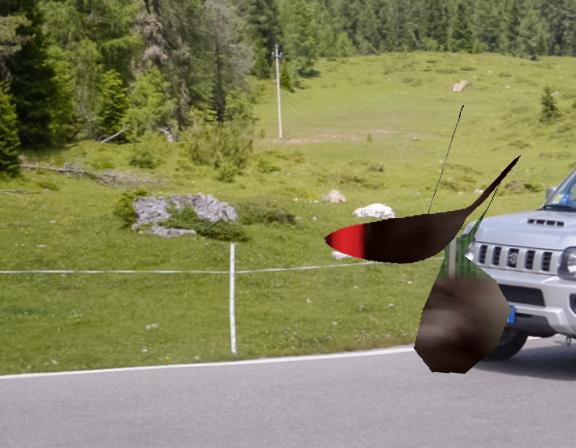}\,%
  \includegraphics{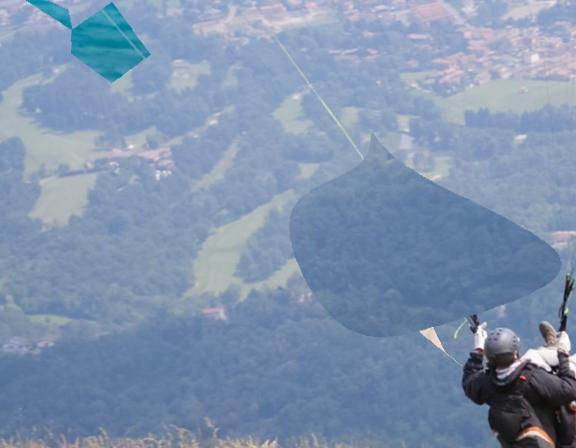}\,%
  \includegraphics{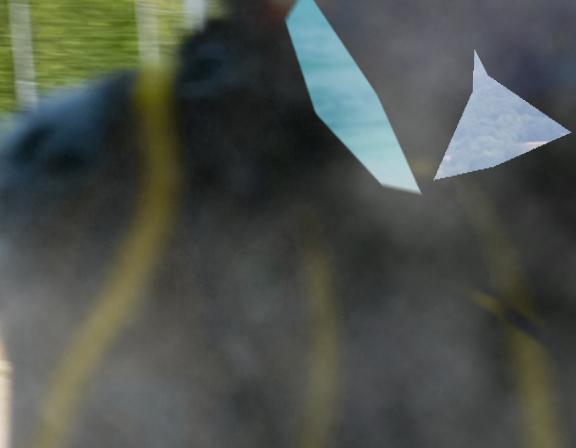}

  \includegraphics{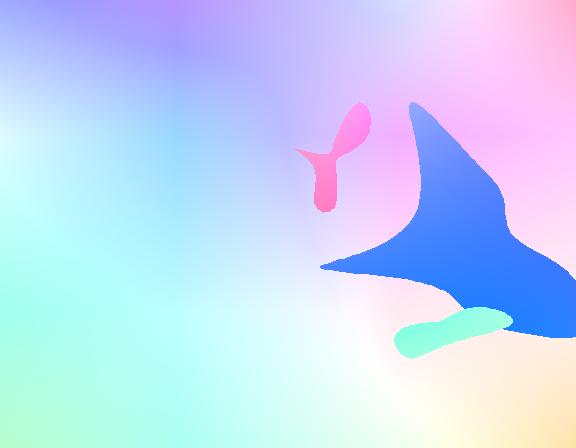}\,%
  \includegraphics{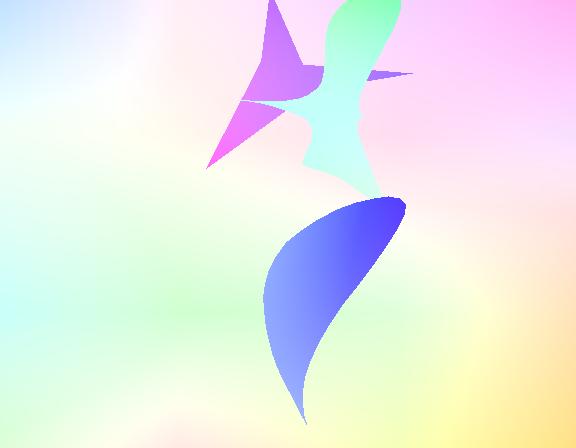}\,%
  \includegraphics{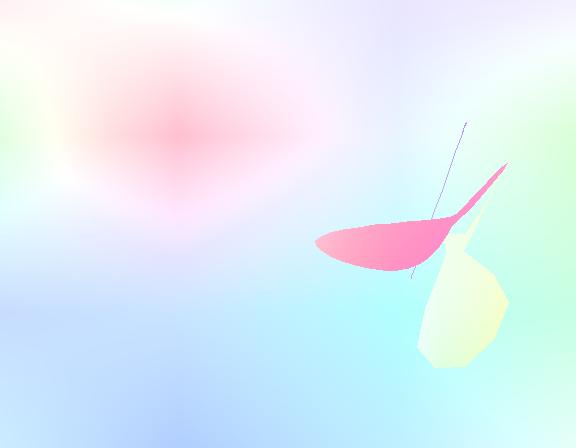}\,%
  \includegraphics{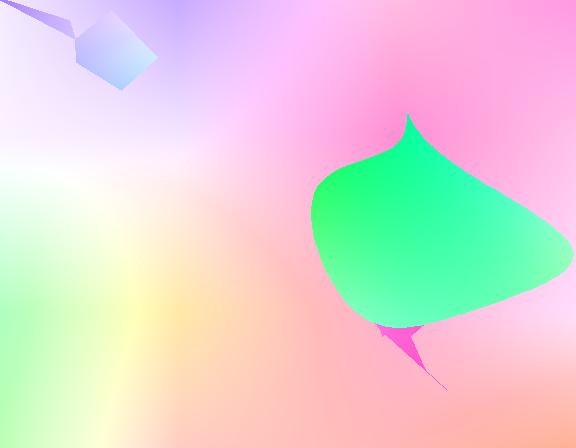}\,%
  \includegraphics{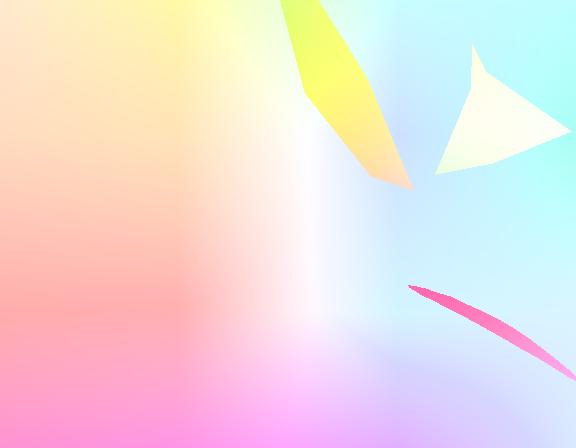}

  \includegraphics{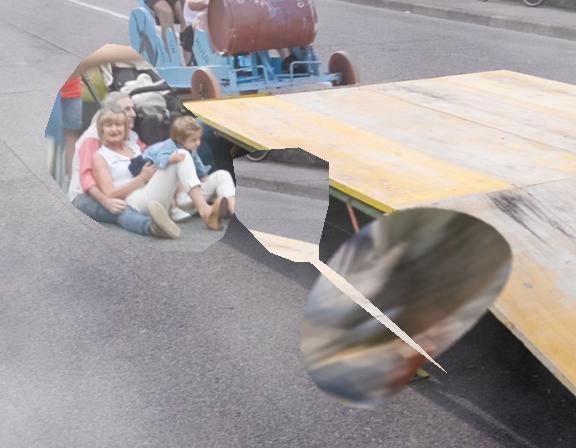}\,%
  \includegraphics{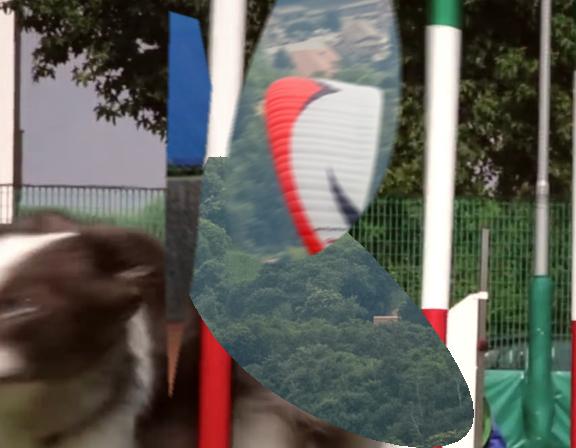}\,%
  \includegraphics{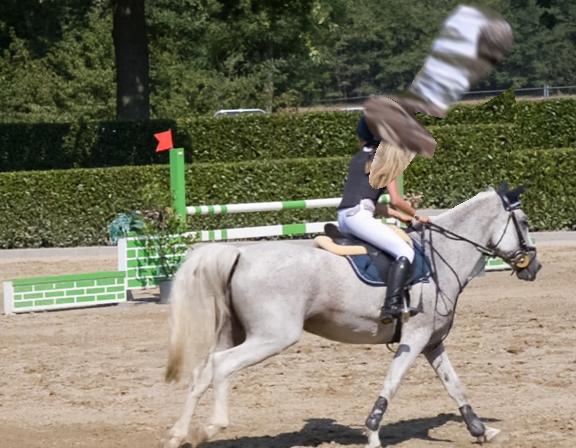}\,%
  \includegraphics{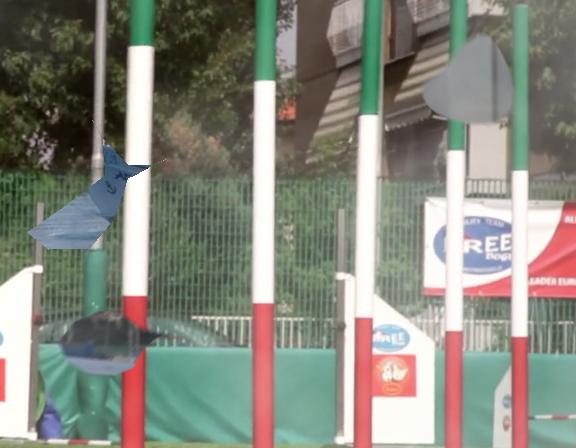}\,%
  \includegraphics{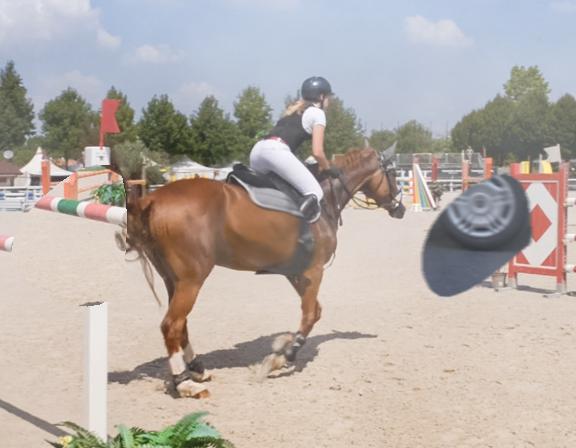}

  \includegraphics{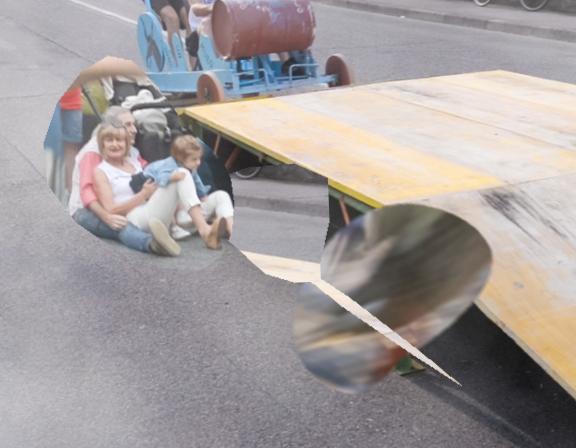}\,%
  \includegraphics{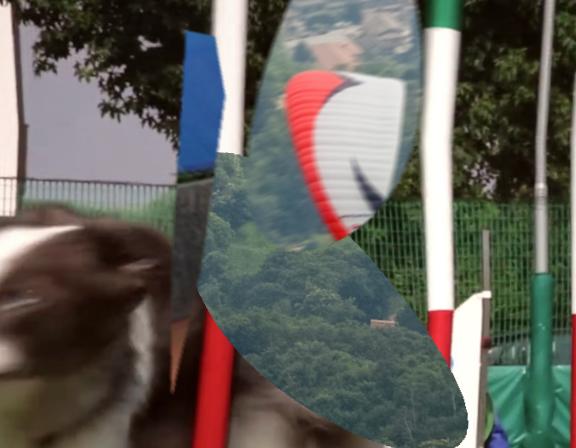}\,%
  \includegraphics{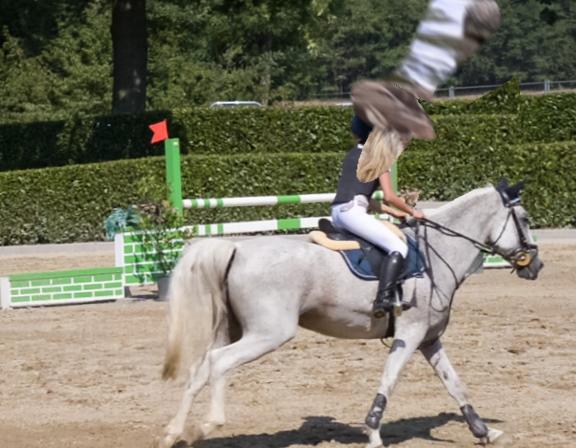}\,%
  \includegraphics{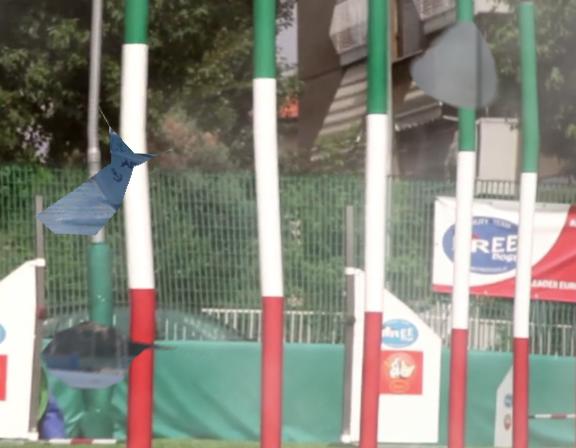}\,%
  \includegraphics{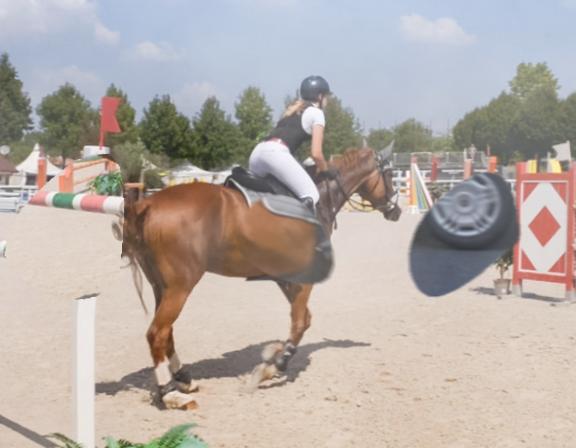}

  \includegraphics{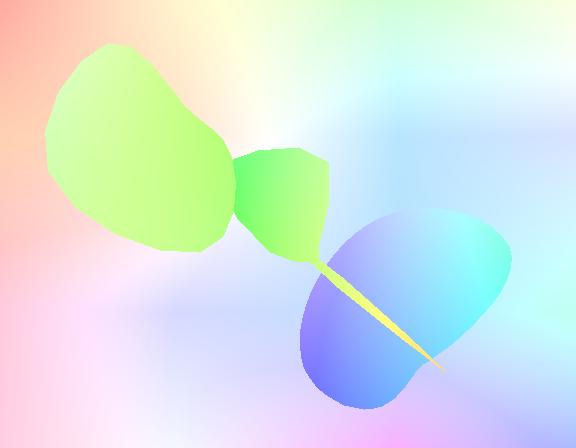}\,%
  \includegraphics{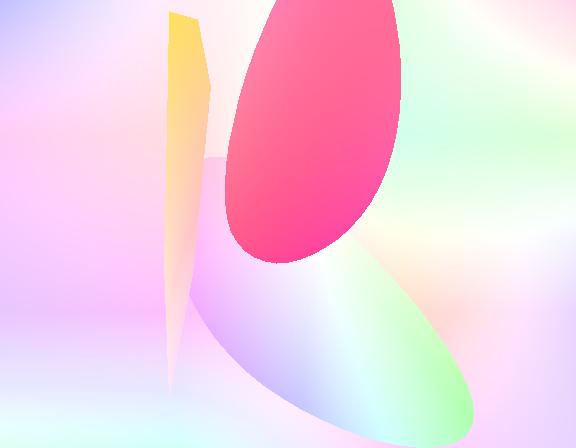}\,%
  \includegraphics{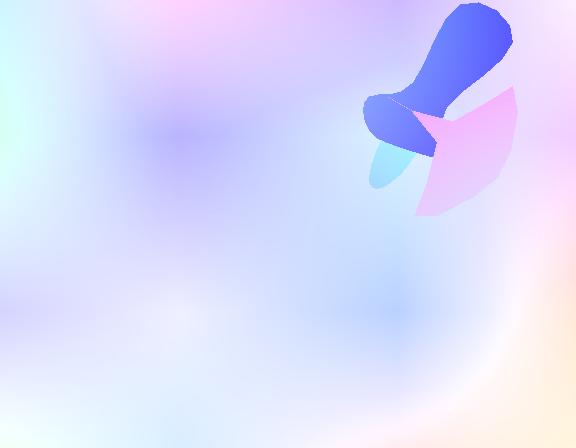}\,%
  \includegraphics{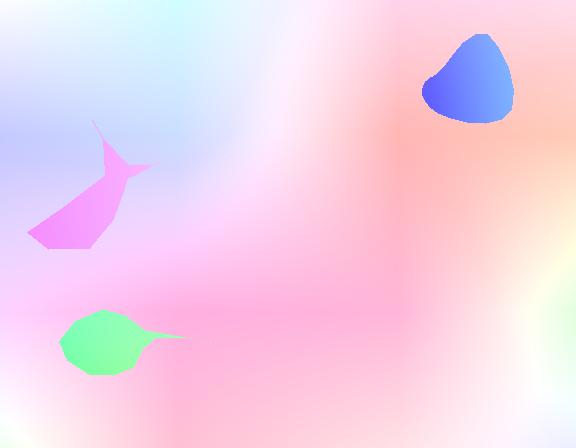}\,%
  \includegraphics{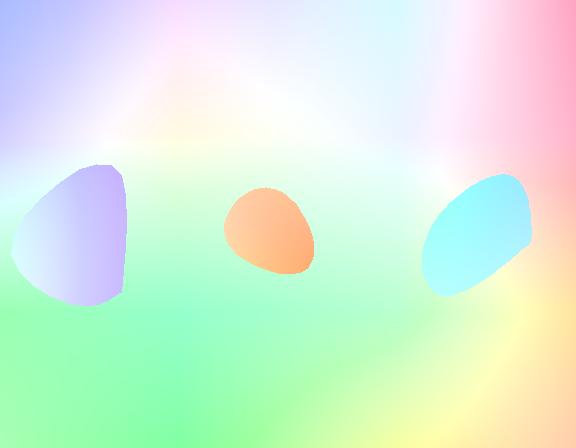}
  
  \caption{\textbf{AutoFlow samples  with three foreground objects}. First/fourth row: first image; second/fifth row: second image; third/sixth row: visualized optical flow. \textbf{Please go to our webpage \url{autoflow-google.github.io} to see the gif images.}}    
\label{fig:more:samples3}
\end{figure*}

\begin{figure*}[t]
  \centering
  \setkeys{Gin}{width=0.2\linewidth}
  
  \includegraphics{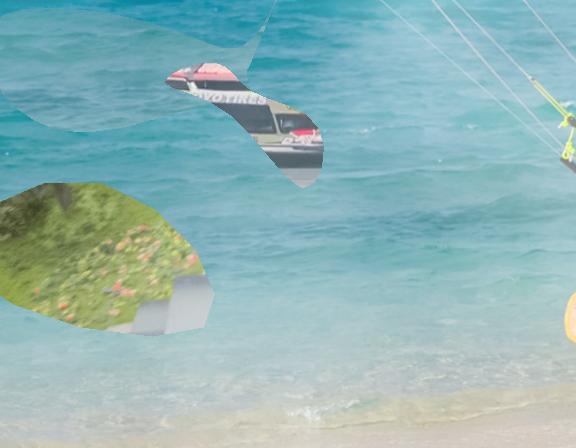}\,%
  \includegraphics{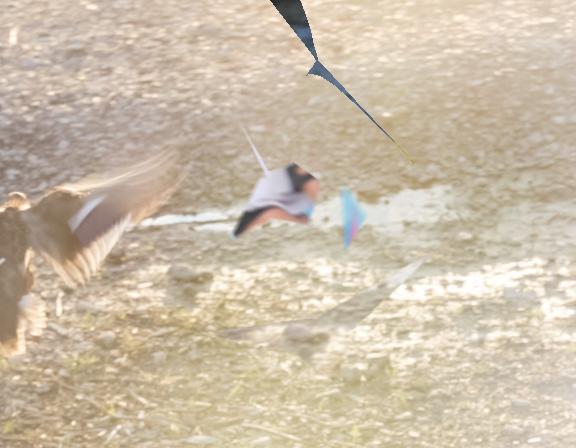}\,%
  \includegraphics{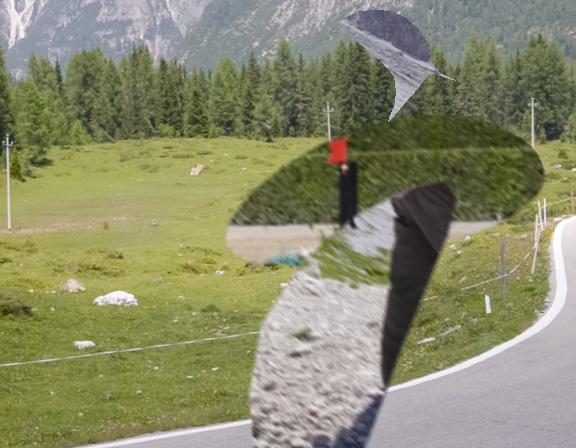}\,%
  \includegraphics{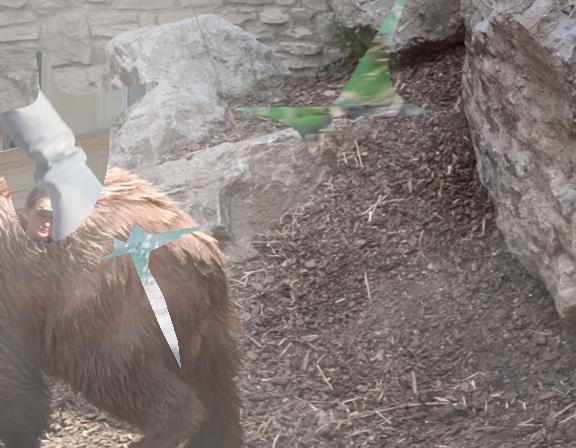}\,%
  \includegraphics{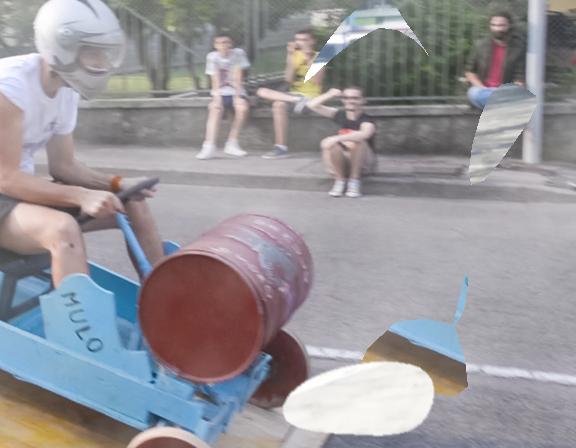}

  \includegraphics{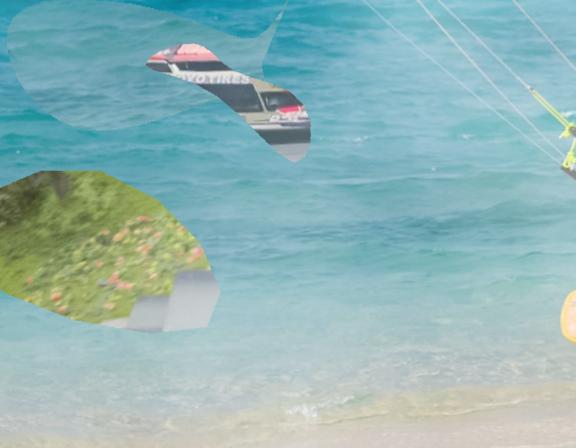}\,%
  \includegraphics{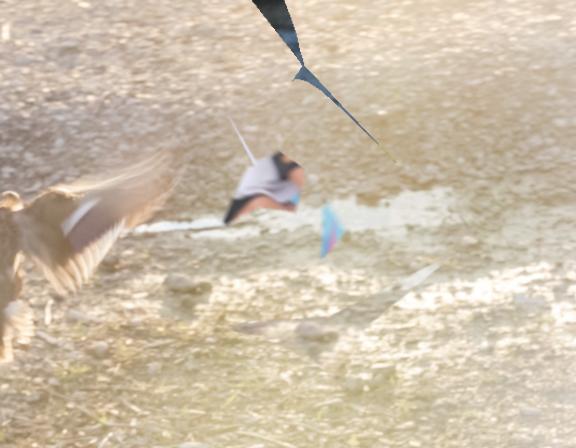}\,%
  \includegraphics{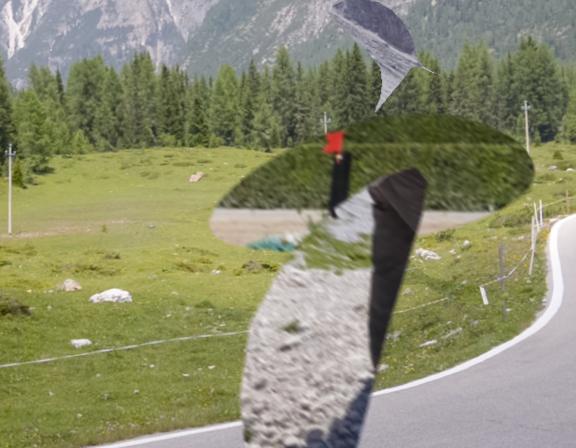}\,%
  \includegraphics{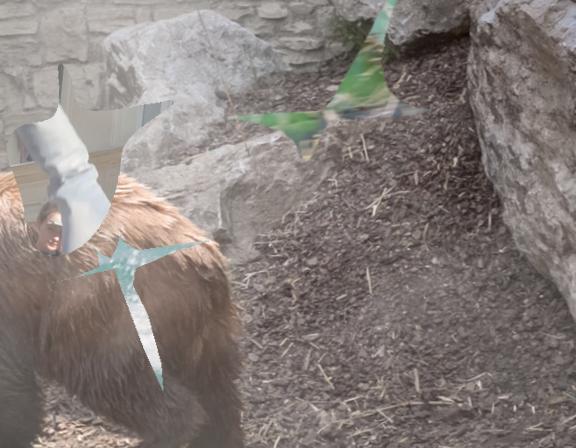}\,%
  \includegraphics{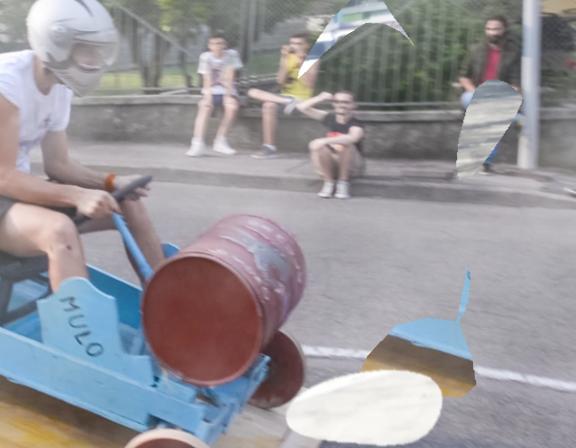}

  \includegraphics{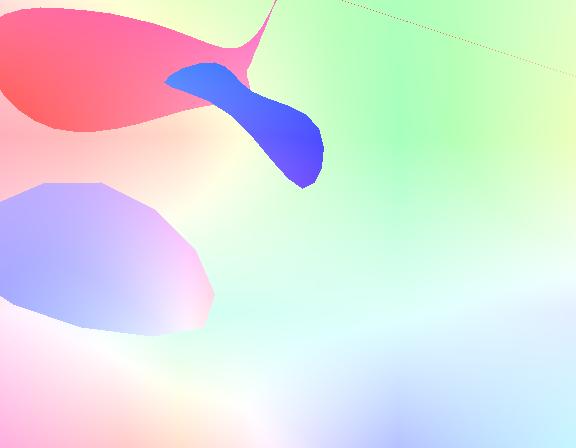}\,%
  \includegraphics{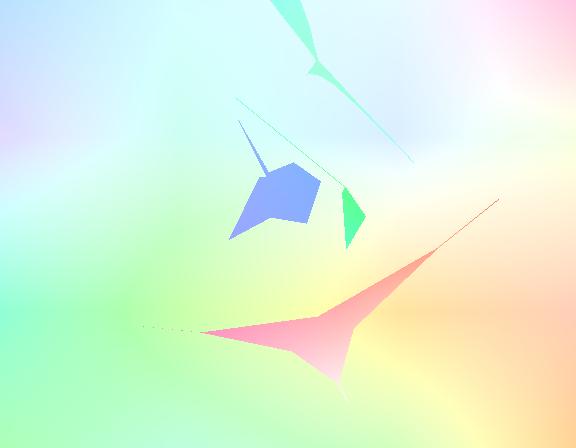}\,%
  \includegraphics{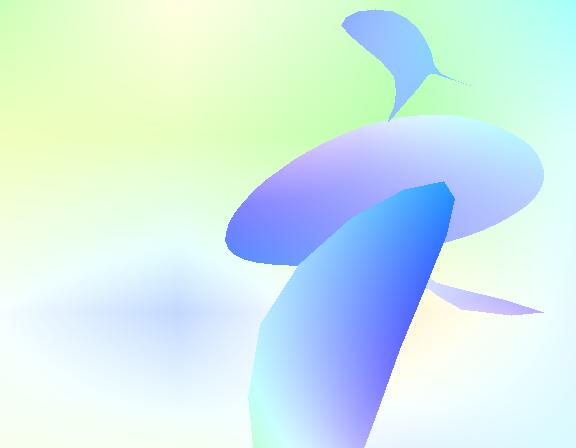}\,%
  \includegraphics{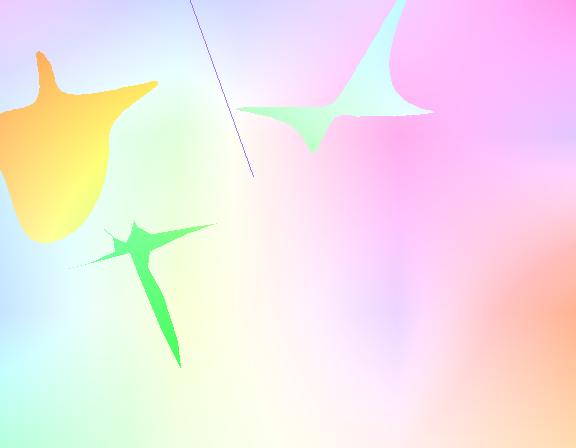}\,%
  \includegraphics{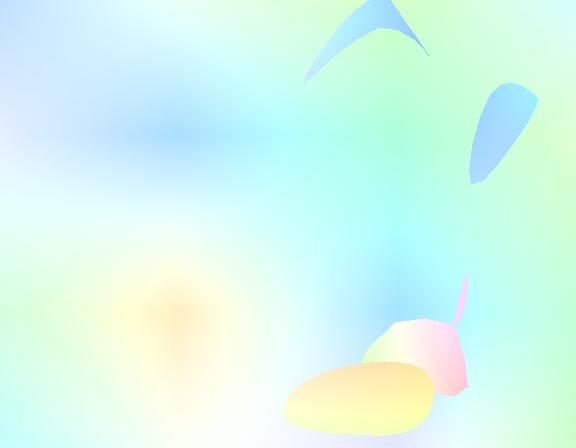}

  \includegraphics{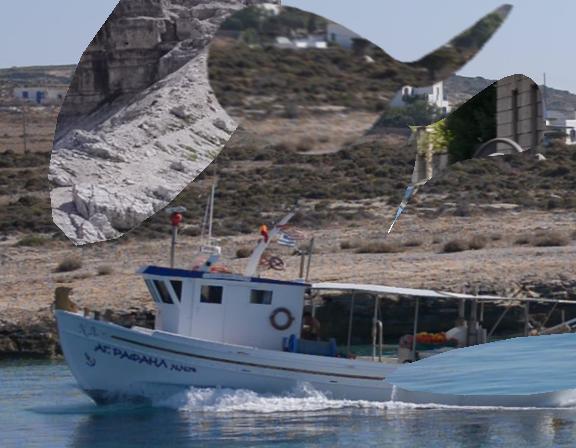}\,%
  \includegraphics{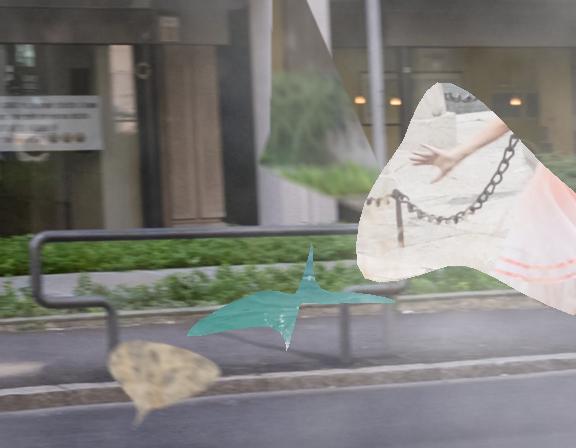}\,%
  \includegraphics{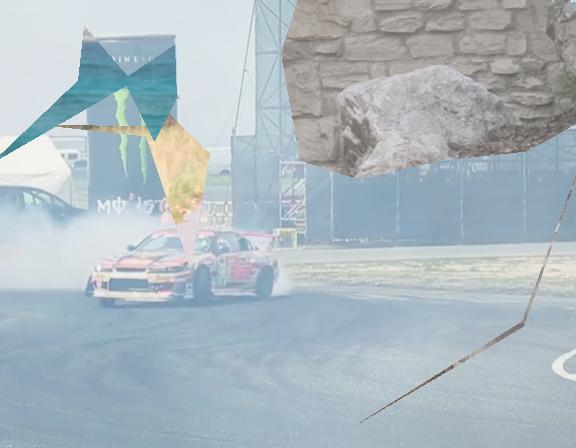}\,%
  \includegraphics{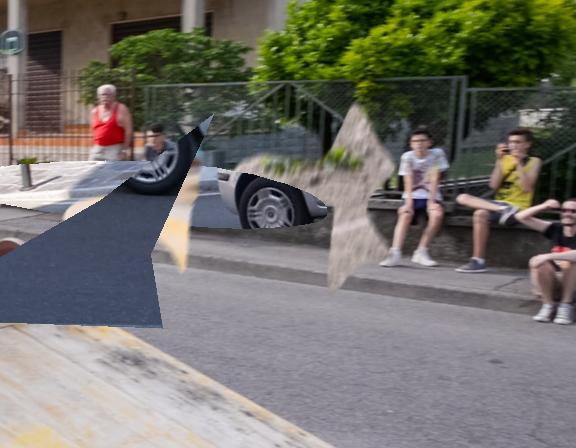}\,%
  \includegraphics{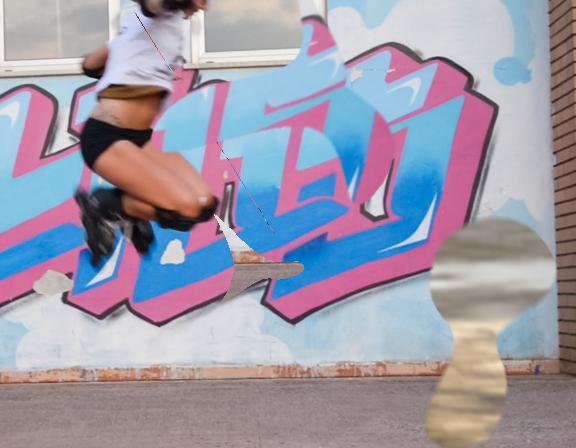}

  \includegraphics{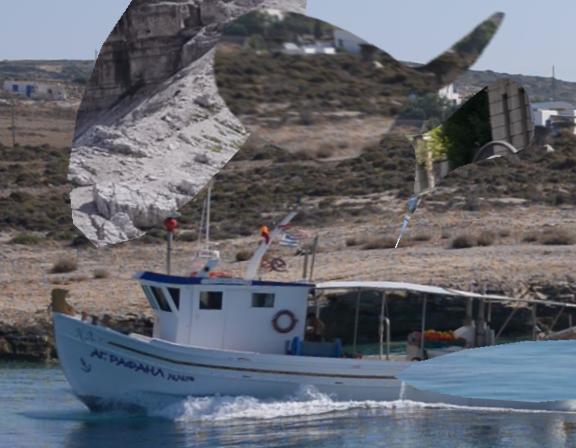}\,%
  \includegraphics{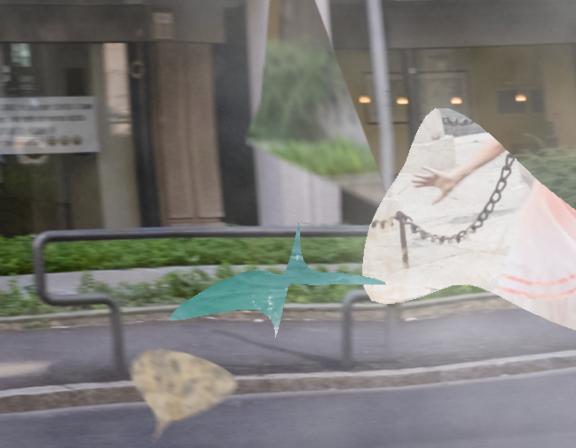}\,%
  \includegraphics{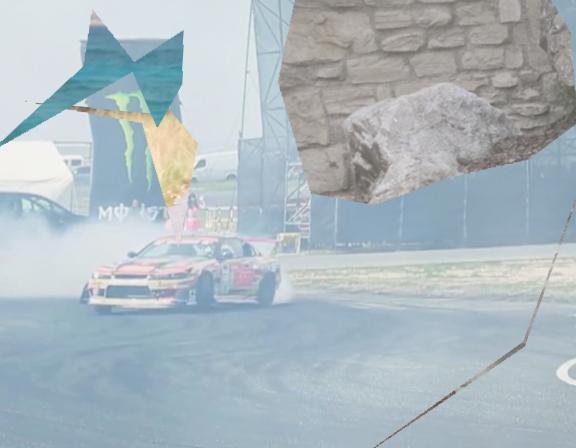}\,%
  \includegraphics{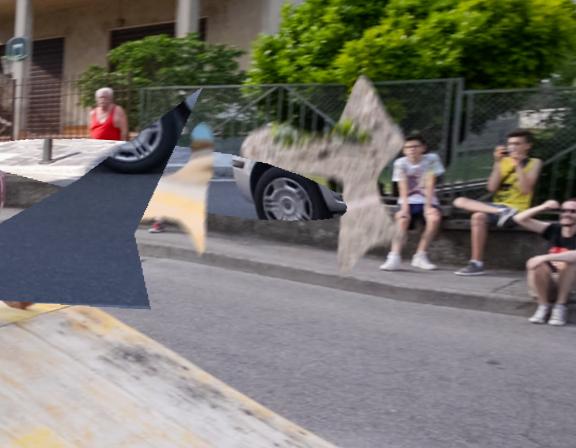}\,%
  \includegraphics{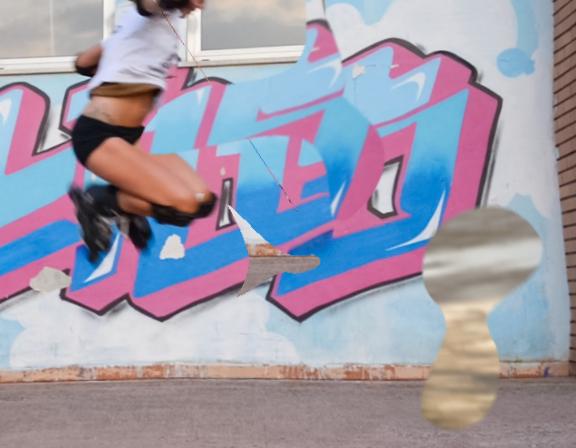}

  \includegraphics{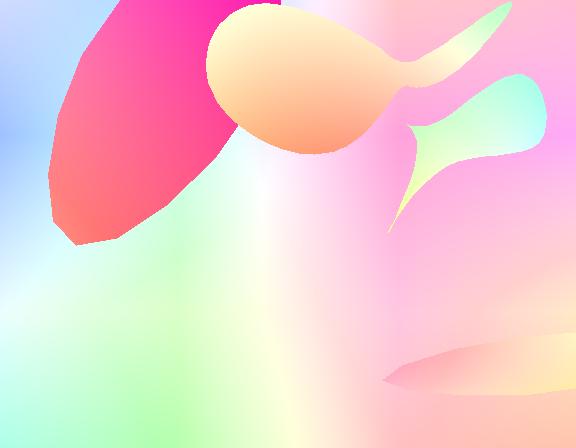}\,%
  \includegraphics{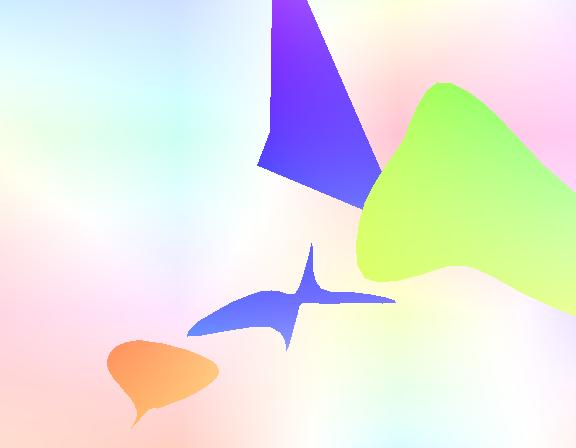}\,%
  \includegraphics{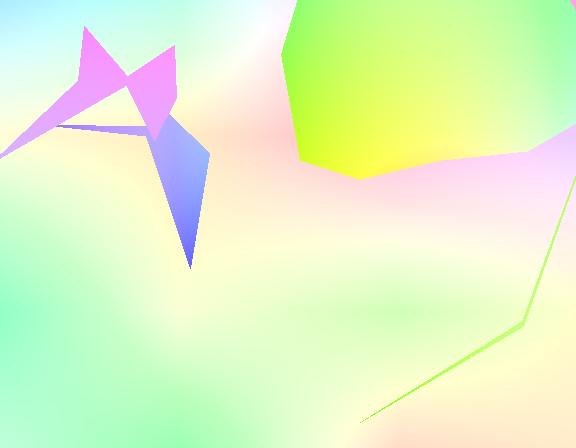}\,%
  \includegraphics{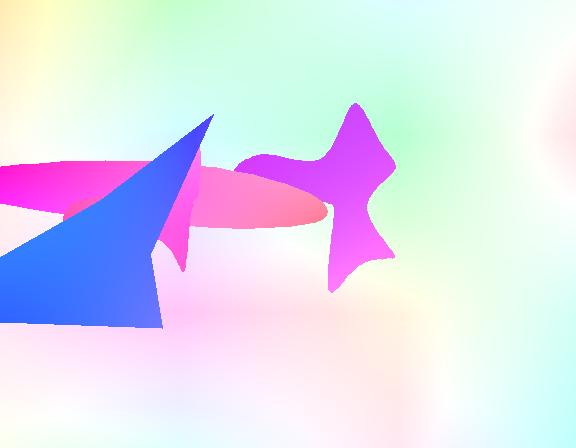}\,%
  \includegraphics{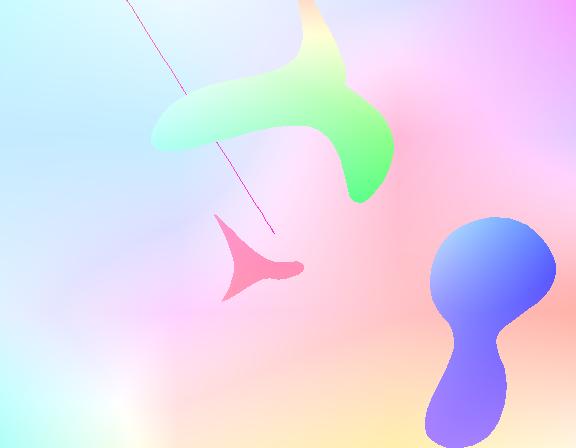}
  
  \caption{\textbf{AutoFlow samples  with four foreground objects}. First/fourth row: first image; second/fifth row: second image; third/sixth row: visualized optical flow. \textbf{Please go to our webpage \url{autoflow-google.github.io} to see the gif images.}}    
\label{fig:more:samples4}
\end{figure*}

\begin{figure*}[t]
  \centering
  \setkeys{Gin}{width=0.2\linewidth}
  
  \includegraphics{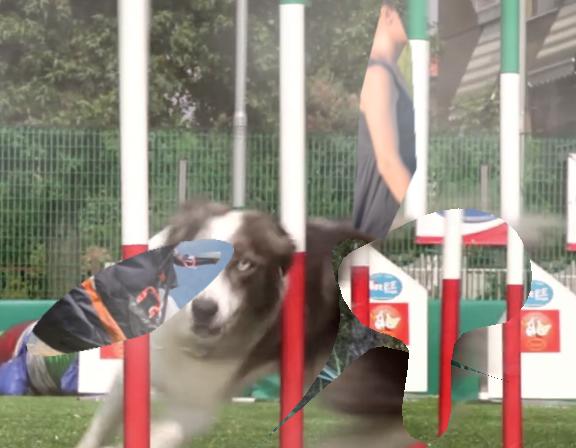}\,%
  \includegraphics{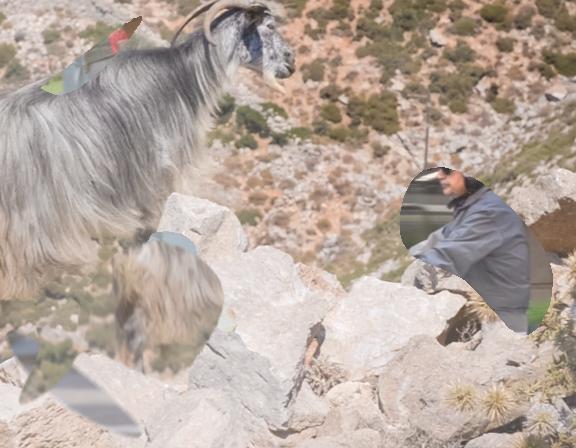}\,%
  \includegraphics{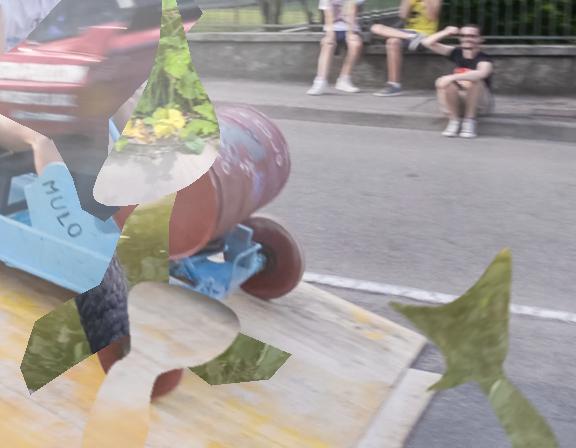}\,%
  \includegraphics{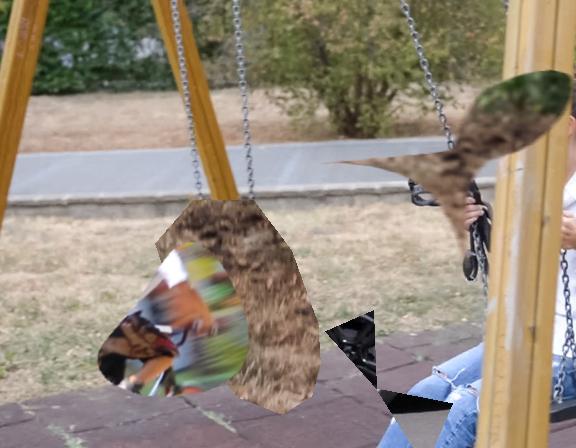}\,%
  \includegraphics{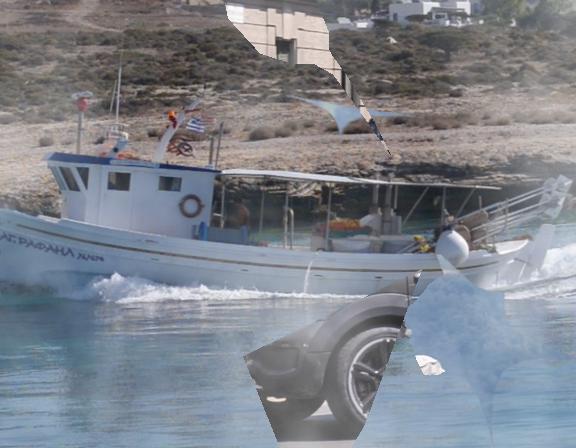}

  \includegraphics{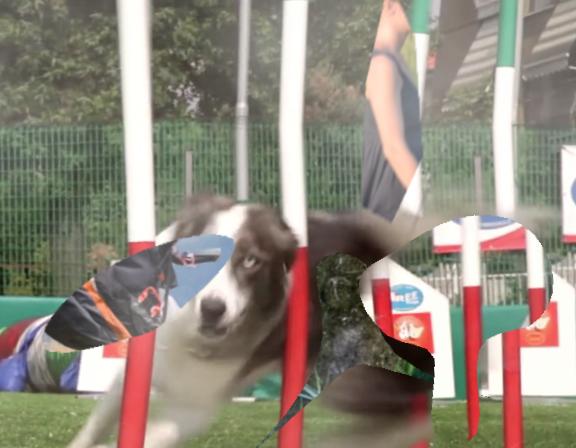}\,%
  \includegraphics{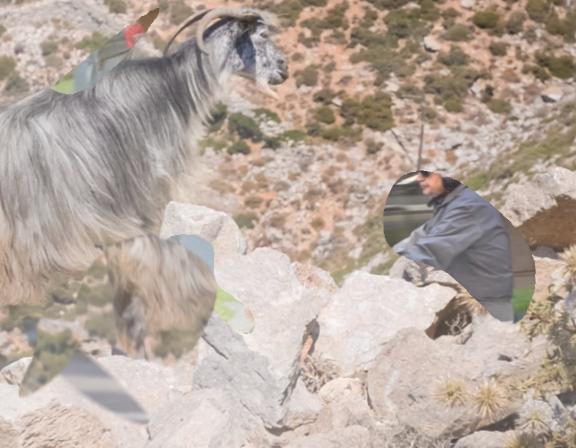}\,%
  \includegraphics{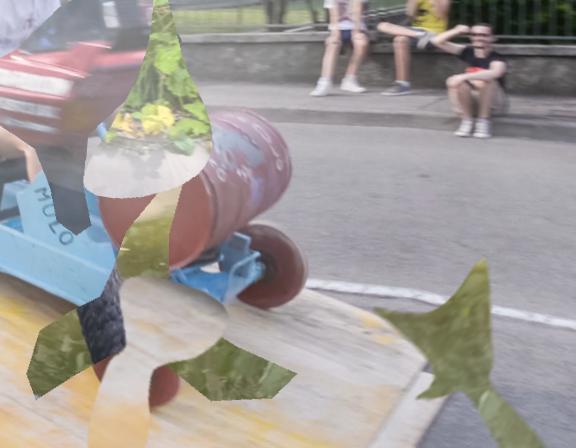}\,%
  \includegraphics{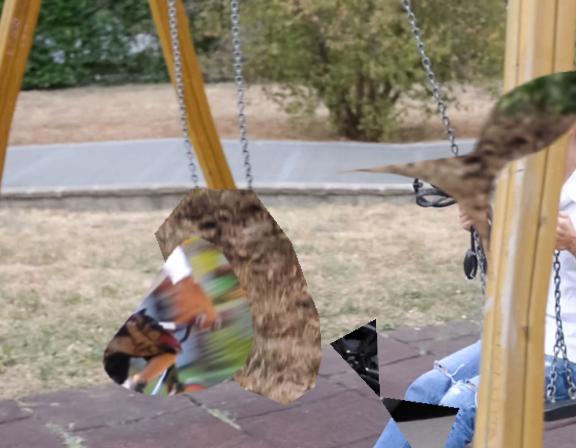}\,%
  \includegraphics{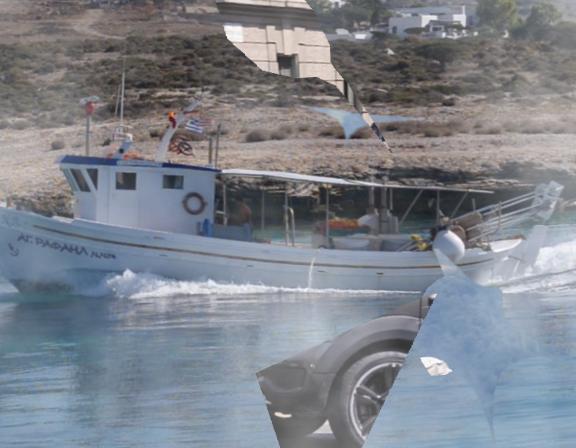}

  \includegraphics{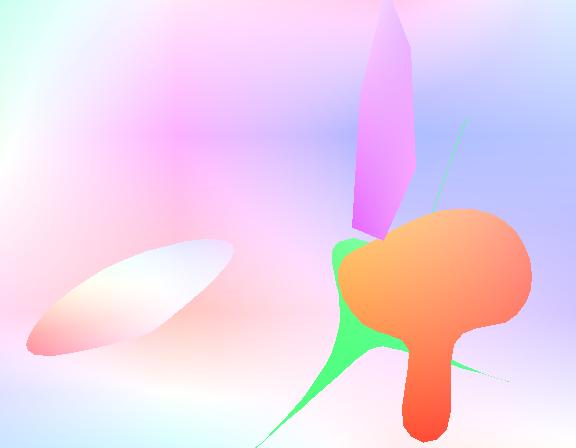}\,%
  \includegraphics{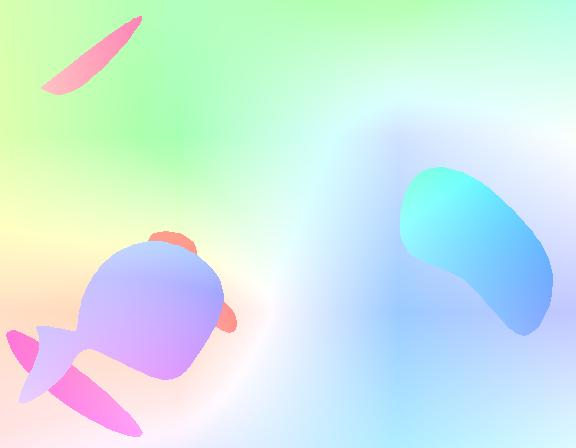}\,%
  \includegraphics{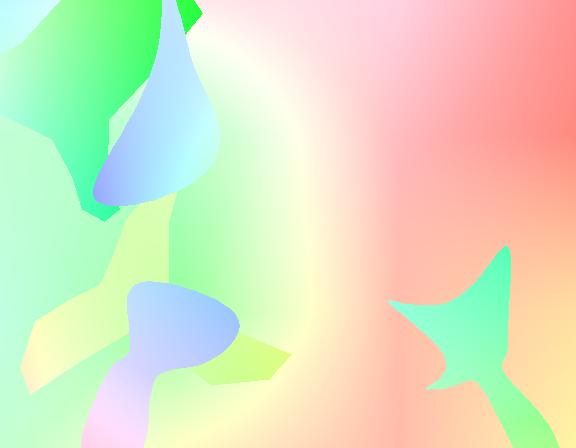}\,%
  \includegraphics{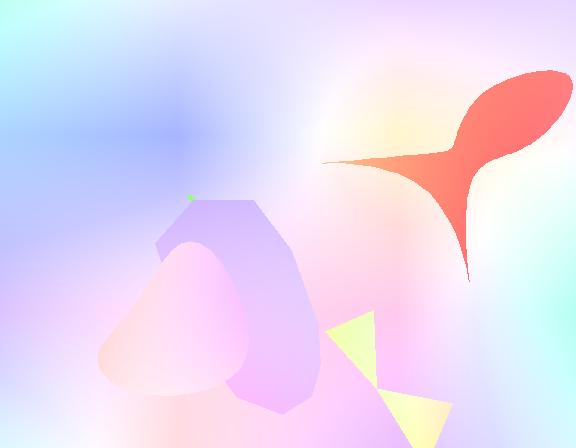}\,%
  \includegraphics{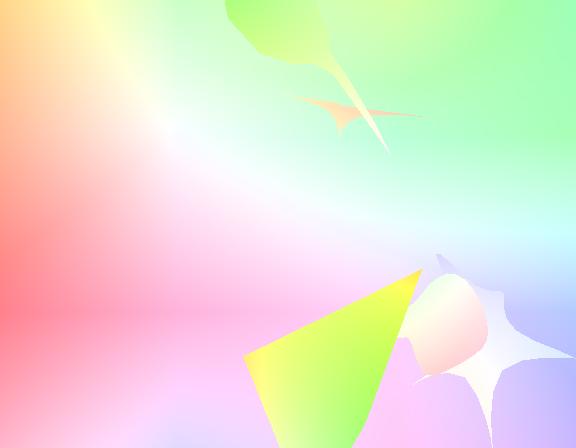}

  \includegraphics{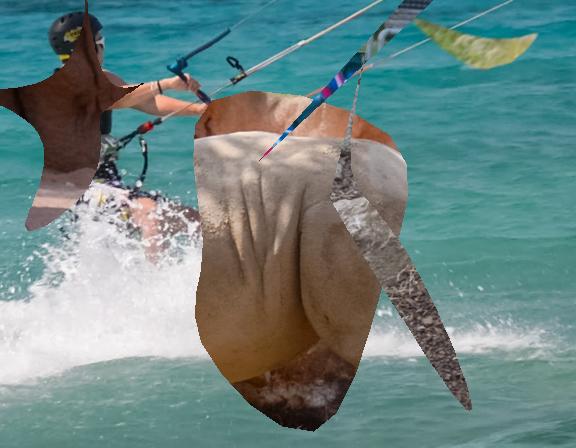}\,%
  \includegraphics{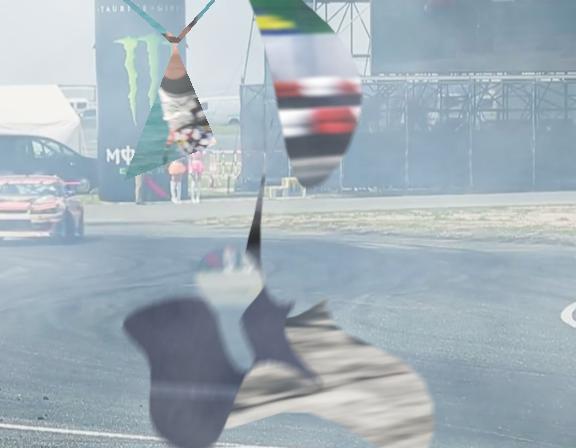}\,%
  \includegraphics{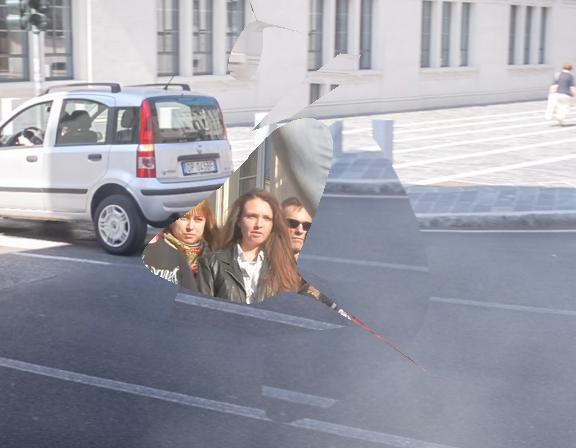}\,%
  \includegraphics{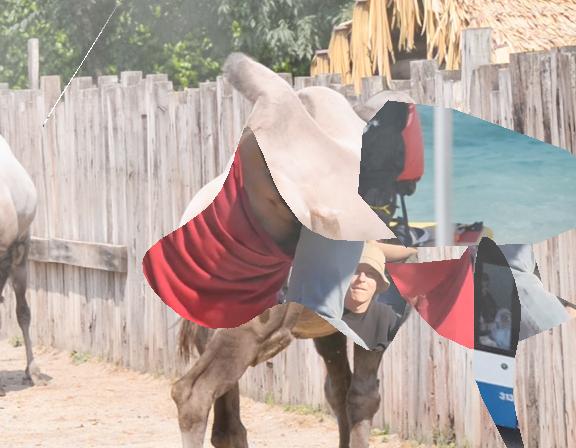}\,%
  \includegraphics{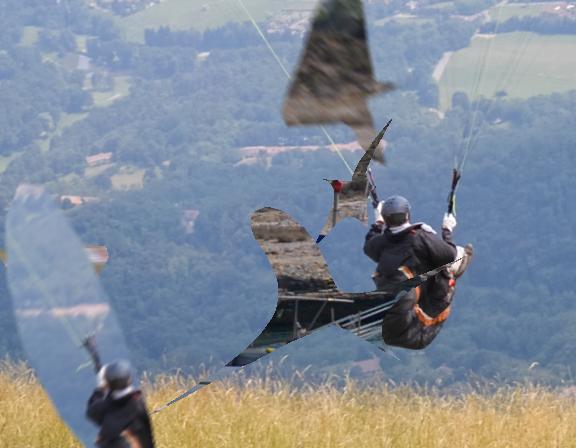}

  \includegraphics{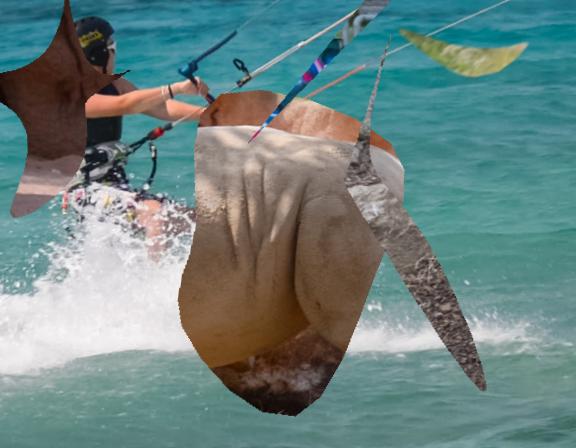}\,%
  \includegraphics{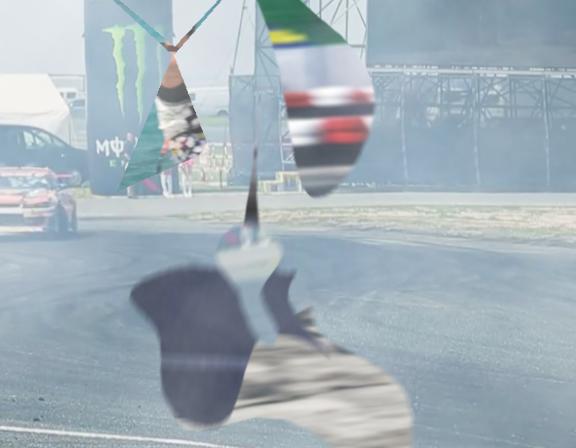}\,%
  \includegraphics{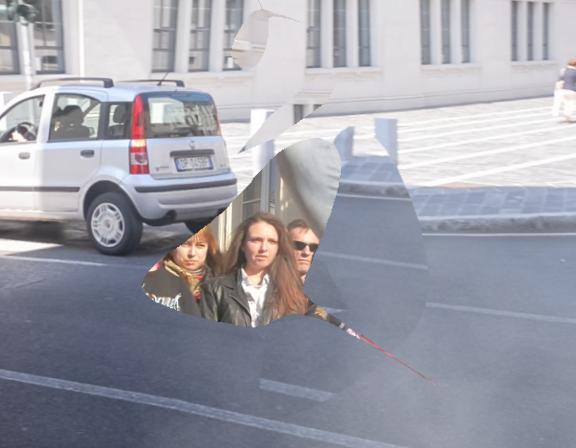}\,%
  \includegraphics{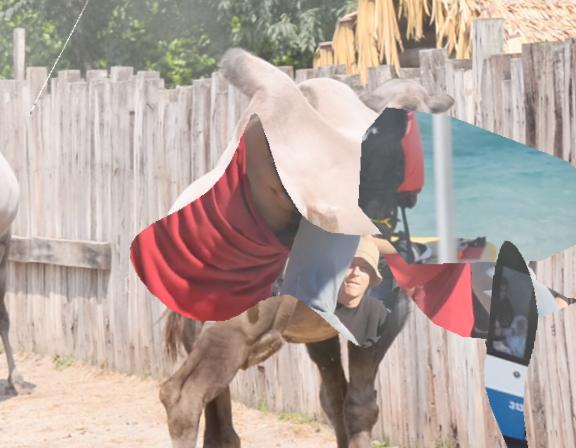}\,%
  \includegraphics{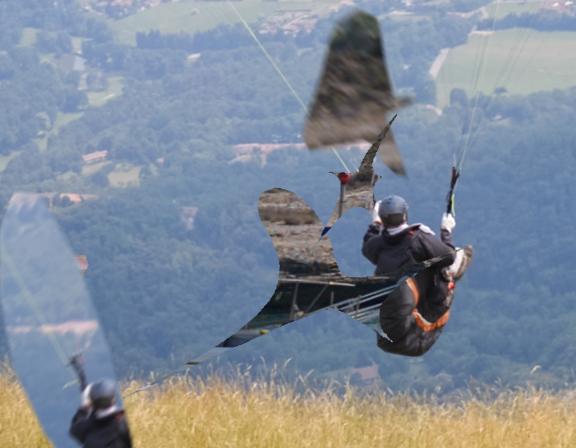}

  \includegraphics{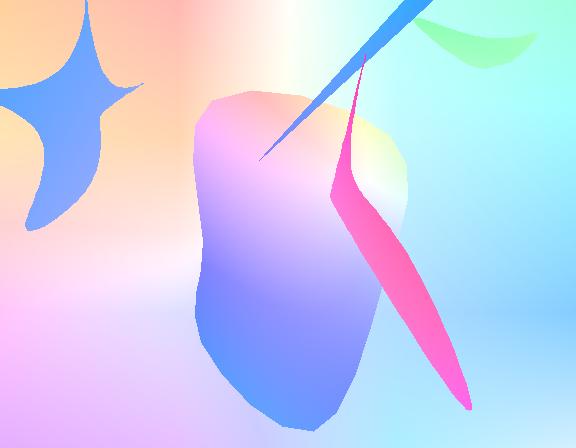}\,%
  \includegraphics{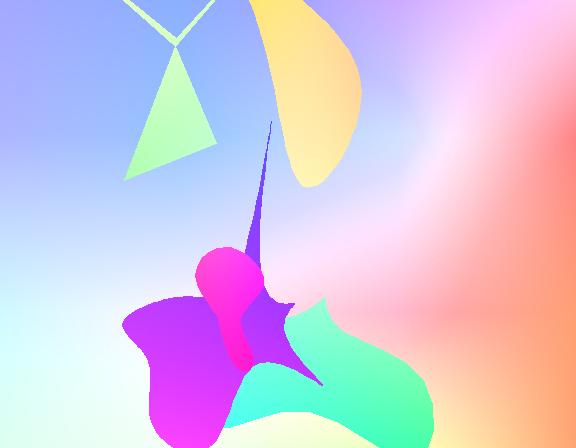}\,%
  \includegraphics{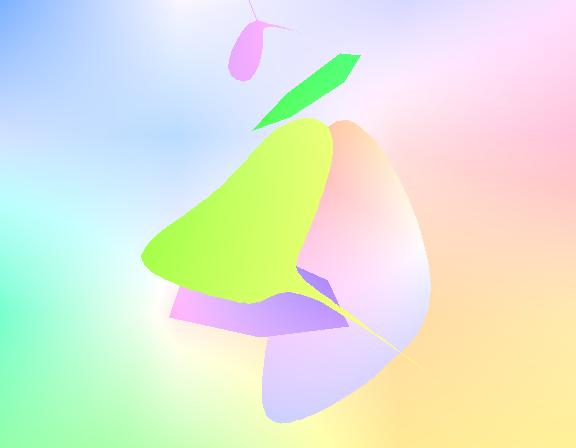}\,%
  \includegraphics{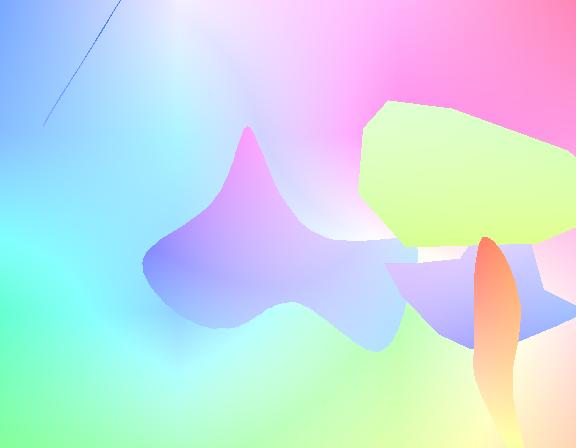}\,%
  \includegraphics{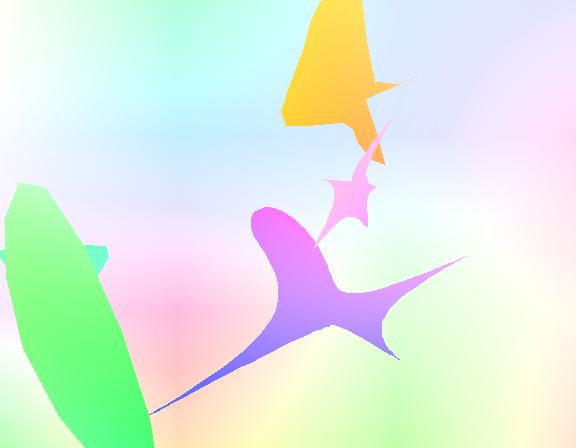}
  
  \caption{\textbf{AutoFlow samples  with five foreground objects}. First/fourth row: first image; second/fifth row: second image; third/sixth row: visualized optical flow. \textbf{Please go to our webpage \url{autoflow-google.github.io} to see the gif images.}}    
\label{fig:more:samples5}
\end{figure*}

Figures~\ref{fig:more:samples1}-\ref{fig:more:samples5} show some more samples of the image pairs and flow field. \textbf{Please go to  our webpage \url{autoflow-google.github.io} to see the gif images.}

\end{document}